%% file: main.tex
\documentclass[letterpaper, 10 pt, conference]{ieeeconf}  % Comment this line out if you need a4paper

\IEEEoverridecommandlockouts                              % This command is only needed if 
\usepackage{times}
\usepackage{listings}
\usepackage{multicol}
\usepackage[bookmarks=true]{hyperref}
\usepackage{xspace}

\usepackage{hyperref}
\usepackage{url}

\usepackage{amsmath,amssymb,amsthm}
\usepackage{cleveref}
\usepackage{placeins}
\usepackage{graphicx}
\usepackage{multirow}
\usepackage{booktabs}
\usepackage[table]{xcolor}
\usepackage{float} % in preamble
\usepackage{arydshln} % in preamble
\usepackage{rotating} % for vertical text if needed
\usepackage[percent]{overpic} % put this in the preamble
\usepackage{subcaption} % for subtables
\usepackage{algorithm}
\usepackage{algpseudocode}
\usepackage{tikz}
\usepackage[percent]{overpic}
\usepackage{soul}
\usepackage{calc}   
\usepackage{cuted}
\usepackage{censor}
\usepackage{cite}
\usepackage{dblfloatfix} % or \usepackage{stfloats}
\definecolor{lightblue}{RGB}{220, 235, 255}
\definecolor{lightgreen}{RGB}{180, 255, 180}
\definecolor{lightorange}{RGB}{255,200,150}
\definecolor{lightred}{RGB}{255,180,180} 

\definecolor{myGreen}{HTML}{D5E8D4} % example hex code
\definecolor{myOrange}{HTML}{FFE6CC} \definecolor{myRed}{HTML}{F8CECC}% soft 

\newcommand{\acronym}{\texttt{CAPLR}\xspace}
\newcommand{\acronymS}{\texttt{CAPLR$_s$}\xspace}
\newcommand{\acronymM}{\texttt{CAPLR$_m$}\xspace}
\title{\LARGE \bf
Unsupervised Domain Adaptation for Sim-to-Real Object Pose Estimation with Contrastive Alignment and Pseudo-Label Refinement
}

\author{{Nidhal Eddine Chenni, Arunkumar Rathinam and Djamila Aouada }
\thanks{The authors are associated with SnT, University of Luxembourg. Contact emails: \{nidhal.chenni, arunkumar.rathinam, djamila.aouada\}@uni.lu}% <-this % stops a space
\thanks{The work was funded by Luxembourg National Research Fund under the project reference BRIDGES/2025/IS/19855391/U-ADAPT, and LMO space.}% <-this % stops a space
}

\begin{document}

\maketitle
\thispagestyle{empty}
\pagestyle{empty}
\begin{strip}
\centering
\input{sections/fig1_content_only} % The tabular and caption
\end{strip}

%%%%%%%%%%%%%%%%%%%%%%%%%%%%%%%%%%%%%%%%%%%%%%%%%%%%%%%%%%%%%%%%%%%%%%%%%%%%%%%%

\input{sections/0.Abstarct}
\input{sections/1.Introduction}
\input{sections/2.RelatedWorks}

\input{sections/fig2}
\input{sections/3.Methodology}

\input{sections/4.Experiments}

\input{sections/5.Conclusion}

%%%%%%%%%%%%%%%%%%%%%%%%%%%%%%%%%%%%%%%%%%%%%%%%%%%%%%%%%%%%%%%%%%%%%%%%%%%%%%%%



% \section*{ACKNOWLEDGMENT}

% The preferred spelling of the word ÒacknowledgmentÓ in America is without an ÒeÓ after the ÒgÓ. Avoid the stilted expression, ÒOne of us (R. B. G.) thanks . . .Ó  Instead, try ÒR. B. G. thanksÓ. Put sponsor acknowledgments in the unnumbered footnote on the first page.

%%%%%%%%%%%%%%%%%%%%%%%%%%%%%%%%%%%%%%%%%%%%%%%%%%%%%%%%%%%%%%%%%%%%%%%%%%%%%%%%

\vspace{-0.1cm}

% \scriptsize
\bibliographystyle{IEEEtran}
\bibliography{IEEEabrv,references}

\newpage
\twocolumn[
\begin{center}
\vspace{0.5em}
{\Large \bfseries Supplementary Material}
\vspace{0.15em}
\end{center}
]
\setcounter{section}{0}
\setcounter{figure}{0}
\input{appendix/1.analysis}

\input{appendix/2.implementation}

\input{appendix/3.visuals}
\end{document}

%% file: sections/fig1_content_only.tex
\vspace{-5em}
\centering
\setlength{\tabcolsep}{2pt} 

\resizebox{0.9\linewidth}{!}{% % Forces the table to stay inside the page margins
\begin{tabular}{cccccccc}
    % & \multicolumn{2}{c}{LineMOD} & \multicolumn{2}{c}{Occluded-LineMOD} & \multicolumn{1}{c}{HomeBrewedDB} & \multicolumn{1}{c}{SPEED+ Li.box} & SPEED+ S.lamp \\
    \rotatebox{90}{\small\hspace{4pt}Groundtruth}  
    & \includegraphics[width=0.13\textwidth]{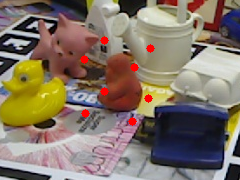}  
    & \includegraphics[width=0.13\textwidth]{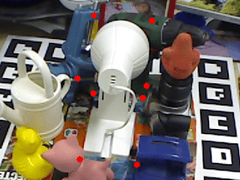}  
    & \includegraphics[width=0.13\textwidth]{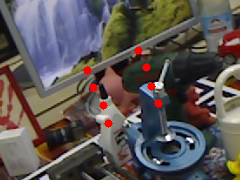}  
    & \includegraphics[width=0.13\textwidth]{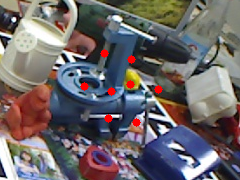}  
    & \includegraphics[width=0.13\textwidth]{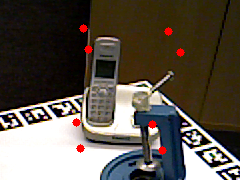}  
    & \includegraphics[width=0.13\textwidth]{figures/fig1/4740_light.png}  
    & \includegraphics[width=0.13\textwidth]{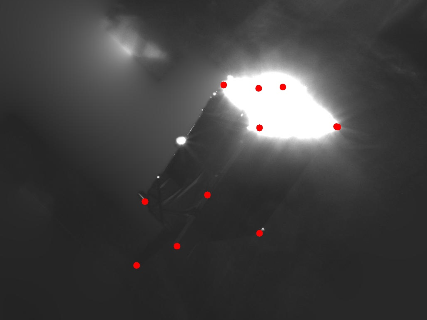} \\
    
    \rotatebox{90}{\small\hspace{8pt}w/o UDA}
    & \includegraphics[width=0.13\textwidth]{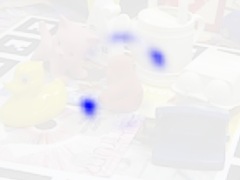}  
    & \includegraphics[width=0.13\textwidth]{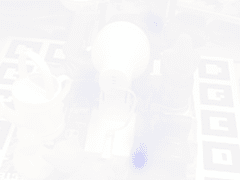}  
    & \includegraphics[width=0.13\textwidth]{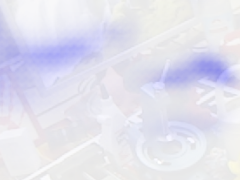}  
    & \includegraphics[width=0.13\textwidth]{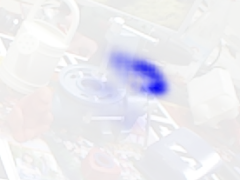}  
    & \includegraphics[width=0.13\textwidth]{figures/fig1/phone_pred.png}  
    & \includegraphics[width=0.13\textwidth]{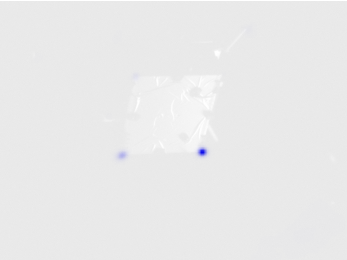}  
    & \includegraphics[width=0.13\textwidth]{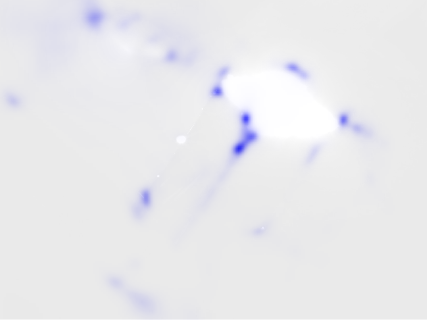} \\

    \rotatebox{90}{\small\hspace{8pt}\acronym}  
    & \includegraphics[width=0.13\textwidth]{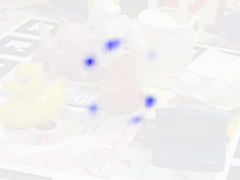}  
    & \includegraphics[width=0.13\textwidth]{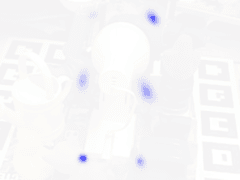}  
    & \includegraphics[width=0.13\textwidth]{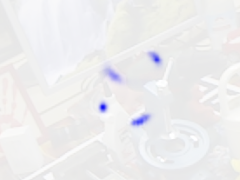}  
    & \includegraphics[width=0.13\textwidth]{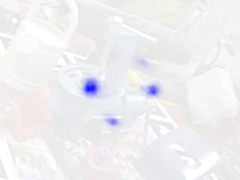}  
    & \includegraphics[width=0.13\textwidth]{figures/fig1/phone_caplr.png}  
    & \includegraphics[width=0.13\textwidth]{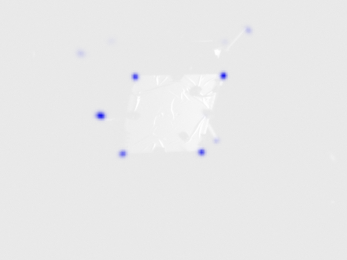}  
    & \includegraphics[width=0.13\textwidth]{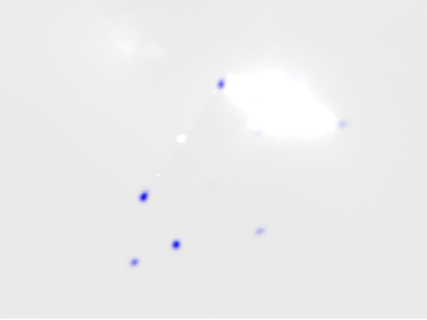} \\

    \rotatebox{90}{\small\hspace{3pt}Object pose}  
    & \includegraphics[width=0.13\textwidth]{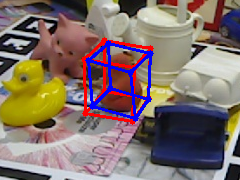}  
    & \includegraphics[width=0.13\textwidth]{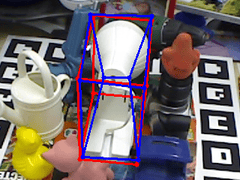}  
    & \includegraphics[width=0.13\textwidth]{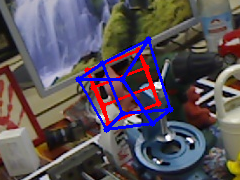}  
    & \includegraphics[width=0.13\textwidth]{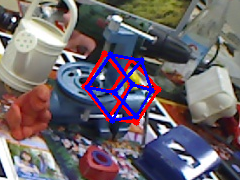}  
    & \includegraphics[width=0.13\textwidth]{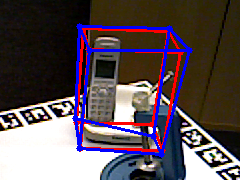}  
    & \includegraphics[width=0.13\textwidth]{figures/fig1/light_3d.png}  
    & \includegraphics[width=0.13\textwidth]{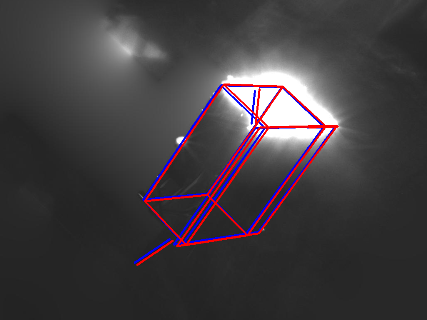} \\
\end{tabular}
}

% \vspace{-5pt}
\captionof{figure}{Qualitative results on target-domain samples. Row-1: GT keypoints (\textcolor{red}{red}). Row-2: heatmaps from synthetic pretrained model. Row-3: refined heatmaps after \acronym adaptation. Row-4: recovered Object pose with \textcolor{red}{GT} and \textcolor{blue}{prediction} overlaid.}
\label{fig:comparison}
\vspace{-1em}

%% file: sections/0.Abstarct.tex
\begin{abstract}
Unsupervised domain adaptation (UDA) enables robust transfer of knowledge from simulated to real environments while exploiting a subset of unlabeled target data to improve real-world performance. Existing UDA methods for Object pose estimation often rely on global feature matching, multi-stage larger frameworks, or image translation pipelines, which tend to overlook the pose-specific information embedded in feature representations. To bridge this limitation, we introduce~\acronym~that targets the adaptation of pose-sensitive features in localized regions, ensuring that domain alignment preserves the geometric cues essential for accurate pose estimation. \acronym~achieves UDA with three key components: (1)~Efficient Cross-Domain Pairing strategy leveraging intermediate features to identify pose similar image pairs across domains without supervision; (2) Contrastive Alignment to perform feature alignment at localised regions in both intermediate and task-specific representations; and (3) Consistency-Based Pseudo-Label Refinement to improve reliability by encouraging stable target predictions. Extensive experiments demonstrate that \acronym~achieves state-of-the-art performance across multiple well-known object pose estimation benchmarks featuring diverse and challenging scenarios.
\end{abstract}

%% file: sections/1.Introduction.tex
\section{INTRODUCTION}
%NEW ONE
% Deep learning (DL) models trained exclusively on synthetic data often underperform on real data, primarily due to the inherent distributional discrepancies between synthetic and real domains, a phenomenon commonly referred to as the \textit{domain gap}~\cite{denninger2020blenderproc} or \textit{sim-to-real} transfer \cite{pitkevich2024survey}. However, obtaining such ground-truth annotations in real-world scenarios is often impractical due to the associated costs, labor, and scalability challenges. To mitigate domain gap challenges, domain adaptation methods, particularly Unsupervised Domain Adaptation (UDA)~\cite{farahani2021} were adapted to leverage any available unlabeled real-world data (i.e., target domain) along with abundant labeled synthetic data (i.e., source domain) to train DL models. UDA has shown success in classification and segmentation~\cite{schwonberg2023survey}; however, regression tasks such as object pose estimation present unique challenges due to the continuous nature of the output space, ambiguities of object symmetries, and multimodal uncertainties~\cite{ikeda2024diffusionnocs}. Moreover, the domain gap can include not only appearance variations, but also geometric and structural differences~\cite{bauer2024challenges}, affecting both feature extraction and output prediction.  

Deep learning (DL) models trained solely on synthetic data often underperform on real data due to distributional discrepancies between synthetic and real domains, commonly referred to as the \textit{domain gap}~\cite{denninger2020blenderproc} or \textit{sim-to-real} transfer \cite{pitkevich2024survey}. Yet collecting ground-truth annotations in real-world settings is often impractical given the cost, labor, and scalability constraints. To mitigate these gaps, Unsupervised Domain Adaptation (UDA)~\cite{farahani2021} leverages unlabeled real data (target) alongside labeled synthetic data (source) to train DL models. While UDA has been effective for classification and segmentation~\cite{schwonberg2023survey}, regression tasks such as object pose estimation remain challenging due to continuous outputs, object symmetries, and multimodal uncertainty~\cite{ikeda2024diffusionnocs}. Moreover, the domain gap can span not only appearance changes but also geometric and structural differences~\cite{bauer2024challenges}, affecting both feature extraction and output prediction.

Recent efforts have increasingly focused on self-supervised adaptation, often via heavy pipelines with anchor/cluster partitioning and multi-stage optimization that are sensitive to hyperparameter and can struggle with continuous, multimodal pose distributions \cite{zhang2023manifold}. In contrast, image-to-image translation methods \cite{chen2023texpose, wang2023bridging}  focus on narrowing the appearance gap by rendering synthetic data more realistic; however, they cannot fully preserve the geometric or structural information essential for pose accuracy, especially under large domain gaps, often resulting in noisy target pseudo labels. Multi-view consistency approaches \cite{tan2023smoc, tan2025onda} can help, but rely on assumptions that rarely hold in standard single-view scenarios. An alternative, still relatively underexplored in pose estimation, is direct feature-level alignment, which has shown strong generalization in other tasks \cite{zhang2019category, chen2021towards}.  Extending this concept to pose estimation introduces additional challenges, as indiscriminate feature alignment can disrupt task-specific spatial and geometric relationships essential for accurate prediction, rather than enhancing them. To overcome these limitations, a practical adaptation strategy must ensure that (1) features are aligned at a fine-grained spatial scale that preserves geometric consistency, and (2) pseudo labels in the target domain remain reliable despite the domain-induced appearance and structural distortions.

% This paper introduces \acronym, a unified keypoint-based UDA framework for object pose estimation to address the synthetic-to-real domain gap through \textbf{C}ontrastive \textbf{A}lignment and \textbf{P}seudo-\textbf{L}abel \textbf{R}efinement, with the following key contributions: 
% \textbf{First}, a cross-domain pairing strategy that identifies source-target pairs with consistent poses, enabling effective supervision in the absence of target annotations.
% \textbf{Second}, a patch-level contrastive alignment mechanism to jointly align feature embeddings and task-specific representations.
% \textbf{Third}, a consistency-based pseudo-label refinement mechanism that leverages augmented views to improve and stabilize target predictions. \textbf{Finally}, the framework’s effectiveness is validated on multiple Object pose estimation benchmarks with varied domain gaps and complexities, and model sizes, complemented by an ablation study showing each component's role and contribution.

This paper introduces \acronym, a unified keypoint-based UDA framework for object pose estimation to address the synthetic-to-real domain gap via \textbf{C}ontrastive \textbf{A}lignment and \textbf{P}seudo-\textbf{L}abel \textbf{R}efinement. wiith the following key contributions:
\textbf{First}, a cross-domain pairing strategy that identifies source–target pairs with consistent poses, enabling supervision without target annotations.
\textbf{Second}, a patch-level contrastive alignment mechanism that jointly aligns feature embeddings and task-specific representations.
\textbf{Third}, a consistency-based pseudo-label refinement scheme that uses augmented views to improve and stabilize target predictions.
\textbf{Finally}, the framework’s effectiveness is validated on multiple object pose estimation benchmarks spanning diverse domain gaps, complexities, and model sizes, supported by an ablation study isolating each component’s impact.

% The remainder of this paper is organized as follows. Section~\ref{sec:related} reviews the relevant literature on UDA and Object pose estimation. Section~\ref{sec:solution} introduces the proposed \acronym~framework, detailing its design and implementation. Section~\ref{sec:results} presents the experimental setup and quantitative results on benchmark datasets. Finally, Section~\ref{sec:conclusion} summarizes the key findings and outlines future directions.  

The remainder of this paper is organized as follows: Section~\ref{sec:related} reviews related work on UDA and object pose estimation. Section~\ref{sec:solution} presents the proposed \acronym~framework and its implementation. Section~\ref{sec:results} describes the experimental setup and reports results on benchmark datasets. Section~\ref{sec:conclusion} concludes with key findings and future work.

%% file: sections/2.RelatedWorks.tex
\section{RELATED WORKS}
\label{sec:related}

% For DL-based Object pose estimation of known objects from an image, two widely known approaches are \textit{Direct regression} and \textit{Hybrid approaches}. Direct approaches aim to learn an end-to-end mapping by directly regressing the continuous pose values from an image (e.g., \cite{xiang2018posecnn}). These methods often struggle with the inherent non-linearity of the rotation space, leading to complex loss function design, lower accuracy, and limited interpretability. Hybrid methods (e.g., \cite{yang2024pvspe}) predict the 2D image coordinates of predefined keypoints via a regressor; subsequently, given the known 3D coordinates of these keypoints in the object's reference frame, and the camera intrinsic parameters, Perspective-n-Point (PnP) algorithms are employed to recover the object's pose. Owing to their better geometric interpretability and flexibility in incorporating spatial priors, hybrid approaches have become the dominant paradigm for pose estimation.

For DL-based object pose estimation of known objects from an image, two common paradigms are \textit{Direct regression} and \textit{Hybrid approaches}. Direct methods learn an end-to-end mapping by regressing continuous pose parameters from the image (e.g., \cite{xiang2018posecnn}), but they often struggle with the nonlinearity of rotation space, leading to complex loss design, lower accuracy, and limited interpretability. In contrast, hybrid methods (e.g., \cite{yang2024pvspe}) first predict 2D locations of predefined keypoints and then recover the pose using Perspective-n-Point (PnP), given the corresponding 3D keypoints and camera intrinsics. Owing to their geometric interpretability and flexibility in incorporating spatial priors, hybrid approaches have become the prevailing paradigm for pose estimation.

\subsection{UDA for Keypoint Detection}

UDA aims to transfer knowledge from a labeled source domain to an unlabeled target domain by addressing the distributional shifts in both features and labels. Existing methods are broadly grouped into \textit{Reconstruction-}, \textit{Adversarial-} and \textit{Discrepancy-}based approaches; for a detailed review see~\cite{farahani2021}. UDA for keypoint detection has unique challenges, as keypoints encode precise geometric relationships and are therefore highly sensitive to domain shift \cite{jiang2021regressive}. Shape-consistent frameworks \cite{vasconcelos2021shape} integrate adversarial learning and self-supervision to preserve geometric structure, but they struggle as the domain gap increases and their reliance on implicit shape priors may not generalize to complex 3D objects. In~\cite{jiang2021regressive} a dual-regressor strategy inspired by Disparity Discrepancy Theory \cite{zhang2019bridging} was introduced to stabilize adversarial training by reducing source–target domain conflicts; nevertheless, adversarial methods remain sensitive to hyperparameters and can be unstable. Moreover, their indirect discriminator-driven alignment may match irrelevant superficial characteristics rather than discriminatory task features \cite{zhao2019learning}, which can limit robust transfer.

Among discrepancy-based methods, contrastive learning emerged as an effective paradigm for classification tasks, where discrete labels naturally define positive/negative pairs \cite{xu2024unsupervised}, but constructing such pairs is non-trivial for regression due to continuous outputs \cite{keramati2024conr}. \acronym overcomes this through a cross-domain pairing strategy that establishes meaningful correspondences, enabling effective contrastive alignment for keypoint regression under domain shift.

\subsection{UDA for Object Pose Estimation}

Recent UDA methods for object pose estimation primarily reduce the domain gap through self-supervised adaptation. Typically, models are trained on synthetic data and then adapted using unlabeled real images. For example, Self6D++ \cite{wang2021gdr} performs self-supervised mask alignment with differentiable rendering, but it still faces a rendered-vs-real discrepancy and depends on a costly rendering pipeline that limits scalability to complex scenes. SMOC-Net~\cite{tan2023smoc} introduces a relative-pose geometry constraint using estimated camera poses. MAST \cite{zhang2023manifold} uses manifold-aware self-training that models dependencies among regression targets to encourage domain-invariant representations, though its smooth-transition assumption may miss local structural misalignments across domains. Other works attempt to narrow the gap via synthetic-to-real image translation \cite{chen2023texpose}, which can reduce appearance differences but often generalizes poorly to challenging cases such as occlusion. More recently, \cite{tan2025onda} improves translation robustness by leveraging multi-view images with overlapping regions, an assumption that many real-world datasets do not satisfy. In contrast, direct feature alignment has demonstrated strong performance in other UDA tasks \cite{zhang2019category, chen2021towards}, but remains relatively unexplored for object pose estimation, A key difficulty is that alignment must target task-relevant features, otherwise, it risks amplifying superficial cues (e.g., texture/style) while overlooking pose-critical information.

Motivated by these limitations, our work introduces a patch-based dual-level alignment strategy that jointly aligns backbone embeddings and regression head outputs, ensuring that adaptation captures both marginal and conditional distributions relevant to pose, building on prior findings that joint alignment of features and outputs improves adaptation~\cite{9991175}, as well as theoretical results emphasizing the importance of combining marginal and conditional alignment for regression problems~\cite{yang2024cod}. Furthermore, to mitigate the unreliability of predictions in the unlabeled target domain, we incorporate a consistency-based pseudo-label refinement that filters low-confidence or spatially inconsistent labels, thereby stabilizing supervision. Together, these components form a unified framework named \acronym~that advances UDA for Object pose estimation beyond the need for image translation or computationally expensive rendering pipelines.

% Motivated by these limitations, we propose a patch-based dual-level alignment strategy that jointly aligns both backbone embeddings and regression head outputs, ensuring that adaptation captures both marginal and conditional distributions relevant to pose, in line with prior evidence that joint feature–output alignment improves adaptation~\cite{9991175} and theory highlighting the need to combine marginal and conditional alignment for regression problems~\cite{yang2024cod}. Furthermore, to mitigate the unreliability of predictions in the unlabeled target domain, we introduce a consistency-based pseudo-label refinement that filters low-confidence or spatially inconsistent labels, stabilizing supervision. Together, these components form \acronym, advancing UDA for object pose estimation beyond relying on image translation or expensive rendering pipelines.

% beyond reliance on image translation or rendering-heavy pipelines.

%% file: sections/fig2.tex
\begin{figure*}[t]
    \centering
    % Adjust this 0.95 to scale the entire assembly
    \resizebox{0.95\textwidth}{!}{% 
    \begin{tabular}{@{}c@{}}
        % Top Row
        \includegraphics[width=\textwidth]{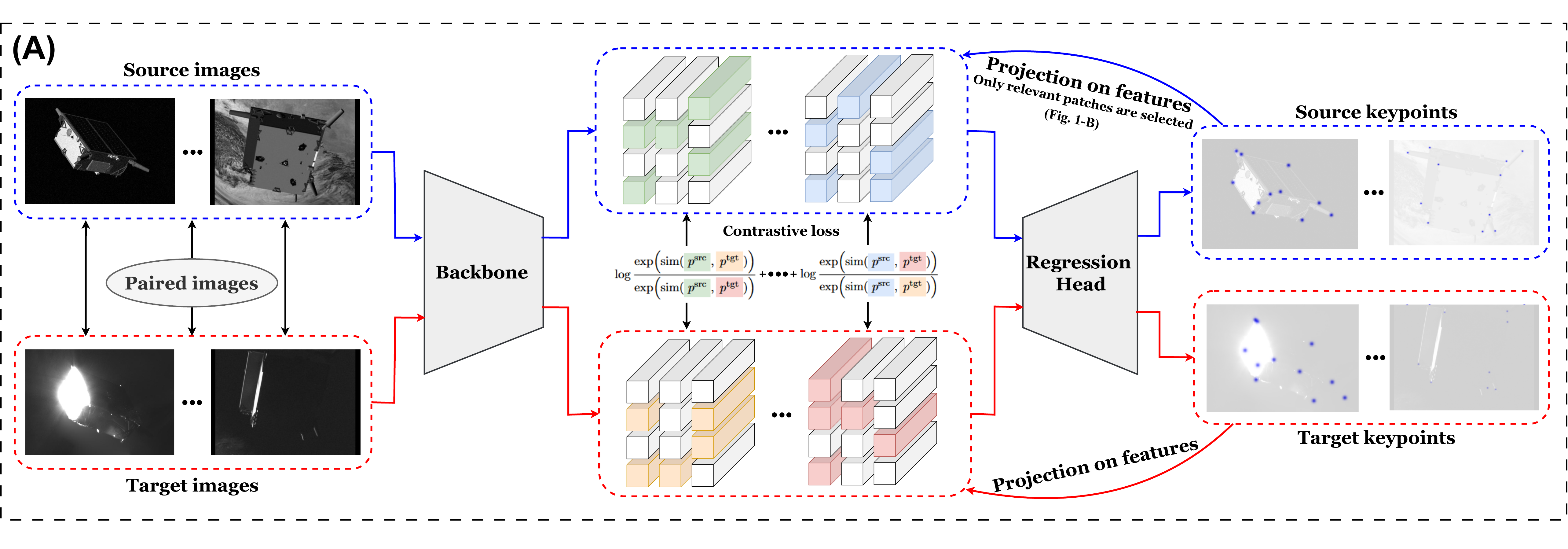} \\ [-1em]       
        % Bottom Row
        \begin{minipage}{0.325\textwidth}
            \includegraphics[width=\linewidth]{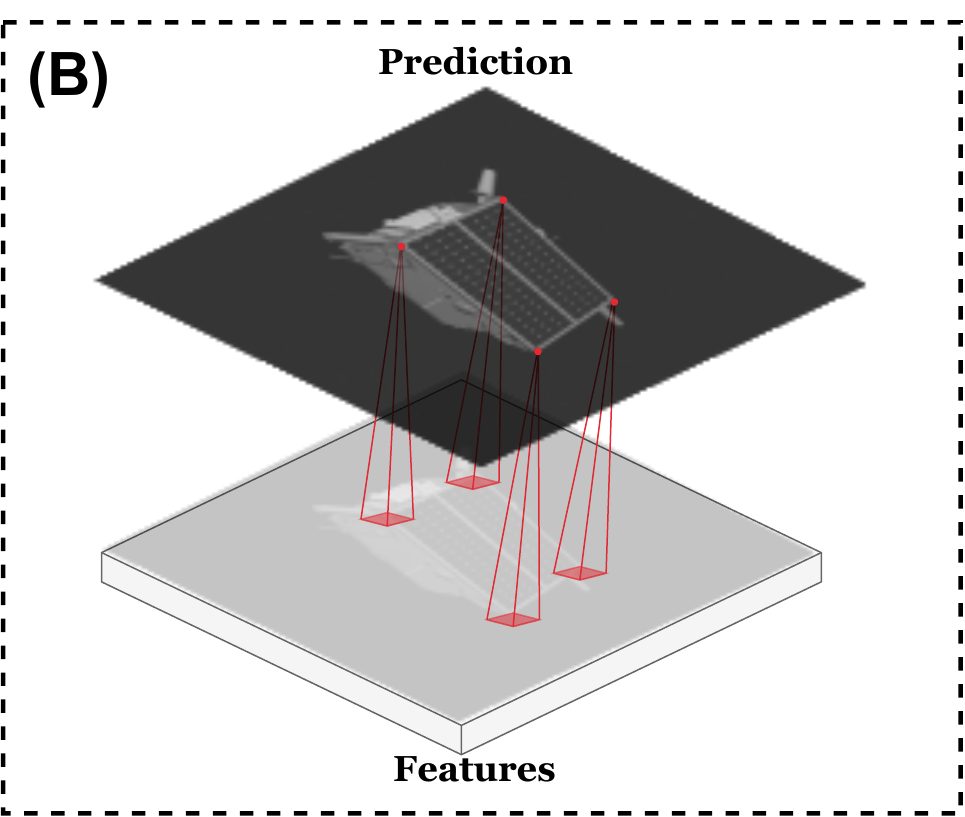}
        \end{minipage}
        \hfill
        \begin{minipage}{0.6643\textwidth}
            \includegraphics[width=\linewidth]{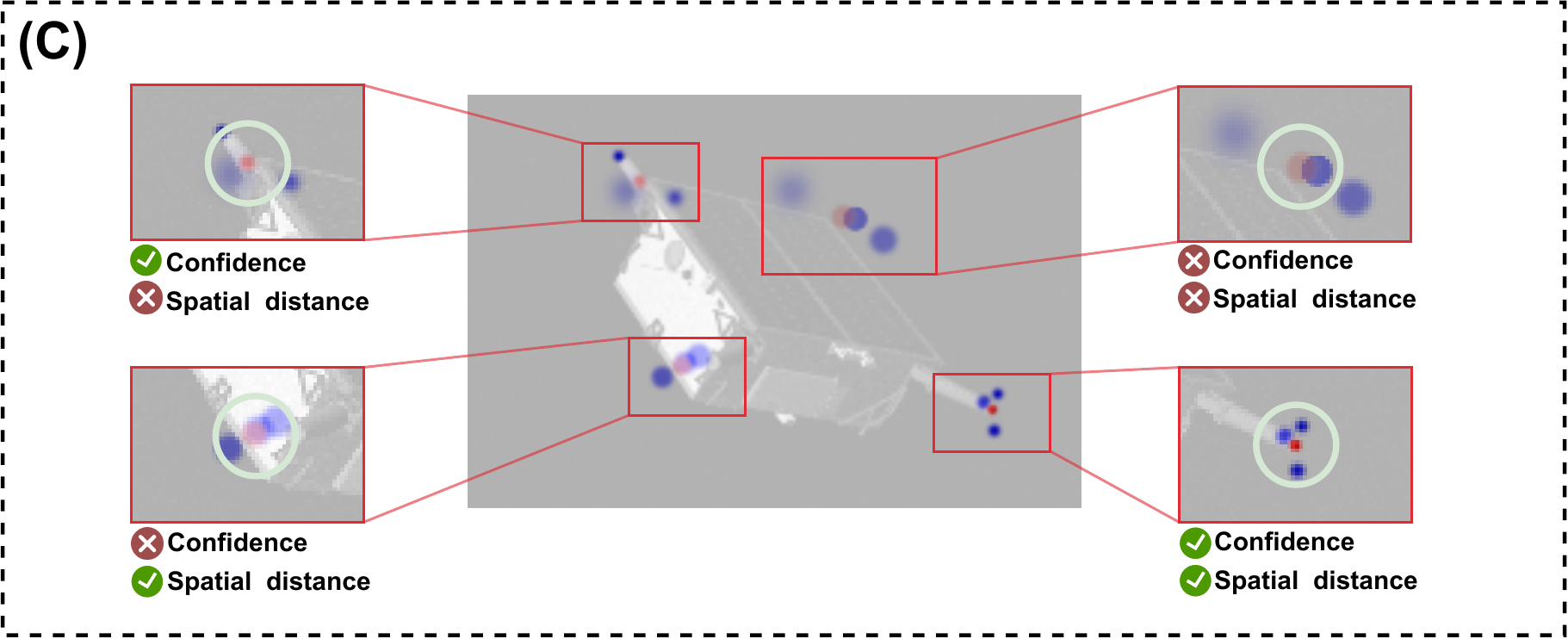}
        \end{minipage}
    \end{tabular}%
    }
    \vspace{-0.1cm}
    \caption{Overview of the proposed framework. 
(A) Contrastive alignment between relevant feature patches (in colors) of paired samples. 
(B) Keypoint projection for local patch extraction from the feature map. 
(C) Consistency-based filtering of pseudo-keypoints (red), incorporating confidence thresholding and spatial consistency checks (green circle).}
    \label{fig:main}
    \vspace{-0.55cm}
\end{figure*}

%% file: sections/3.Methodology.tex
\section{METHODOLOGY}
\label{sec:solution}
\input{sections/3.1.Preliminaries}

\input{sections/3.2.CAPLR}

%% file: sections/3.1.Preliminaries.tex
\subsection{Problem Statement}
Let $(x_i, y_i) \sim \mathcal{D}$ be a pair sampled from a data distribution $\mathcal{D}$ over the input space $\mathcal{X}$ and the output space $\mathcal{Y}$, where $x_i \in \mathbb{R}^{H \times W \times 3}$ is the input image and $y_i = \{(u_k, v_k)\}_{k=1}^{K}$ is a set of annotated 2D keypoints $K$. 
The objective is to train a regressor $f_\theta: \mathcal{X} \rightarrow \mathcal{Y}$ that, given an input image, predicts 2D keypoints by minimizing the prediction error, where $\mathcal{L}$ denotes a regression loss function measuring the discrepancy between the predicted and ground truth keypoints.
\begin{equation}
\mathcal{L}_{\text{task}}(\theta) = \mathbb{E}_{(x, y) \sim \mathcal{D}} \ \mathcal{L}(f_\theta(x), y).
\label{eq:task_loss}
\end{equation}
Given the predicted 2D keypoints $\{(\Tilde{u}_k, \Tilde{v}_k)\}_{k=1}^K$,  the known 3D object keypoints $\{X_k\}_{k=1}^K$, and camera intrinsics $\phi$, the optimal 6D object pose $(R^*, t^*)$ is estimated by solving a PnP problem:
{\setlength{\abovedisplayskip}{0pt}
 \setlength{\belowdisplayskip}{0pt}
\begin{equation}
(R^*, t^*) = \arg\min_{R, t} \sum_{k=1}^{K} \left\| \phi(R X_k + t) - (\Tilde{u}_k, \Tilde{v}_k) \right\|^2.
\label{eq:pnp}
\end{equation}}In the UDA setting, a labeled source domain $\hat{\mathcal{P}} = \{(x_i^s, y_i^s)\}_{i=1}^{n}$ is given, sampled from distribution $\mathcal{P}(x^s, y^s)$ along with an unlabeled target domain $\hat{\mathcal{Q}} = \{x_i^t\}_{i=1}^{m}$ sampled from the marginal $\mathcal{Q}(x^t)$ of the joint target distribution $\mathcal{Q}(x^t,y^t)$ where the labels $y^t$ are not available during training. The objective is to learn a model $f_\theta$ that minimizes the expected prediction error in the target domain:
 \begin{equation}
\text{err}_{\mathcal{Q}} = \mathbb{E}_{(x, y) \sim \mathcal{Q}} \left[ \mathcal{L}(f_\theta(x), y) \right].
\label{eq:uda_error}
\end{equation} However, this task is challenging due to the domain gap, which arises from discrepancies in both input and output distributions between the source and target domains \cite{yang2024cod}:
 \begin{equation}
  \begin{array}{r@{\;\neq\;}l}
    \mathcal{P}(x^s)          & \mathcal{Q}(x^t)\\
    \mathcal{P}(y^s\mid x^s)  & \mathcal{Q}(y^t\mid x^t).
  \end{array}
  \label{eq:domain_gap}
\end{equation}This mismatch hinders generalization capability of models trained on source domain when deployed in target domain.

%% file: sections/3.2.CAPLR.tex
\subsection{\acronym~Framework}
To address the gap described in \cref{eq:domain_gap}, a domain adaptation strategy would aim to align both the input and conditional output distributions \cite{yang2024cod} as follows:
\begin{equation}
  \begin{array}{r@{\;\equiv\;}l@{\;}r}
    \mathcal{P}(x^s)           & \mathcal{Q}(x^t)            & \\
    \mathcal{P}(y^s\mid x^s)    & \mathcal{Q}(y^t\mid x^t).   
  \end{array}
  \label{eq:solution_domain_gap}
\end{equation}
However, aligning the input distributions $\mathcal{P}(x^s)$ and $\mathcal{Q}(x^t)$ is nontrivial due to domain shifts in illumination, texture, and rendering fidelity between source and target images. Similarly, aligning the conditional distributions $\mathcal{P}(y^s \mid x^s)$ and $\mathcal{Q}(y^t \mid x^t)$ is particularly difficult in regression due to the continuous nature of the output space. To overcome this, we reformulate the objective in \cref{eq:solution_domain_gap} within the feature space, where task-relevant representations are more amenable to alignment. Specifically, \cref{eq:solrelaxed_domain_gap} defines alignment over backbone features $g(x)$ and regression head outputs $h(g(x))$, which serve as proxies for the input and conditional output distributions, respectively.
\begin{equation}
  \begin{array}{r@{\;\equiv\;}l}
    \mathcal{P}_g\bigl(g(x^s)\bigr)          & \mathcal{Q}_g\bigl(g(x^t)\bigr)\\
    \mathcal{P}_h\bigl(h(g(x^s))\bigr)  & \mathcal{Q}_h\bigl(h(g(x^t))\bigr),
  \end{array}
  \label{eq:solrelaxed_domain_gap}
\end{equation} where $\mathcal{P}_g$ and $\mathcal{P}_h$ refer to source distributions, and $\mathcal{Q}_g$ and $\mathcal{Q}_h$ to target distributions; subscripts $g$ and $h$ indicate backbone and head features, respectively. Building on this reformulation, the proposed \acronym~framework instantiates the alignment process through three complementary stages: cross-domain pairing, local patch contrastive alignment, and consistency-based pseudo-label refinement. Together, these stages constitute a practical instantiation of the objective in \cref{eq:solrelaxed_domain_gap}. The details of each stage are presented in \cref{cdp}–\ref{cbpr}.

\input{sections/3.2.1.CDP}
\input{sections/3.2.2.CA}
\input{sections/4.1.Table_LM}
\input{sections/3.2.3.PLR}

%% file: sections/3.2.1.CDP.tex
\subsubsection{Cross-Domain Pairing}\label{cdp}
Effective feature alignment requires establishing source–target pairings with similar poses. In the UDA setting, this is particularly challenging due to the absence of target labels and the limited reliability of initial pose estimates. To address this, a two-step Cross-Domain Pairing (CDP) strategy is introduced, comprising (i) feature-based candidate selection and (ii) pose-based refinement.

\textit{\textbf{1.A. Feature-Based Initial Candidate Selection:}}  
\label{p:FBICS} Pose predictions obtained from a model trained exclusively on the source domain often lead to suboptimal pairings when used directly, due to errors caused by domain shift. To mitigate this limitation, the first step of CDP relies on feature embeddings rather than raw pose estimates. Although accurate keypoint localization in the target domain is unreliable, heatmap activations still provide informative cues about the approximate object region (see \cref{fig:comparison}-row2). These activations are therefore used to extract local features by treating predicted keypoints as anchors for cropping patches from the backbone feature map $F \in \mathbb{R}^{C \times H \times W}$.

Formally, for each image $I$, the model outputs $K$ keypoints  $\{k_j = (x_j, y_j)\}_{j=1}^{K}$, each with a confidence score $c_j \in [0,1]$ derived from the predicted heatmap. Around each keypoint $k_j$, a local patch
$p_j \in \mathbb{R}^{C \times h' \times w'}$ is extracted from $F$  as shown in \cref{fig:main}-B. These patches are aggregated into a confidence-weighted image-level embedding:
{\setlength{\abovedisplayskip}{5pt}
\setlength{\belowdisplayskip}{5pt}
\begin{equation}
E = \frac{\sum_{j=1}^{K} c_j\, \mathrm{vec}(p_j)}{\sum_{j=1}^{K} c_j + \epsilon},
\label{eq:image_embedding}
\end{equation}}
where $\mathrm{vec}(\cdot)$ denotes vectorization of the patch. Dividing by the sum of confidences normalizes the embedding ensuring that images with varying numbers of confident keypoints are represented on a similar scale. Cosine similarity is then used to cluster the resulting embeddings, forming an initial pool of structurally similar source–target image pairs.

\textit{\textbf{1.B. Refinement via Pose Prediction:}} The second stage refines the candidate set obtained from the previous step by applying pose-based filtering. While direct pairwise comparison of predicted poses across all source–target images is computationally expensive and error-prone due to noisy target predictions, restricting the operation to the reduced set from feature-based selection makes it feasible and more reliable.This refinement mitigates the risk of retaining structurally similar pairs that differ in pose, which would otherwise degrade feature alignment. For each candidate pair $(I_s, I_t)$ where $I_s$ and $I_t$ denote a source and target image respectively, the similarity between predicted poses $(t_s, q_s)$ and $(t_t, q_t)$ is evaluated using a combined pose error ($E_{\mathrm{pose}}$) defined as:
\begin{equation}
  \left.
  \begin{array}{lll}
    E_T              & = & \|t_t - t_s\| \\[0.5em]
    E_T^{\mathrm{norm}}  & = & E_T/\|t_s\| \\[0.5em]
    E_R               & = & 2\arccos\!\left(|\langle q_t, q_s \rangle|\right) \\[0.5em]
    E_{\mathrm{pose}}    & = & E_T^{\mathrm{norm}} + \lambda_{\mathrm{rot}} \cdot E_R
  \end{array}
  \right\}
  \label{eq:pose_error}
\end{equation}

 where $E_T$ denotes the translation error, $E_T^{\mathrm{norm}}$ is the normalized translation error, which accounts for scale differences, and  $E_R$ is the angular distance between quaternions.  $\lambda_{\mathrm{rot}}$ is a weighting factor balancing the two. The source image with the lowest $E_{\mathrm{pose}}$ is selected as the optimal match for each target image, ensuring that only pairs with closely aligned pose are retained for the subsequent %alignment
stage.

%% file: sections/3.2.2.CA.tex
\subsubsection{Local Patch Contrastive Alignment}\label{lpca}
Keypoint predictions in the target domain are often imprecise due to domain shift, producing dispersed heatmaps, as shown in the second row of \cref{fig:comparison}. Nevertheless, as established in~\cref{p:FBICS}, these heatmaps frequently localize near correct object regions, providing a meaningful signal for patch-level alignment. Local Patch Contrastive Alignment (LPCA) leverages this property by aligning local feature representations across source–target pairs, in contrast to global feature alignment that may capture irrelevant background context (\cref{fig:main}-A). Specifically, feature patches are extracted around predicted keypoints (\cref{fig:main}-B) from both intermediate backbone features and pre-output regression head features. Backbone-level patches preserve spatial structure, while head-level patches encode task-specific semantics related to keypoint prediction. For a source–target pair $(I_s, I_t)$, keypoints are predicted as $\{k_j^s\}_{j=1}^{K} = f(I_s)$ and $\{k_j^t\}_{j=1}^{K} = f(I_t)$ using a network trained on synthetic data. Around each $k_j$, patches $p_j^s$ and $p_j^t$ are cropped from the feature maps. These localized descriptors are then aligned via the InfoNCE loss $L_{\text{iNCE}}$~\cite{oord2018representation}:
\begin{equation}
L_{\text{iNCE}} = -\frac{1}{BK} \sum_{b,k} \log \frac{\exp(\text{sim}(p_{b,k}^{\text{src}}, p_{b,k}^{\text{tgt}})/\tau)}{\sum_{b'=1}^{B} \exp(\text{sim}(p_{b,k}^{\text{src}}, p_{b',k}^{\text{tgt}})/\tau)}
\end{equation}
where $\text{sim}(\cdot)$ denotes cosine similarity, $\tau$ is a temperature hyperparameter, and $B$ is the batch size. The final alignment loss ($L_{\mathrm{align}}$) combines the task objective with contrastive losses at both levels:
{
 \begin{equation}
L_{\mathrm{align}} = L_{\mathrm{task}} + \lambda_{\mathrm{backbone}} L_{\mathrm{iNCE}}^{\mathrm{backbone}} + \lambda_{\mathrm{head}} L_{\mathrm{iNCE}}^{\mathrm{head}},
\end{equation}}
where $\lambda_{\mathrm{backbone}}$ and $\lambda_{\mathrm{head}}$ balance the alignment and the keypoint regression objective $L_{\mathrm{task}}$ (\cref{eq:task_loss}). These weights are chosen to ensure that the contrastive terms have similar magnitude to the task loss at initialization, with greater emphasis placed on backbone-level alignment due to its broader effect on domain invariance.

%% file: sections/4.1.Table_LM.tex
\begin{table*}[!b]
  \centering
  \caption{Comparison on LineMOD. The best and the second best results are in \textbf{bold} and \underline{underlined}. $S$: synthetic data; $R$: annotated real data; $R^{-}$: unannotated real data. Symmetric objects are in \textit{italic}. Metrics: ADD-(S) [\%] $\uparrow$.}
  \label{tab:linemod-results}
  \resizebox{\textwidth}{!}{
  \begin{tabular}{l|c|ccccccccccccc|c}
    \toprule
    Methods & Data & Ape & Bvise. & Cam & Can & Cat & Drill & Duck & \textit{Eggbox} & \textit{Glue} & Holep. & Iron & Lamp & Phone & \textbf{Mean} $\uparrow$ \\
    \midrule
    GDR-Net \cite{wang2021gdr} & \multirow{3}{*}{$S$} & 50.9 & 99.4 & 89.2 & 97.2 & 79.9 & 98.7 & 24.6 & 81.1 & 81.2 & 41.9 & 98.8 & 98.9 & 64.3 & 77.4 \\
    DPODv2 \cite{shugurov2021dpodv2} & & 62.1 & 88.3 & 92.5 & 96.6 & 86.1 & 90.1 & 54.8 & 98.6 & 95.4 & 27.0 & 98.2 & 91.0 & 74.3 & 81.2 \\
    MAR \cite{zhang2023manifold} & & 68.6 & 97.4 & 79.4 & 98.3 & 87.1 & 94.2 & 61.3 & 82.0 & 87.1 & 56.7 & 94.3 & 92.3 & 68.8 & 82.1 \\
    \midrule
    GDR-Net \cite{wang2021gdr} & \multirow{2}{*}{$R$} & 85.0 & \underline{99.8} & 96.5 & 99.3 & 93.0 & \textbf{100.0} & 65.3 & 99.9 & 98.1 & 73.4 & 86.9 & \textbf{99.6} & 86.3 & 91.0 \\
    DPODv2 \cite{shugurov2021dpodv2} & & 80.0 & 99.7 & \textbf{99.2} & 99.6 & 95.1 & 98.9 & 79.5 & 99.6 & \textbf{99.8} & 72.3 & 99.4 & 96.3 & \textbf{96.8} & \underline{93.5} \\
    \midrule
    Self6D++ \cite{wang2021occlusion} & \multirow{7}{*}{$S+R^{-}$} & 76.0 & 91.6 & 97.1 & \underline{99.8} & 85.6 & 98.8 & 56.5 & 91.0 & 92.2 & 35.4 & 99.5 & 97.4 & 91.8 & 85.6 \\
    MAST \cite{zhang2023manifold} & & 73.5 & 97.2 & 80.8 & 98.6 & 89.1 & 93.9 & 66.9 & 95.3 & 95.4 & 69.8 & 95.5 & \underline{98.9} & 93.4 & 93.2 \\
    SMOC-Net \cite{tan2023smoc} & & \underline{85.6} & 96.7 & 97.2 & \textbf{99.9} & 95.0 & \textbf{100.0} & 76.0 & 98.3 & 99.2 & 45.6 & \textbf{99.9} & \underline{98.9} & 94.0 & 91.3 \\
    Tex-Pose \cite{chen2023texpose} & & 80.9 & 99.0 & 94.8 & 99.7 & 92.6 & 97.4 & 83.4 & 94.9 & 93.4 & 79.3 & 99.8 & 98.3 & 78.9 & 91.7 \\
    ONDA-Pose \cite{tan2025onda} & & 83.0 & \textbf{99.9} & \underline{98.0} & 99.5 & 96.4 & 99.4 & 76.5 & 96.3 & 87.1 & 83.4 & 99.6 & \underline{98.9} & 93.4 & 93.2 \\
    \rowcolor[gray]{0.93}  \acronymS (Ours) & & 82.1 & 95.2 & 93.2 & 97.4 & \underline{98.8} & 97.0 & \underline{89.5} & \textbf{100.0} & 96.9 & \underline{95.1} & 98.8 & 78.0 & 93.7 & \underline{93.5} \\
    \rowcolor[gray]{0.93}  \acronymM (Ours) & & \textbf{90.7} & 99.5 & 93.2 & 97.9 & \textbf{99.3} & 97.7 & \textbf{98.8} & \textbf{100.0} & \underline{99.5} & \textbf{97.0} & \textbf{99.9} & 94.1 & \underline{96.6} & \textbf{97.2} \\
    \bottomrule
  \end{tabular}
  } \vspace{-0.1cm}
\end{table*}

%% file: sections/3.2.3.PLR.tex
\subsubsection{Consistency-Based Pseudo Label Refinement}\label{cbpr}
To improve robustness on the unlabeled target domain, a Consistency-Based Pseudo Label Refinement (CBPR)  strategy  is applied after the LPCA stage, leveraging model confidence to identify reliable keypoint predictions and enforce consistency across multiple augmented variants of the same image. Given a target image $I_t$ a set of $M$ augmented variants $\{I^{(m)}_t\}_{m=1}^{M}$ is generated using the same augmentations applied to source-domain images during training. Each variant is passed through the current model $f$ producing keypoint predictions $K^{(m)} \in \mathbb{R}^{K \times 2}$ with associated confidence scores $c^{(m)} \in \mathbb{R}^K$:
\begin{equation}
K^{(m)},\, c^{(m)} = f\Bigl(I^{(m)}_t\Bigr), \quad m \in \{1,\dots,M\}.
 \end{equation}
To construct pseudo-labels, predictions across these variants are aggregated via confidence-weighted averaging. For each keypoint $k_k$, the pseudo-label is computed as follows:
 {
 \begin{equation}
k_{\mathrm{pseudo}} = \frac{\sum_{m=1}^{M} c_{k}^{(m)}\, k_{k}^{(m)}}{\sum_{m=1}^{M} c_{k}^{(m)} + \epsilon},
 \end{equation}}
where $\epsilon$ ensures numerical stability. This aggregation prioritizes high-confidence predictions while down-weighting uncertain ones.
Two filtering criteria are applied to improve reliability (\cref{fig:main}-C):
\begingroup
  \setlength\topsep{0pt}    % above & below
  \setlength\partopsep{0pt} % paragraph separation
  \setlength\itemsep{0pt}   % between items
  \setlength\parsep{0pt}  
\begin{enumerate}
    \item[3.A.] \textit{\textbf{Confidence Thresholding:}} A pseudo-keypoint is retained only if at least one variant yields a confidence score above a threshold $c_{\mathrm{thresh}}$:
     { \begin{equation}
    \exists m \in \{1,\dots,M\}: \quad c_k^{(m)} > c_{\mathrm{thresh}}.
    \end{equation}}
    \item[3.B. ] \textit{\textbf{Spatial Consistency Filtering:}} Retained keypoints must also exhibit low spatial variance across variants. Specifically, the maximum distance between any prediction and the pseudo-label must be below $d_{\mathrm{thresh}}$:
     {
 \begin{equation}
    \max_{m \in \{1,\dots,M\}} \| k_k^{(m)} - k_{\mathrm{pseudo}} \|_2 < d_{\mathrm{thresh}}.
    \end{equation}}
\end{enumerate}
\endgroup
Only keypoints that satisfy both criteria are used as pseudo-labels and further used in a consistency loss to regularize the model. For each training image $b$ in batch $B$, with $K_b$ valid keypoints, the loss encourages predictions across all augmented variants to remain close to the pseudo-label:
{\setlength{\abovedisplayskip}{2pt}
 \begin{equation}
L_{\mathrm{pseudo}} = \frac{1}{B} \sum_{b=1}^{B} \frac{1}{K_b} \sum_{k \in K_b} \sum_{m=1}^{M} \| k_{b,k}^{(m)} - k_{b,k}^{\mathrm{pseudo}} \|_2^2.
 \end{equation}}
This loss reinforces prediction stability under diverse augmentations, effectively regularizing the model and enhancing robustness to noisy keypoints in the unlabeled target domain.

%% file: sections/4.Experiments.tex
\input{sections/4.2.Table_O-LM}

\input{sections/4.3.Table_speedplus}

\section{EXPERIMENTS AND RESULTS}
\label{sec:results}
\subsection{Experimental Setup} 
\label{ss:experiments}
\subsubsection{Datasets} \acronym~is evaluated on different benchmarks for object pose estimation. \textbf{LineMOD} dataset \cite{hinterstoisser2012model} contains individual sequences of 13 household objects in cluttered scenes with challenging lighting with $\sim$1.2k real images per sequence. Notably, two of the objects (Eggbox and Glue) exhibit symmetry. Following prior works, we use 15\% of images for domain adaptation (without annotations) and the rest for testing. \textbf{Occluded-LineMOD} \cite{brachmann2014learning}, a subset of LineMOD with 8 objects under severe occlusions,  we use
the BOP split \cite{hodavn2019photorealistic} to sample data for testing and use the rest for adaptation. \textbf{HomebrewedDB} \cite{kaskman2019homebreweddb} provides new set of images of three LineMOD objects (bvise, driller, phone). This dataset differs from LineMOD in scene layouts, background appearance, camera intrinsics, and slightly varying CAD model sizes, making it a suitable benchmark to evaluate the generalization ability of \acronym~to unseen variations. We use the second sequence for testing. These three real-world datasets serve as target domains, all sharing the same synthetic source domain from BOP challenge \cite{hodavn2020bop}, which contains 50k PBR-generated images. For a more complex scenario, we use \textbf{SPEED+}~\cite{speedplus2022} designed for spacecraft pose estimation, with 59,960 synthetic images (80/20 train/validation split) and two real subsets captured under varied lighting: Lightbox~(6,740 image) and Sunlamp (2,791 image). For fair comparison with SOTA, all test images are used for adaptation discarding their labels. To mitigate this overlap, we also evaluate on the recent \textbf{SHIRT} dataset \cite{park2023adaptive}, which provides two real sequences (roe1 and roe2) captured under lighting conditions similar to Lightbox. SHIRT serves as an external real-world validation set, enabling us to assess whether the adaptation learned on SPEED+ transfers beyond its own test imagery, and to further evaluate generalization to unseen real images.

\subsubsection{Metrics}
For BOP datasets, we use the Average Distance of Model Points (ADD) \cite{hinterstoisser2011gradient} for asymmetric objects, which measures the average distance between 3D object points transformed by the estimated pose and those transformed by the ground truth. For symmetric objects, we use ADD-S \cite{xiang2018posecnn}, which computes the average distance to the closest model points. Combining both, we report ADD(-S) the Average Recall (\%) of poses with error below 10\% of the object’s diameter on all three datasets. Following~\cite{park2024robust}, evaluation on SPEED+ uses rotation error ($E_R$), translation and normalized translation errors ($E_T$, $E_T^{norm}$), and overall pose error ($E_{pose}$).

\subsubsection{Network Architecture}
Our network follows the design of~\cite{park2024robust}, consisting of an EfficientNet backbone, a Bidirectional Feature Pyramid Network (BiFPN), and a regression head with deconvolution layers that produce $K$ heatmaps, each representing one of the $K$ designated keypoints. The final 6D pose is recovered from the predicted 2D keypoints using PnP-RANSAC, which requires at least four keypoints. For the BOP datasets, all eight keypoints are used, while for the more challenging SPEED+ dataset, only the top five keypoints (ranked by prediction confidence) are employed, yielding better performance. The robustness of \acronym~is evaluated across model scales using two variants: \acronymS, based on EfficientNet-B0 with $\sim$3.9M params, and \acronymM, based on EfficientNet-B5 with $\sim$35M params.

\subsection{Comparative Evalaution} \label{ss:resDis} 
\input{sections/4.8_Table_combined}

% \input{sections/4.4.Table_gen_LMO_HB}
% \input{sections/4.4.Table_gen_HB_SHIRT}

\subsubsection{LineMOD} The results are compared against a range of methods employing different levels of supervision, including models trained solely on synthetic data (GDR-Net \cite{wang2021gdr}, DPODv2 \cite{shugurov2021dpodv2} and MAR \cite{zhang2023manifold}), models trained on annotated real data (GDR-Net and DPODv2), and models trained on synthetic data combined with unlabeled real images (Self6D++ \cite{wang2021occlusion}, MAST \cite{zhang2023manifold}, SMOC-Net \cite{tan2023smoc}, Tex-pose \cite{chen2023texpose} and ONDA-pose \cite{tan2025onda}). Detailed results are reported in Table~\ref{tab:linemod-results}. In particular, \acronymS alone outperforms all methods and even reaches a performance comparable to that of DPODv2, which leverages real annotated images during training. \acronymM achieves even stronger results, demonstrating both the effectiveness of \acronym~for domain adaptation in 6D object pose estimation and its robustness to model size.

\subsubsection{Occluded-LineMOD} Results are compared against methods trained solely on synthetic images (CosyPose \cite{labbe2020cosypose}, GDR-Net and MAR) as well as those leveraging unlabeled real data (Self6D++, MAST, SMOC-Net, Tex-pose and ONDA-pose). Table~\ref{tab:olinemod-results} shows that combining synthetic and unlabeled real images reduces the domain gap, outperforming purely synthetic models.
% As shown in Table~\ref{tab:olinemod-results}, approaches that combine synthetic and unlabeled real images generally reduce the domain gap and outperform purely synthetic trained models. 
Nevertheless, \acronym~surpasses all UDA baselines by a margin of at least 10\% in the ADD(-S) score, demonstrating the effectiveness of the framework even under challenging conditions of severe occlusions.

\subsubsection{SPEED+} The results are presented in Table~\ref{tab:speedplus-results}, with the performance further analyzed by model size, an aspect of particular importance in space applications. Comparisons are made against the top three winners of the SPEC21 challenge \cite{park2023satellite} as well as recent state-of-the-art approaches (Yolov8s \cite{bechini2025robust}, UAKD \cite{ousalah2025uncertainty}, SPNv2 \cite{park2024robust}, PVSPE \cite{yang2024pvspe} and FA-VAE \cite{yanfang2024feature}). Notably, PVSPE and FA-VAE are trained exclusively on synthetic images, whereas SPNv2 and the top SPEC21 winners leverage both synthetic and unlabeled real images, similar to our setup. For smaller models, \acronymS proves highly competitive, achieving the best performance in the Lightbox domain and the second-best in the Sunlamp domain, despite having roughly half the size of the leading model (Lava1302). For medium and larger models, \acronymM continues to perform strongly, securing the second-best result in Lightbox, outperforming models 5x larger, such as VPU, and the fourth-best in Sunlamp.  When averaging performance across both domains, \acronymM ($E_p=0.101$) ranks third overall after VPU ($E_p=0.081$) and TangoU ($E_p=0.082$) models, highlighting that \acronym~achieves consistently strong results in varied real-world space conditions.

% \begin{table*}[t]
% \centering
% \input{sections/4.5.Table-CDP}
% \vspace{0.5em}
% \input{sections/4.6.Table-DUAL}
% \vspace{0.5em}
% \input{sections/4.7.Table_PLR}
% \vspace{-0.5em}
% \end{table*}

\subsubsection{Generalization to unseen domains} Direct feature alignment may raise concerns regarding generalization to unseen domains. To evaluate this, we assess two setups: (i) \acronymM models adapted using only 15\% of real LineMOD images, tested on HomebrewedDB; and (ii) models adapted on the Lightbox domain of SPEED+, tested on the SHIRT dataset. The results are summarized in Tables~\ref{tab:homebrewed-generalization} and \ref{tab:shirt-generalization}. In the HomebrewedDB dataset, \acronymM achieves the second-best overall performance compared to MAST, Tex-Pose and ONDA-Pose, demonstrating strong robustness to changes in scene layout, background, and camera intrinsics. In SHIRT, we compare with \cite{cassinis2023leveraging,park2023adaptive}, where \acronymM achieves better performance. We also report baselines trained only on synthetic data and on synthetic+real SPEED+ images as lower and upper bounds; the performance gap notably narrows, indicating that \acronymM effectively transfers beyond its adaptation domain.

\subsection{Ablation Studies}

To assess the design choices in our framework and quantify the contribution of its components, we conduct a comprehensive ablation study, as detailed below.

\subsubsection{Effectiveness of Two-Stage Pairing Strategy}
\label{sssec:ablation-pairing}
The cross-domain pairing approach consists of two stages: initial selection of candidate based on features followed by refinement based on poses. To assess the effectiveness of this two-step strategy, comparisons are made against single-stage alternatives and ground-truth-based pairings as an upper bound. To measure pair quality, we define a compatibility metric based on pose error thresholds, as shown below:

\vspace{-1em}
\begin{equation}
\text{Compatible}(I_s, I_t) =
\begin{cases}
1, & \text{if } \begin{aligned} 
    &d_{\mathrm{rot}}(q_s,q_t) \leq \text{R}_{th} \ \text{\&} \\ 
    &d_{\mathrm{tran}}(t_s,t_t) \leq \text{T}_{th} 
    \end{aligned} \\
0, & \text{otherwise}
\end{cases}
\label{eq:compatible}
\end{equation}

Based on the above \cref{eq:compatible}, we calculate the number of compatible pairs from the total pairs as the accuracy reported in Table~\ref{tab:domain-accuracy} that shows that the two-stage strategy significantly outperforms the single-stage methods and approaches the performance of ground truth-based pairings, confirming the effectiveness of the two-stage pairing strategy.

\subsubsection{Effectiveness of Dual-Level Alignment} 
% The proposed dual-level alignment strategy simultaneously aligns both backbone and head features to bridge domain gaps at multiple representational levels.
The proposed dual alignment strategy simultaneously aligns the backbone and head features across the source and target domains, bridging the gap at multiple levels. To evaluate its effectiveness, we compare configurations that perform alignment exclusively on the backbone (SL1) features and head (SL2) features against the dual-level (DL) alignment strategy. As shown in \colorbox{green!7}{Table~\ref{tab:few-target-images}}, dual-level alignment consistently outperforms partial variants, underscoring the complementary benefits of aligning both types of features.

\subsubsection{Generalizability with Limited Target Data}
To assess the effectiveness of \acronym~under limited availability of target data, we add experiments using subsets of 10\%/20\% of SPEED+ target images for adaptation and evaluate on the full test set. The results, presented in \colorbox{blue!7}{Table~\ref{tab:few-target-images}}, demonstrate that \acronym~significantly mitigates the domain gap even with substantially reduced target supervision.

\subsubsection{Effectiveness of the CBPR Stage}  
We evaluate the impact of the CBPR stage by comparing the full \acronym framework with a version that omits CBPR (Alignment only). Although alignment alone achieves strong performance, the addition of the CBPR stage consistently improves results across multiple datasets, as shown in Table~\ref{tab:PLR-importance}, confirming the effectiveness of this refinement step in enhancing the robustness of keypoint prediction.
\input{sections/4.9._table_cbpr}

%% file: sections/4.2.Table_O-LM.tex
\begin{table*}[t]
  \centering
  \vspace{0.1cm}
  \caption{Comparison on the Occluded-LineMOD. Top two results are in \textbf{bold} and \underline{underlined}. Metrics: ADD-(S) [\%] $\uparrow$.}
  \resizebox{0.8\textwidth}{!}{
  \begin{tabular}{l|c|cccccccc|c}
    \toprule
    Method&
    Data
      & Ape
      & Can
      & Cat
      & Drill
      & Duck
          & \textit{Eggbox} 
      & \textit{Glue} 
       & Holep. 
     
      & \textbf{Mean} $\uparrow$\\
       \midrule
        CosyPose \cite{labbe2020cosypose}  & \multirow{3}{*}{$S$}& 44.0 & 69.9  & 42.1 & 67.5 & 47.8  & 24.4 & 60.0 & 17.5 & 46.7  \\
        GDR-Net \cite{wang2021gdr}  && 44.0 & 83.9 &49.1 & 88.5 & 15.0 & 33.9 & 75.0 & 34.9 & 52.9  \\
        MAR \cite{zhang2023manifold}  && 44.9 & 78.4 & 40.3 & 73.5 & 47.9 & 26.9 & 72.1 & 58.0 & 55.3  \\
    \midrule
        Self6D++ \cite{wang2021occlusion} & \multirow{7}{*}{$S$+$R^{-}$} & 57.7 & \textbf{95.0} & 52.6 & 90.5 & 26.7 & 45.0 & \underline{87.1} & 23.5 & 59.8  \\
        MAST \cite{zhang2023manifold}  & & 47.6 & 82.9 & 45.4 & 75.0 & 53.7 & 48.2 & 75.3 & 63.0 & 61.4  \\
       SMOC-Net \cite{tan2023smoc}  & & 60.0 & \underline{94.5} & 59.1 & 93.0 & 37.2 & 48.3 & \textbf{89.3}  & 25.0 & 63.3  \\
Tex-Pose \cite{chen2023texpose}  & & 60.5 & 93.4 & 56.1 & 92.5 & 55.5 & 46.0 & 82.8  & 46.5 & 66.7  \\
ONDA-Pose \cite{tan2025onda}  & & 57.7 & 92.0 & 56.7  & \underline{94.0} & 56.1 & 50.6 & 77.9 & 69.5 & 69.3  \\
\rowcolor[gray]{0.93}  \acronymS (Ours)  & & \underline{62.0} & 91.3 & \underline{61.2} & 86.1 & \underline{69.4} & \underline{95.6} & 75.6 & \textbf{93.6} & \underline{79.3} \\
\rowcolor[gray]{0.93}  \acronymM (Ours)  & & \textbf{67.0} & 93.9 & \textbf{71.3} & \textbf{95.0} & \textbf{83.3} & \textbf{98.0} &85.8 & \underline{92.4} & \textbf{85.8}\\
        
     \bottomrule
  \end{tabular}
  }
  \label{tab:olinemod-results}
  % \vspace{-0.18cm}
\end{table*}

%% file: sections/4.3.Table_speedplus.tex
\begin{table*}[t]
\centering
\vspace{-0.08cm}
% LEFT: table 1
\begin{minipage}[t]{0.58\textwidth}
\centering
\caption{Comparison of pose estimation performance across target domains in the SPEED+. Metrics: $E_{T}^{\text{norm}}$ in [.], $E_{R}$ in [$^\circ$], and $E_{\text{pose}}$ in [.]. Results marked with * are SPEC21 winners.}
\vspace{0.08cm}
\label{tab:speedplus-results}
\setlength{\tabcolsep}{2.2pt} % Slightly reduced to accommodate new padding
\footnotesize
\begin{tabular}{l c c !{\vrule\hspace{6pt}} ccc !{\vrule\hspace{6pt}} ccc}
\toprule
% & & & \multicolumn{6}{c}{\textbf{SPEED+ Dataset}} \\ \cmidrule(lr){4-9}
 & \textbf{Size} & \textbf{Data} & \multicolumn{3}{c}{\textbf{Lightbox}} & \multicolumn{3}{c}{\textbf{Sunlamp}} \\
\cmidrule(lr){4-6} \cmidrule(lr){7-9} 
Method & (M) & Type & $ E_{T}^{\text{norm}} \downarrow$ & $E_{R} \downarrow$ & $E_{\text{pose}} \downarrow$ & $E_{T}^{\text{norm}} \downarrow$ & $E_{R} \downarrow$ & $E_{\text{pose}} \downarrow$  \\
\midrule
\multicolumn{2}{l}{\textit{Small models}}&&&&&&& \\
YOLOv8s~\cite{bechini2025robust} & 11.5 & \scriptsize{$S$} &  0.118 & 18.0 & 0.432 &  0.086 & 19.8 & 0.432 \\
UAKD~\cite{ousalah2025uncertainty} & 3.8 & \scriptsize{$S$} &  0.049 & 11.42 & 0.248 &  0.063 & 17.03 & 0.360  \\
% \cmidrule(lr){1-3}
Lava1302 \cite{wang2023bridging}$^{*}$ & 8.2 & \scriptsize{$S+R^-$} &  \underline{0.048} & \textbf{6.66} & \textbf{0.165} & \textbf{0.011} & \textbf{2.73} & \textbf{0.059} \\
\rowcolor[gray]{0.93}  \acronymS (Ours) & 3.9 & \scriptsize{$S+R^-$}  & \textbf{0.029} & \underline{7.80} & \textbf{0.165} &  \underline{0.035} & \underline{11.17} & \underline{0.230}  \\
\midrule
\multicolumn{2}{l}{\textit{Large/Medium models}}&&&&&&& \\
PVSPE~\cite{yang2024pvspe} & 70 & \scriptsize{$S$} & \textbf{0.017} & 4.81 & 0.101  & 0.022 & 8.94 & 0.178  \\
FA-VAE~\cite{yanfang2024feature} & 42 & \scriptsize{$S$}&  0.027 & 4.94 & 0.114   & 0.028 & 5.19 & 0.118 \\
% \cmidrule(lr){3}
TangoU \cite{park2023satellite}$^{*}$ & 50 &\scriptsize{$S+R^-$} &  \underline{0.018} & \textbf{3.19} & \textbf{0.073}  & \underline{0.015} & \underline{4.30} & \underline{0.090} \\
VPU \cite{perez2022spacecraft}$^{*}$ & 190 & \scriptsize{$S+R^-$}  & 0.022 & 4.58 & 0.101  & \textbf{0.012} & \textbf{2.83} & \textbf{0.061}  \\
SPNv2~\cite{park2024robust} & 53 & \scriptsize{$S+R^-$} & 0.024 & 5.58 & 0.122 &  0.030 & 9.79 & 0.197 \\
\rowcolor[gray]{0.93}  \acronymM (Ours) & 35 & \scriptsize{$S+R^-$} &  \underline{0.018} & \underline{3.54} & \underline{0.080} & 0.022 & 5.77 & 0.122  \\
\bottomrule
\end{tabular}

\end{minipage}\hfill
% RIGHT: table 2 (top) + table 3 (bottom)
\begin{minipage}[t]{0.4\textwidth}
\centering

\caption{Generalization on HomebrewedDB (LineMOD Adaptation). Metrics: ADD-(S) [\%] $\uparrow$.}
\vspace{-0.15cm}
\label{tab:homebrewed-generalization}
\setlength{\tabcolsep}{8pt}
\small
\resizebox{\columnwidth}{!}{%
\begin{tabular}{l ccc c}
\toprule
% & \multicolumn{4}{c}{\textbf{HomebrewedDB}} 
% \vspace{0.1em} \\ 
% \cmidrule(lr){2-5}
\textbf{Method} & \textbf{Bvise} & \textbf{Drill} & \textbf{Phone} & \textbf{Mean} \\
\midrule
% Self6D++ \cite{wang2021occlusion} & 75.7 & 89.4 & 76.8 & 80.6 \\
MAST \cite{zhang2023manifold}      & \underline{93.8} & 91.5 & 81.8 & 89.0 \\
Tex-Pose \cite{chen2023texpose}   & 93.1 & \underline{94.8} & 79.3 & 89.1 \\
ONDA-Pose \cite{tan2025onda}      & \textbf{98.2} & \textbf{97.1} & \underline{92.1} & \textbf{95.8} \\

\rowcolor[gray]{0.95} \acronymM (Ours) & 93.0 & 91.5 & \textbf{94.1} & \underline{92.9} \\
\bottomrule
\end{tabular}
}

\vspace{0.18cm} % adjust gap between the two right tables

\caption{Generalization results on the SHIRT dataset (SPEED+ Adaptation).}
\vspace{-0.17cm}
\label{tab:shirt-generalization}
\setlength{\tabcolsep}{3pt}
\renewcommand{\arraystretch}{1.3}
\small
\resizebox{\columnwidth}{!}{%
\begin{tabular}{l c cc cc}
\toprule
% & & \multicolumn{4}{c}{\textbf{SHIRT}} \vspace{0.1em} \\ 
% \cmidrule(lr){3-6}
& \textbf{Data} & \multicolumn{2}{c}{\textbf{roe1}} & \multicolumn{2}{c}{\textbf{roe2}} \\
\cmidrule(lr){3-4} \cmidrule(lr){5-6}
\textbf{Method} & Type & $E_{T} [\text{m}] \downarrow$ & $E_{R} [^\circ] \downarrow$ & $E_{T} [\text{m}] \downarrow$ & $E_{R} [^\circ] \downarrow$ \\
\midrule
%Cassinis \textit{et.al.}
U-AUKF\cite{cassinis2023leveraging} & \scriptsize{$S+A$} & \textbf{0.250}{\tiny$\pm$0.06} & 14.00 {\tiny$\pm$11.0} & 0.380 {\tiny$\pm$0.06} & 9.00 {\tiny$\pm$7.5} \\
%Park \textit{et.al.}
A-UKF\cite{park2023adaptive} & \scriptsize{$S+B$} & 0.303{\tiny$\pm$0.35} & 17.17 {\tiny$\pm$42.2} & 0.203 {\tiny$\pm$0.23} & 4.67 {\tiny$\pm$14.9} \\
\rowcolor[gray]{0.95} & \scriptsize{$S$} & 0.348 {\tiny$\pm$2.30} & 10.74 {\tiny$\pm$27.7} & 0.110 {\tiny$\pm$0.30} & 6.04 {\tiny$\pm$22.2} \\
\rowcolor[gray]{0.95} \acronymM & \scriptsize{$S+R^-$} & \underline{0.266} {\tiny$\pm$0.78} & \underline{10.65} {\tiny$\pm$28.0} & \textbf{0.087} {\tiny$\pm$0.17} & \underline{4.49}{\tiny$\pm$17.7} \\
\rowcolor[gray]{0.95}  & \scriptsize{$S+R$} & 0.269{\tiny$\pm$1.41} & \textbf{10.31}{\tiny$\pm$28.0} & \underline{0.097}{\tiny$\pm$0.42} & \textbf{3.74}{\tiny$\pm$14.2} \\
\bottomrule
\end{tabular}
}
\end{minipage}
\vspace{-0.6cm}
\end{table*}

%% file: sections/4.8_Table_combined.tex
% \begin{table*}[t]
% \centering
% \setlength{\tabcolsep}{5pt}
% \small

% % --- TABLE 4.5: CDP PERFORMANCE ---
% \caption{CDP performance evaluated with the proposed compatibility metric. Results compare Upper Bound (GT), Single-Stage variants, and our Two-Stage framework.}
% \label{tab:domain-accuracy}
% \resizebox{0.9\textwidth}{!}{
% \begin{tabular}{l cccc cccc}
% \toprule
% & \multicolumn{4}{c}{\textbf{Lightbox} (Accuracy \% $\uparrow$)} & \multicolumn{4}{c}{\textbf{Sunlamp} (Accuracy \% $\uparrow$)} \\
% \cmidrule(lr){2-5} \cmidrule(lr){6-9}
% \textbf{Method} & $[.5m, 10^\circ]$ & $[1m, 10^\circ]$ & $[.5m, 15^\circ]$ & $[1m, 15^\circ]$ & $[.5m, 10^\circ]$ & $[1m, 10^\circ]$ & $[.5m, 15^\circ]$ & $[1m, 15^\circ]$\\
% \midrule
% Ground-Truth pairs (Upper Bound) & \underline{47.06} & \underline{59.23} & \underline{79.79} & \underline{88.50} & \underline{48.15} & \underline{55.18} & \underline{80.94} & \underline{89.29} \\
% \midrule
% Single-Stage (Features) & 7.39 & 16.71 & 16.22 & 35.16 & 5.98 & 12.76 & 13.62 & 28.84 \\
% Single-Stage (Predictions) & 36.90 & 44.79 & \textbf{64.66} & 75.95 & 32.32 & 38.45 & \textbf{56.50} & 66.46 \\
% \rowcolor[gray]{0.93} \textbf{Two-Stage (Ours)} & \textbf{37.83} & \textbf{56.88} & 59.63 & \textbf{83.26} & \textbf{32.78} & \textbf{46.76} & 52.38 & \textbf{72.62}\\
% \bottomrule
% \end{tabular}
% }
% \vspace{-0.7em}
% \end{table*}

\begin{table*}[t]
\centering
 \vspace{0.1cm}
\begin{minipage}[t]{0.58\textwidth}
  \centering
  % --- TABLE 4.5: CDP PERFORMANCE ---
 
\caption{CDP performance evaluated with the proposed compatibility metric. Results compare Upper Bound (GT), Single-Stage variants, and our Two-Stage framework.}
\label{tab:domain-accuracy}
\resizebox{\textwidth}{!}{
\begin{tabular}{l ccc ccc}
\toprule
& \multicolumn{3}{c}{\textbf{Lightbox} (Accuracy \% $\uparrow$)} & \multicolumn{3}{c}{\textbf{Sunlamp} (Accuracy \% $\uparrow$)} \\
\cmidrule(lr){2-4} \cmidrule(lr){5-7}
\textbf{Method} & $[.5m, 10^\circ]$ & $[1m, 10^\circ]$  & $[1m, 15^\circ]$ & $[.5m, 10^\circ]$ & $[1m, 10^\circ]$  & $[1m, 15^\circ]$\\
\midrule
Ground-Truth pairs & \underline{47.06} & \underline{59.23}  & \underline{88.50} & \underline{48.15} & \underline{55.18}  & \underline{89.29} \\
\midrule
Single-Stage (Features) & 7.39 & 16.71 &35.16 & 5.98 & 12.76 &  28.84 \\
Single-Stage (Predictions) & 36.90 & 44.79 &  75.95 & 32.32 & 38.45  & 66.46 \\
\rowcolor[gray]{0.93} \textbf{Two-Stage (Ours)} & \textbf{37.83} & \textbf{56.88} &  \textbf{83.26} & \textbf{32.78} & \textbf{46.76}  & \textbf{72.62}\\
\bottomrule
\end{tabular}}
\end{minipage}\hfill
\begin{minipage}[t]{0.40\textwidth}
  \centering
% --- TABLE 4.5: CDP PERFORMANCE ---
\caption{Impact of Feature-alignment levels (Backbone vs Head vs Dual) and Sensitivity to target data availability.}
% \vspace{0.05cm}
\label{tab:few-target-images}
\resizebox{\textwidth}{!}{
 \begin{tabular}{lccc}
    \toprule
    Config. & Data & \textbf{Lightbox} $E_{\!p} \downarrow$ & \textbf{Sunlamp} $E_{\!p} \downarrow$ \\
    \midrule
     \rowcolor{green!7} SL1: Backbone & 100\% & 0.166 & 0.384 \\
     \rowcolor{green!7} SL2: Head & 100\%& 0.177 & 0.260 \\
    \rowcolor{blue!7} DL: Dual &10\% & 0.179 & 0.291 \\
    \rowcolor{blue!7} DL: Dual&  20\% & 0.179 & 0.259 \\
    \rowcolor[gray]{0.93}  DL: Dual & 100\% & \textbf{0.165} & \textbf{0.230} \\
    \bottomrule
    \end{tabular}
    }
\end{minipage}
\vspace{-0.65cm}
\end{table*}

% \begin{table}[t]
% \centering
% \setlength{\tabcolsep}{5pt}
% \small

% % --- TABLE 4.5: CDP PERFORMANCE ---
% \caption{Impact of Feature-alignment levels (Backbone vs Head vs Dual) and Sensitivity to target data availability.}
% \label{tab:few-target-images}
% \resizebox{0.45\textwidth}{!}{
%  \begin{tabular}{lccc}
%     \toprule
%     Config. & Data & \textbf{Lightbox} $E_{\!p} \downarrow$ & \textbf{Sunlamp} $E_{\!p} \downarrow$ \\
%     \midrule
%      SL1: Backbone & 100\% & 0.166 & 0.384 \\
%      SL2: Head & 100\%& 0.177 & 0.260 \\
%     DL: Dual &10\% & 0.179 & 0.291 \\
%     DL: Dual&  20\% & 0.179 & 0.259 \\
%     \rowcolor[gray]{0.93}  DL: Dual & 100\% & \textbf{0.165} & \textbf{0.230} \\
%     \bottomrule
%     \end{tabular}
%     }
% \vspace{-0.3em}
% \end{table}

%% file: sections/4.9._table_cbpr.tex
\vspace{-0.2cm}
\begin{table}[!h]
\centering
\setlength{\tabcolsep}{5pt}
\small

 % --- TABLE 4.7: PLR IMPORTANCE ---
\caption{Effectiveness of \acronym pseudo label refinement. Compared to contrastive alignment only.}
\vspace{-0.1cm}
\label{tab:PLR-importance}
\resizebox{0.45\textwidth}{!}{
\begin{tabular}{l cccc}
\toprule
& \multicolumn{2}{c}{\textbf{ADD(-S) [\%] $\uparrow$}} & \multicolumn{2}{c}{\textbf{$E_{\!pose}$ [-] $\downarrow$}} \\
\cmidrule(lr){2-3} \cmidrule(lr){4-5}
\textbf{Configuration} & \textbf{LM} & \textbf{O-LM} & \textbf{Lightbox} & \textbf{Sunlamp} \\
\midrule
Syn. only & 84.4 & 70.3 & 0.193 & 0.395 \\
Alignment Only & 93.3 & 78.7 & 0.177 & 0.256 \\
\rowcolor[gray]{0.93} \acronymS (Full) & \textbf{93.5} & \textbf{79.3} & \textbf{0.165} & \textbf{0.230} \\
\bottomrule
\end{tabular}
}
\vspace{-0.3cm}
\end{table}

%% file: sections/5.Conclusion.tex
\section{CONCLUSION}
\label{sec:conclusion}
\vspace{-0.1cm}
% This paper introduced \acronym, a unified keypoint-based unsupervised domain adaptation framework for object pose estimation, which integrates cross-domain pairing, contrastive alignment, and pseudo-label refinement. Unlike existing self-supervised methods, \acronym~ addresses domain gaps through explicit feature alignment. We evaluated \acronym~on diverse domains with varying complexities and across different model sizes, demonstrating state-of-the-art performance on multiple benchmarks. Despite its effectiveness, \acronym~has some limitations: it relies on finding pairings between source and target images, which may be challenging when synthetic CAD models differ significantly from real ones, and its patch extraction process depends on uncertain predictions rather than being fully learnable. Future work could address these limitations by focusing on eliminating the need for explicit pairings and exploring learning-based patch extraction.

This paper introduces \acronym, a unified keypoint-based UDA framework for object pose estimation that combines cross-domain pairing, contrastive alignment, and pseudo-label refinement. Unlike prior self-supervised pipelines, \acronym tackles domain gaps via explicit feature alignment. Experiments across diverse domains, varying scene complexity, and multiple model sizes show state-of-the-art results on several benchmarks. \acronym nonetheless has limitations: it depends on finding source–target pairings, which can be difficult when synthetic CAD models differ substantially from real objects, and its patch extraction relies on uncertain predictions rather than being fully learnable. Future work will focus on reducing or removing the need for explicit pairing and exploring learning-based patch extraction.

%% file: appendix/1.analysis.tex
\section{Additional Analysis}
\subsection{Impact of Synthetic Domain Pretraining}
\noindent
We evaluate the effect of synthetic pretraining on the SPEED+ dataset by comparing two settings prior to adaptation with \acronymS~: (i) initializing the backbone (EfficientNet, as described in Section~\ref{ss:experiments}) with ImageNet-pretrained weights and adapting only the regression head, and (ii) pretraining the entire network on the full synthetic dataset before adaptation. \cref{tab:domain-alignment-results} shows that synthetic pretraining of the full network results in a significant improvement in post-adaptation performance compared to ImageNet-only pretraining, confirming that pretraining on the synthetic domain is essential.
\vspace{-0.1cm}
\begin{table}[h]
  \centering
  \vspace{-0.5em}
  \captionof{table}{Performance comparison with partial (i) and full (ii) network synthetic pretraining on SPEED+ domains.}
  \label{tab:domain-alignment-results}
  \begin{tabular}{ccc}
    \toprule
     
    % & Lightbox 
    % & Sunlamp  \\
    Config.  & Lightbox-$E_{pose}[-] \downarrow$ & Sunlamp-$E_{pose}[-] \downarrow$\\
    \midrule
    (i)           
            & 1.65                 & 2.38         \\
    (ii)                
      & 0.165  & 0.230 \\
    \bottomrule
  \end{tabular}
  \vspace{-0.5cm}
\end{table}

\subsection{Effectiveness of CBPR stage}
\noindent
The consistency-based pseudo-label Refinement (CBPR) stage constitutes the final stage of \acronym~and is specifically designed to enhance the model’s prediction confidence on the target domain. This process is executed over $N$ training epochs ($N = 40$ in our experiments), and for each run, we select the epoch yielding the highest number of accepted pseudo-keypoints according to the criteria described in \ref{cbpr}. In \cref{fig:lightbox-refinement} and \ref{fig:sunlamp-refinement}, the blue curve shows the maximum average number of accepted keypoints per 100 images achieved at this stage, revealing a clear upward trend. The red curve depicts the best linear fit computed across all epochs (not restricted to the selected best ones), and likewise indicates a consistent improvement across both domains. Together, these results demonstrate a steady increase in model confidence.
\vspace{-0.1cm}

\subsection{Validation Performance Across CBPR stage}
\begin{figure}[H]
    \centering

    % ===================== Figure 1 =====================
    \begin{minipage}{\linewidth}
        \centering
        \begin{subfigure}{0.8\linewidth}
            \centering
            \includegraphics[width=\linewidth]{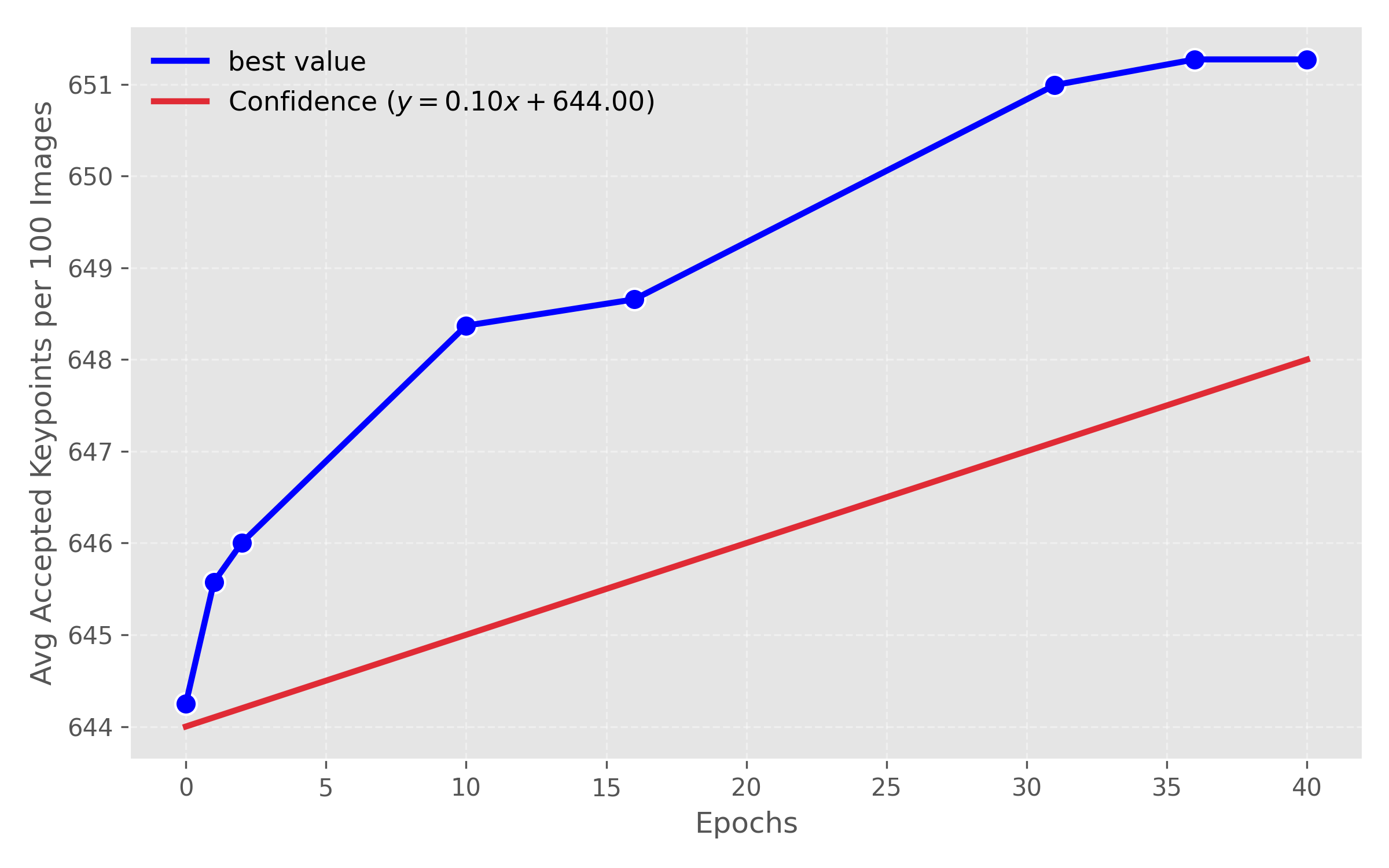}
            \caption{Lightbox}
            \label{fig:lightbox-refinement}
        \end{subfigure}

        \vspace{0.3em}

        \begin{subfigure}{0.8\linewidth}
            \centering
            \includegraphics[width=\linewidth]{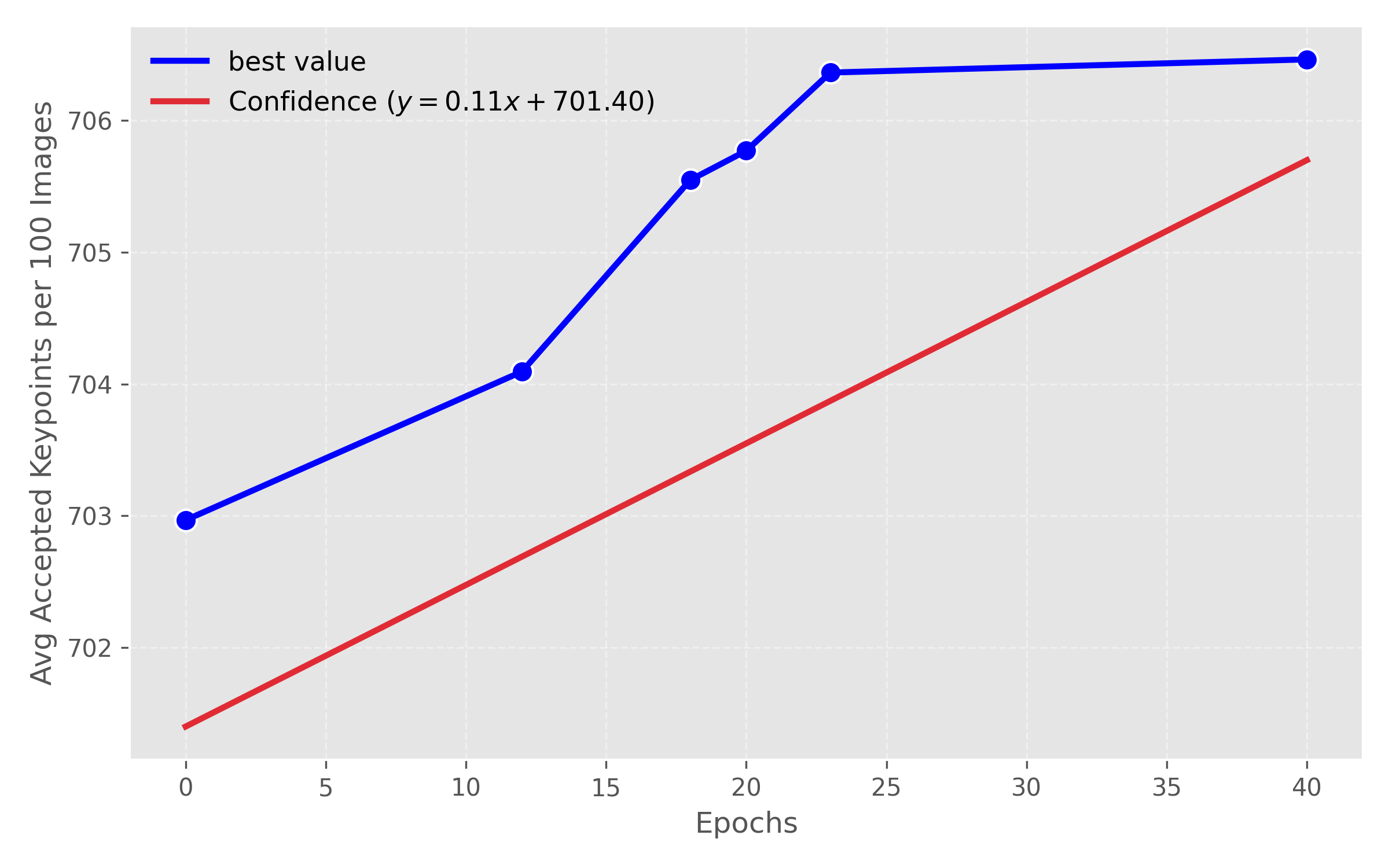}
            \caption{Sunlamp}
            \label{fig:sunlamp-refinement}
        \end{subfigure}

        \captionof{figure}{Accepted pseudo-labeled keypoints across epochs (blue) following the two criterias defined in \ref{cbpr}. The red line indicates the linear trend fitted to the data.}
        \label{fig:refinement-evolution}
    \end{minipage}

    \vspace{2.5em}

    % ===================== Figure 2 =====================
    \begin{minipage}{\linewidth}
        \centering
        \begin{subfigure}{0.9\linewidth}
            \centering
            \includegraphics[width=0.95\linewidth]{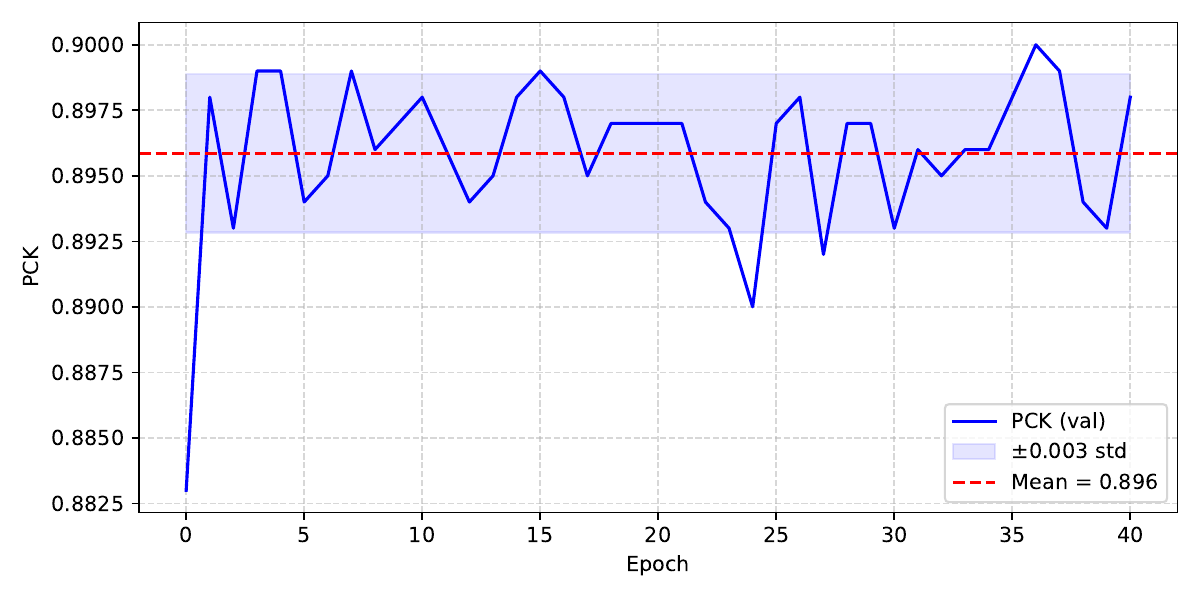}
            \caption{Lightbox}
            \label{fig:cbpr-lightbox}
        \end{subfigure}

        \vspace{0.3em}

        \begin{subfigure}{0.9\linewidth}
            \centering
            \includegraphics[width=0.95\linewidth]{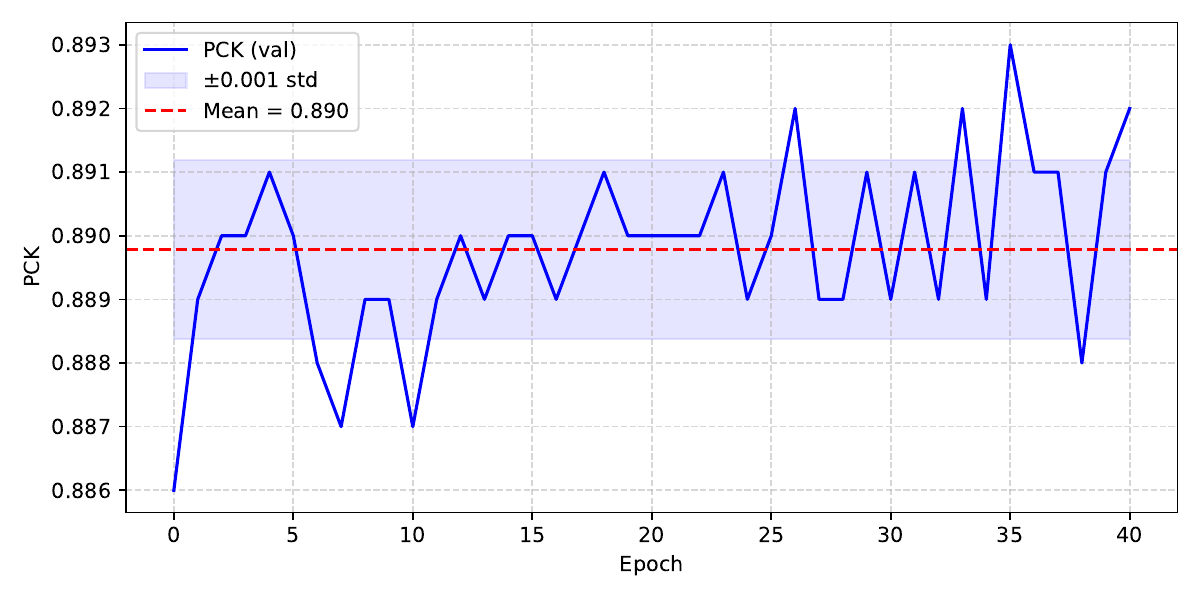}
            \caption{Sunlamp}
            \label{fig:cbpr-sunlamp}
        \end{subfigure}

        \captionof{figure}{Validation performance (PCK, 5\%) on real images during CBPR stage. Performance remains stable across epochs.}
        \label{fig:cbpr_val_comparison}
    \end{minipage}

\end{figure}

\noindent
During the CBPR stage, training relies solely on the loss between predicted keypoints on the augmented images and the pseudo labels, omitting direct task supervision. This could raise concerns about training stability and potential performance degradation. To investigate this, we report the percentage of correct keypoints (within 5\% of image width) on the SPEED+ domains during this stage, shown in Figure \ref{fig:cbpr_val_comparison}. The results exhibit only minor fluctuations, indicating that performance remains largely stable throughout CBPR.

This stability is primarily ensured by two factors. First, the augmentations used are simple and similar to those seen during training, preventing the model from encountering overly unfamiliar inputs. Second, update acceptance conditions based on prediction confidence and spatial consistency, filter out unreliable pseudo-label updates, avoiding detrimental updates from poorly aligned keypoints. Together, these mechanisms maintain stable performance across CBPR epochs.

% \begin{figure}[t]
%     \centering
%     \begin{subfigure}{0.75\linewidth}
%         \centering
%         \includegraphics[width=\linewidth]{appendix_figures/sunlamp.png}
%         \vspace{
%         -0.6cm
%         }
%         \caption{Lightbox}
%         \label{fig:lightbox-refinement}
%     \end{subfigure}
%     % \hfill
%     \vspace{0.5em}
%     \begin{subfigure}{0.75\linewidth}
%         \centering
%         \includegraphics[width=\linewidth]{appendix_figures/lightbox.png}
%         \vspace{
%         -0.6cm
%         }
%         \caption{Sunlamp}
%         \label{fig:sunlamp-refinement}
%     \end{subfigure}
%     \vspace{-0.3cm}
%     \caption{Accepted pseudo-labeled keypoints across epochs (blue) following the two criterias defined in \ref{cbpr}. The red line indicates the linear trend fitted to the data.}
%     \label{fig:refinement-evolution}
% \end{figure}
% \begin{figure}[t]
%     \centering
%     \begin{subfigure}[t]{0.88\linewidth}
%         \centering
%         \includegraphics[width=0.9\linewidth]{appendix_figures/lightbox_pck.pdf}
%         \vspace{
%         -0.3cm
%         }
%         \caption{Lightbox}
%     \end{subfigure}
%     % \hfill
%     \begin{subfigure}[t]{0.88\linewidth}
%         \centering
%         \includegraphics[width=0.9\linewidth]{appendix_figures/sunlamp_pck.pdf}
%         \vspace{
%         -0.3cm
%         }
%         \caption{Sunlamp}
%     \end{subfigure}
%     \vspace{
%     -0.2cm
%     }
% \caption{Validation performance (PCK, 5\%) on real images during CBPR stage. Performance remains stable across epochs.}

%     \label{fig:cbpr_val_comparison}
% \end{figure}
\input{appendix/supp_table_LM}
\input{appendix/supp_table_lmo}
\input{appendix/supp_Table_speed}
\FloatBarrier

\subsection{Extended Results and Ablation on \acronym~Performance}
To better illustrate the gains attributable to \acronym, we report in this section the lower and upper bounds of model performance. The lower bound corresponds to the model trained only on synthetic images, while the upper bound corresponds to a model trained on synthetic images and fine-tuned on real images for the same number of alignment epochs as in \acronym. Results are reported on LineMOD (\autoref{tab:supp-linemod-results}), Occluded-LineMOD (\autoref{tab:supp-linemodo-results}), and SPEED+ (\autoref{tab:supp-speedresults}). For the first two datasets, the real images used for fine-tuning are the same as those used for adaptation; for SPEED+, we use the first 20\% of real images. 

As evident from tables, \acronym~achieves consistent and substantial gains over the synthetic-only baseline. Moreover, in most cases, the improvement from synthetic-only to \acronym~is larger than the difference between the upper-bound (syn+real) model and \acronym, highlighting the effectiveness of \acronym~in closing the gap towards real-data performance.

%% file: appendix/supp_table_LM.tex
\begin{table*}[h]
% \scriptsize
% \footnotesize
  \centering
  \caption{Results on LineMOD – Baseline (S), \acronym~(S+R$^-$), and Real-Finetuned Model (S+R)}
  
  \resizebox{0.95\textwidth}{!}{
  \begin{tabular}{l|c|ccccccccccccc|c}
    \toprule
    Method&
    Data
      & Ape
      &bvise
      &cam
      & Can
      & Cat
      & Drill
      & Duck
          & \textit{Eggbox} 
      & \textit{Glue} 
       & Holep 
       & iron
       &lamp
       &phone
     
      & \textbf{Mean} $\uparrow$\\
       \midrule
    \multirow{3}{*}{\acronymS} & {S}& 49.7 & 94.5 & 74.1 & 93.3 & 95.1 & 94.3 & 78.3 & 96.9 & 84.2 & 91.1 & 98.7 & 63.8 & 82.8 & 84.4 \\
 &{S+R$^-$}& 82.1 & 95.2 & 93.2  & 97.4 & 98.8 & 97.0 & 89.5 & 100.0 &  96.9 & 95.1 & 98.8 & 78.0 & 93.7 & 93.5  \\
  &{S+R}& 95.1 & 98.9 & 95.1  & 95.8 & 93.1 & 99.7 & 96.3 & 100 &  96.1 & 98.7 & 99.8 & 86.6 & 96.8 & 96.3 \\
     \midrule
    \multirow{3}{*}{\acronymM} &{S}& 75.2 & 99.5 & 85.1 & 95.6 & 96.6 & 95.1 & 96.0 & 99.8 & 84.6 & 92.9 & 99.7 & 75.2 & 90.9 & 90.8 \\
      &{S+R$^-$}& 90.7 & 99.5 & 93.2 & 97.9 & 99.3 & 97.7 & 98.8 & 100.0 & 99.5 & 97.0 & 99.9 & 94.1 & 96.6 & 97.2 \\        
  &{S+R}& 98.1 & 99.9 & 97.7 & 98.4 & 100 & 99.9 & 99.9 & 100 & 98.6 & 99.7 & 100 & 96.8 & 99.9 & 99.2 \\
        
     \bottomrule
  \end{tabular}
  }
  \label{tab:supp-linemod-results}
  
\end{table*}

%% file: appendix/supp_table_lmo.tex
\begin{table*}[h]
  \centering
  \caption{Results on Occluded-LineMOD – Baseline (S), \acronym~(S+R$^-$), and Real-Finetuned Model (S+R)}
  \resizebox{0.7\textwidth}{!}{
  \begin{tabular}{l|c|cccccccc|c}
    \toprule
    Method&
    Data
      & Ape
      & Can
      & Cat
      & Drill
      & Duck
          & \textit{Eggbox} 
      & \textit{Glue} 
       & Holep 
     
      & \textbf{Mean} $\uparrow$\\
       \midrule

    \multirow{3}{*}{\acronymS} &{S}& 54.6 & 80.3 & 56.1 & 79.6 & 55.8 & 91.1 & 62.9 & 80.3 & 70.3  \\
  & {S+R$^-$} & 62.0 & 91.3 & 61.2 & 86.1 & 69.4 & 95.6 & 75.6 & 93.6 & 79.3 \\

 &{S+R}& 85.0 & 95.8 & 87.7  & 90.3 & 82.9 & 95.3 & 80.6 & 98.4 &  89.5 \\
     \midrule
         \multirow{3}{*}{\acronymM} &{S}& 54.6 & 84.0 & 54.5 & 92.5 & 79.5 & 97.4  & 68.6 & 86.7  & 77.2  \\
           &{S+R$^-$} & 67.0 & 93.9 & 71.3 & 95.0 & 83.3 & 98.0 &85.8 & 92.4 & 85.8\\         
 &{S+R}& 90.0 & 97.4 & 95.1 & 96.8 & 88.4 & 98.4 & 87.9 & 99.0 & 94.1  \\
        
     \bottomrule
  \end{tabular}
  }
  \label{tab:supp-linemodo-results}
\end{table*}

%% file: appendix/supp_Table_speed.tex
\begin{table*}[htb!]

\centering
\caption{ Results on SPEED+ – Baseline (S), \acronym~(S+R$^-$), and Real-Finetuned Model (S+R)}
\resizebox{\textwidth}{!}{%
\begin{tabular}{l|
    c |  % Size
    c c c|   % Lightbox metrics
    c c c | 
     c% Sunlamp metrics
}
\toprule
& \textbf{DATA}
& \multicolumn{3}{c}{\textbf{Lightbox}} & \multicolumn{3}{c}{\textbf{Sunlamp}} & \multicolumn{1}{c}{\textbf{Mean}} \\
\cmidrule(lr){3-5} \cmidrule(lr){6-8} \cmidrule(lr){9-9}
      Method & (M) & $E_{T} [m] $ / ${E}_{T}^{norm} [.] \downarrow$ & $E_{R}$ [\textdegree ] $\downarrow$ & $E_{\!pose} [.] \downarrow$  & $E_{T} [m] $ /  ${E}_{T}^{norm} [.] \downarrow$ & $E_{R}$ [\textdegree ] $\downarrow$ & $E_{\!pose} [.] \downarrow$ &$E_{\!pose} [.]\downarrow$ \\
      \midrule
         % \underline{Small models} &&&&&&&\\
            \multirow{3}{*}{\acronymS}&S
        & 0.221 / 0.036  & 8.98   & 0.193 &   0.426/0.065  & 18.93   & 0.395 &0.294  \\
         & S+R$^-$        &  0.173 / 0.029   & 7.80   & 0.165  & 0.215 / 0.035  & 11.17   & 0.230 & 0.198\\
         &S+R
        & 0.127 / 0.020   & 4.74   & 0.103 & 0.131 / 0.021  & 5.38   & 0.115 & 0.109\\
        \midrule
        \multirow{3}{*}{\acronymM}&S
        & 0.162 / 0.026   & 5.31   & 0.119  & 0.288 / 0.044  & 11.09   & 0.237 & 0.178\\
        
        & S+R$^-$ & 0.103 / 0.018   & 3.54   & 0.080  & 0.126 / 0.022  & 5.77   & 0.122 & 0.101\\

        & S+R & 0.097 / 0.016   & 2.99   & 0.068  & 0.106 / 0.017  & 3.69   & 0.082 & 0.075\\
    
         \bottomrule   
\end{tabular}
}
\label{tab:supp-speedresults}

\end{table*}

%% file: appendix/2.implementation.tex
\section{Technical details}
\subsection{Network Architecture and Training}
\begin{figure*}[h]
\includegraphics[width=\textwidth]{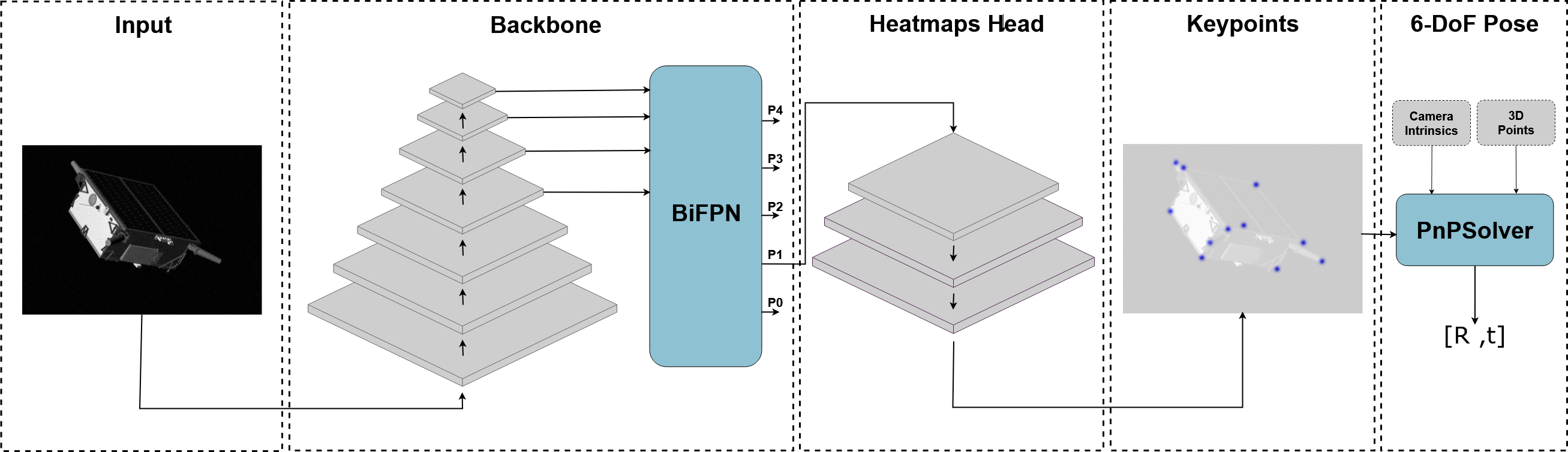}
\caption{Architecture used in experiments, comprising an EfficientNet backbone, a BiFPN layer, and a heatmap head with three deconvolution layers.}
\label{fig:architecture}
\end{figure*}

The adopted architecture consists of an EfficientNet backbone, followed by a Bidirectional Feature Pyramid Network (BiFPN) and a regression head with three deconvolution layers that generate $K$ heatmaps, each corresponding to one of the $K$ annotated keypoints (\cref{fig:architecture}). The BiFPN produces features at multiple levels, each capturing different amounts of information. For the CDP stage, we use the P4 features due to their large receptive fields and robustness to stylistic variations \cite{park2024robust}. For the alignment, we rely on the P1 features, as they directly drive the final predictions.

For the BOP datasets, training is performed exclusively on synthetic images without any data augmentation. Input images are padded by 16 pixels on the top and bottom (resulting in a resolution of $640 \times 512$). The corresponding output heatmaps are generated at a resolution of $320 \times 256$. Training is conducted for 5 epochs using the AdamW optimizer~\cite{loshchilov2017decoupled}, with a learning rate of $10^{-4}$ and a batch size of 16.

For the SPEED+ dataset, synthetic data is also used, but augmented with the transformations proposed in~\cite{park2024robust}. Input images are padded by 40 pixels on the top and bottom to match the original image ratio ($1920 \times 1280$), and resized to $768 \times 512$. The heatmap resolution is set to $384 \times 256$. Training follows the same optimizer and batch size settings, but is extended to 80 epochs.

All experiments are conducted on a single NVIDIA A6000 GPU.

During the Cross-Domain Pairing (CDP) stage, the number of initial candidates is fixed at 1000, with patch sizes of 
$12 \times 12$
($10 \times 10$) for backbone features and 
$10 \times 10$
($10 \times 10$) for head features on SPEED+ (BOP datasets), as this configuration yields the best empirical performance. We do not follow a strict rule for choosing the patch size; rather, we observe that it seems to depend on how well the synthetic-trained model performs on the real domain, where better performance allows the use of smaller patches, and also on the object size, with smaller objects tending to favor smaller patches, The weighting parameter  $\lambda_{rot}$ in \cref{eq:pose_error} is set to 2 to give more emphasis to orientation differences, as the patch-level alignment is inherently more sensitive to angular variations than to translational ones. In the Local Patch Contrastive Alignment (LPCA) stage, the contrastive loss usually converges within five epochs. The model is evaluated on the target domain at each epoch, and the checkpoint with the highest percentage of correct keypoints score is retained.

Finally, the Consistency-Based Pseudo-Refinement (CBPR) stage is performed exclusively on the target domain. We use the augmentation set consisting of Blur, Emboss, and FancyPCA from the Albumentations library. Training runs for 40 epochs, and the checkpoint selected for final evaluation corresponds to the one that yields the highest number of accepted keypoints under the consistency criterion defined in \cref{cbpr}.
\vspace{0.1cm}
\subsection{Loss Function Choice}
\textit{\textbf{A. Heatmap Supervision with KL Divergence Loss:}}

The heatmap regression model outputs $K$ heatmaps, each corresponding to one of the $K$ keypoints. To supervise the heatmap regression, the Kullback-Leibler divergence (KLD) loss is used. The KLD loss makes the predicted heatmaps more uniformly distributed around the correct keypoint locations by reducing spatial divergence \cite{luo2021rethinking}. An alternative approach is to extract keypoints from the predicted heatmaps (e.g., using argmax or soft-argmax) and compute the mean squared error (MSE) between the predicted and ground-truth keypoint coordinates. Two training strategies were initially explored: (a) using only the KLD loss and (b) combining KLD with the MSE loss. In the combined setting, a weighting factor $\lambda$ was introduced to balance the two losses to ensure  that they operate on a similar scale. However, MSE supervision at the coordinate level may weaken the training signal by smoothing gradients or encouraging less sharply defined heatmaps, as it does not directly enforce spatial distribution alignment. In contrast, KLD loss alone promotes accurate localization and well-formed heatmaps. The results of training with synthetic data (\Cref{tab:kldiv}) indicate that using KLD loss alone produces the best performance, supporting its selection as the final loss function.
\begin{table}[H]
  \centering
  \caption{Pose Error ($E_{pose}$) Comparison of performance using KLD loss alone versus a combination of KLD on heatmaps and MSE on keypoints.}
  \label{tab:kldiv}
  \begin{tabular}{lcc}
    \toprule ~\vspace{0.5em}
    Configuration
    & Lightbox - $E_{pose}[-]~\downarrow$    & Sunlamp - $E_{pose}[-]~\downarrow$   \\
    \midrule
    KLD     
      & 0.193                           & 0.395          \\
    KLD + $\lambda$MSE                
      & 0.257 & 0.512 \\
    \bottomrule
  \end{tabular}
\end{table}
\vspace{0.2cm}

\textit{\textbf{B. Contrastive Alignment with InfoNCE Loss:}} 

To align the features between the synthetic and real domains, InfoNCE loss is used due to its strong theoretical foundation and practical effectiveness in learning contrastive representation. InfoNCE maximizes a lower bound on mutual information between positive pairs, promoting semantic consistency across domains while distinguishing them from unrelated negative samples~\cite{oord2018representation}. Its softmax-based formulation over large negative sets generally ensures stable gradients and efficient use of negatives, which is particularly beneficial in unsupervised domain adaptation, where explicit supervision is not available. Compared to alternative contrastive losses, such as triplet loss, InfoNCE benefits from using many negatives per anchor without the need for hard negative mining or manually tuned margins, leading to more robust and scalable training. While losses like NT-Xent (used in SimCLR) are also effective, they are essentially temperature-scaled and normalized variants of InfoNCE \cite{chen2020simple}. However, InfoNCE is preferred here because of its interpretability in terms of mutual information maximization and its ease of integration into the alignment stage of our framework.

\subsection{Heatmaps generation for training}
For keypoint regression, the ground-truth 2D coordinates are converted into target heatmaps using a Gaussian rendering function. For each keypoint, a 2D Gaussian distribution centered at its annotated location is generated with the $render\_gaussian2d$ function from the DSNT library:
{\footnotesize
\begin{lstlisting}[breaklines=true, breakatwhitespace=true]
target = dsnt.render_gaussian2d(mean=keypoints, std=[1,1], size=(384,256))
\end{lstlisting}
}

% {\footnotesize
% \begin{verbatim}
% target = dsnt.render_gaussian2d(mean=keypoints, std=[1,1], size=(384,256))
% \end{verbatim}
% }

This procedure yields one heatmap per keypoint, where the peak intensity encodes the confidence of the keypoint localization. The peak value ranges between 0 and 1, serving as a confidence score. During inference, the predicted keypoint location is obtained by identifying the pixel with maximum value in the corresponding heatmap.

\subsection{Keypoint Extraction}
Predicted keypoints are extracted from the output heatmaps through the following procedure:
\begin{itemize}
\item[A. ] \textit{\textbf{Initial Localization:}} Identification of coordinates corresponding to the maximum value in each heatmap, referred to as coarse keypoints.
\item[B. ] \textit{\textbf{Subpixel Refinement:}} Enhancement of coarse keypoints via second-order Taylor expansion applied on a local 3×3 patch around each coarse prediction. This refinement technique improves localization accuracy beyond the heatmap resolution, as demonstrated in Table \ref{tab:refinement_comparison}. The refinement approach adapted from \cite{Xiao_2018_ECCV} calculates a subpixel offset utilizing image gradients and a local quadratic approximation as detailed in \cref{alg:subpixel-refinement} below.

\begin{algorithm}[!h]
  \caption{Subpixel Keypoint Refinement}
  \label{alg:subpixel-refinement}
  \begin{algorithmic}[1]
    \Require A set of predicted keypoints and their heatmaps
    \Ensure  Refined keypoints 
    \ForAll{predicted keypoint $k$}
      \State Extract a $3\times3$ patch from the heatmap around $k$
      \State Compute gradients $\;d_x,d_y\;$ and Hessian matrix $H$ from the patch
      \If{$H$ is invertible}
        \State Compute subpixel offset :
        \State  $ \quad
            \Delta = H^{-1}
            \begin{bmatrix}
              -d_x  -d_y
            \end{bmatrix}
          $
        \State Refine keypoint:
          $ \quad
            k \gets k + \Delta
          $
      \EndIf
    \EndFor
    \State \Return refined keypoints
  \end{algorithmic}
\end{algorithm}
\end{itemize}
\vspace{-0.4cm}
\subsection{6-DoF Pose Recovery}
To estimate the 6-DoF pose from the predicted keypoints, the Perspective-n-Point (PnP) algorithm with RANSAC is initially employed, followed by pose refinement. The procedure is as follows:
\begin{itemize}
\item[A. ] \textit{\textbf{Initial Estimation:}} The initial pose estimation is obtained using the \texttt{solvePnPRansac} function from OpenCV, which estimates the rotation and translation vectors by minimizing reprojection errors under potential outliers.
\item[B. ] \textit{\textbf{Refinement:}} To enhance accuracy, the initial pose estimates from \texttt{solvePnPRansac} are further refined using the Levenberg-Marquardt optimization algorithm via the \texttt{solvePnPRefineLM} function, improving the precision of pose estimation, as validated in Table \ref{tab:refinement_comparison}.
\end{itemize}

\begin{table}[!hpb]
  \centering
  \caption{Advantage of keypoint and pose refinement on pose error, comparing the performance of \acronymS with and without refinement on the SPEED+ real domains.}

    % \vspace{0.5em}
      \setlength{\tabcolsep}{5pt} % Adjust the column spacing
  \label{tab:refinement_comparison}
  % \scriptsize
  \begin{tabular}{lccccc}
    \toprule
    & Lightbox & Sunlamp \\
    % \cmidrule(lr){2-3} \cmidrule(lr){4-5}
    Configuration &   $E_{pose}[-]~\downarrow$ &  $E_{pose}[-]~\downarrow$ \\
    \midrule
    w/o refine         & 0.179  & 0.248 \\
    w/ keypoints refine   & 0.176 & 0.247 \\
    w/ pose refine    & 0.166  & 0.232 \\
    w/ keypoints \& pose refine    & 0.165 & 0.230 \\
    \bottomrule
  \end{tabular}
  \vspace{-0.3cm}
\end{table}

% \begin{table}[!hpb]
%   \centering
%   \caption{Advantage of keypoint and pose refinement on pose error, comparing the performance of $CAPLR_s^{-}$ (synthetic only) and $CAPLR_s^{+}$ (synthetic + unannotated real) with and without refinement on the SPEED+ real domains.}

%     \vspace{0.5em}
%       \setlength{\tabcolsep}{5pt} % Adjust the column spacing
%   \label{tab:refinement_comparison}
%   % \scriptsize
%   \begin{tabular}{lccccc}
%     \toprule
%     & \multicolumn{2}{c}{Lightbox - $E_{pose}[-]~\downarrow$} & \multicolumn{2}{c}{Sunlamp - $E_{pose}[-]~\downarrow$} \\
%     \cmidrule(lr){2-3} \cmidrule(lr){4-5}
%     Configuration &  $CAPLR_s^{-}$ &  $CAPLR_s^{+}$ &  $CAPLR_s^{-}$ &  $CAPLR_s^{+}$ \\
%     \midrule
%     w/o refine        & 0.207 & 0.179 & 0.420 & 0.248 \\
%     w/ keypoints refine   & 0.202 & 0.176 & 0.418 & 0.247 \\
%     w/ pose refine   & 0.194 & 0.166 & 0.397 & 0.232 \\
%     w/ keypoints \& pose refine   & 0.193 & 0.165 & 0.395 & 0.230 \\
%     \bottomrule
%   \end{tabular}
% \end{table}

%% file: appendix/3.visuals.tex
\section{Failure Cases Analysis}
\input{appendix/pairs_analysis}
\input{appendix/failure}

\subsection{UDA Failure Cases}
Although \acronym achieves consistent improvements on most samples, there remain cases where the gains are marginal or even where performance degrades. To better understand these limitations, we selected a few representative examples from SPEED+ and conducted a detailed analysis. This examination (Fig.\ref{fig:failrure}) provides insights into the underlying causes of reduced performance and highlights directions for future research.

\textbf{Sunlamp:} In both samples, the discovered image pairs are very similar (in pose) to the target 
and are generally acceptable. However, the UDA improvements remain limited. For the image in first row, this can be hypothesized due to two main factors. First, the predicted heatmaps fail to highlight all relevant regions in the target image, leading to inaccurate patch extraction that misses important cues. This issue is likely associated to the harsh lighting, which obscure object details leading to confusion in the predictions. Second, the object’s orientation in the image exposes only a single plane with relatively few discriminative features, further reducing the effectiveness of alignment.

For the image in second row, the degradation is more pronounced. In addition to the aforementioned heatmap limitation, the presence of the Earth background in the synthetic image appears to mislead the patch-based alignment, since the corresponding real image only contains a dark background as well as affected by the scattered light around the keypoint regions. This discrepancy pushes the alignment towards unreliable  feature correspondences.

\textbf{Lightbox:} For the third row, the predicted heatmaps successfully highlight relevant regions, enabling accurate patch localization. Nevertheless, the improvement in rotation accuracy remains modest. This is likely due to background differences between the synthetic and real images, combined with the fact that the real image appears significantly darker, limiting the availability of prominent object features.

For the fourth row, the results again show degradation, despite the heatmaps highlighting appropriate regions. Similar to the second row, this behavior can be explained by the influence of backgrounds in the patch between the synthetic and real images, which undermines the alignment quality.
\section{Qualitative Results}
\vspace{-0.4em}
This last section presents qualitative visualizations that illustrate two aspects of the proposed method: (i)~image pairs discovered during the cross-domain pairing stage, and (ii) \acronym~improved predictions over the baseline (only synthetic trained network) on real-world samples.
\vspace{-0.4em}

% \subsection{Cross-Domain Pairs}
% \vspace{-0.4em}

% Pairs discovered by the CDP stage are shown below for the real domains, highlighting semantic and geometric correspondences; rotation and translation differences are also reported.

\input{appendix/3.1.BOP_pairs}

\input{appendix/3.2.Light_pairs}

\input{appendix/3.3.Sun_pairs}

% \vspace{-1em}
% \subsection{Performance Improvements over Baseline}
% \vspace{-0.5em}
% We present qualitative comparisons of outputs before adaptation (w/o UDA) and after applying the $CAPLR$ framework, using samples from three target domains: Lightbox, Sunlamp, and LineMOD. In each visualization, the ground-truth mesh or 3D bounding box is overlaid in \textcolor{red}{red}, while predictions are shown in \textcolor{blue}{blue}, facilitating a clear visual assessment of pose alignment accuracy.

\input{appendix/3.6.viz_bop}

\input{appendix/3.4.viz_light}

\input{appendix/3.5.viz_sun}

%% file: appendix/pairs_analysis.tex
\subsection{Cross-Domain Pairing}
In this section, we analyze the performance of the CDP stage compared to ground-truth pairing. We then examine pairing failure cases, discuss their likely causes, and outline their impact on the overall adaptation performance.

\subsubsection{Comparison with Ground-Truth Pairing: }

\begin{figure}[h]
    \centering
    \vspace{-0.3cm}
    \begin{subfigure}[t]{0.49\linewidth}
        \centering
        \includegraphics[width=\linewidth]{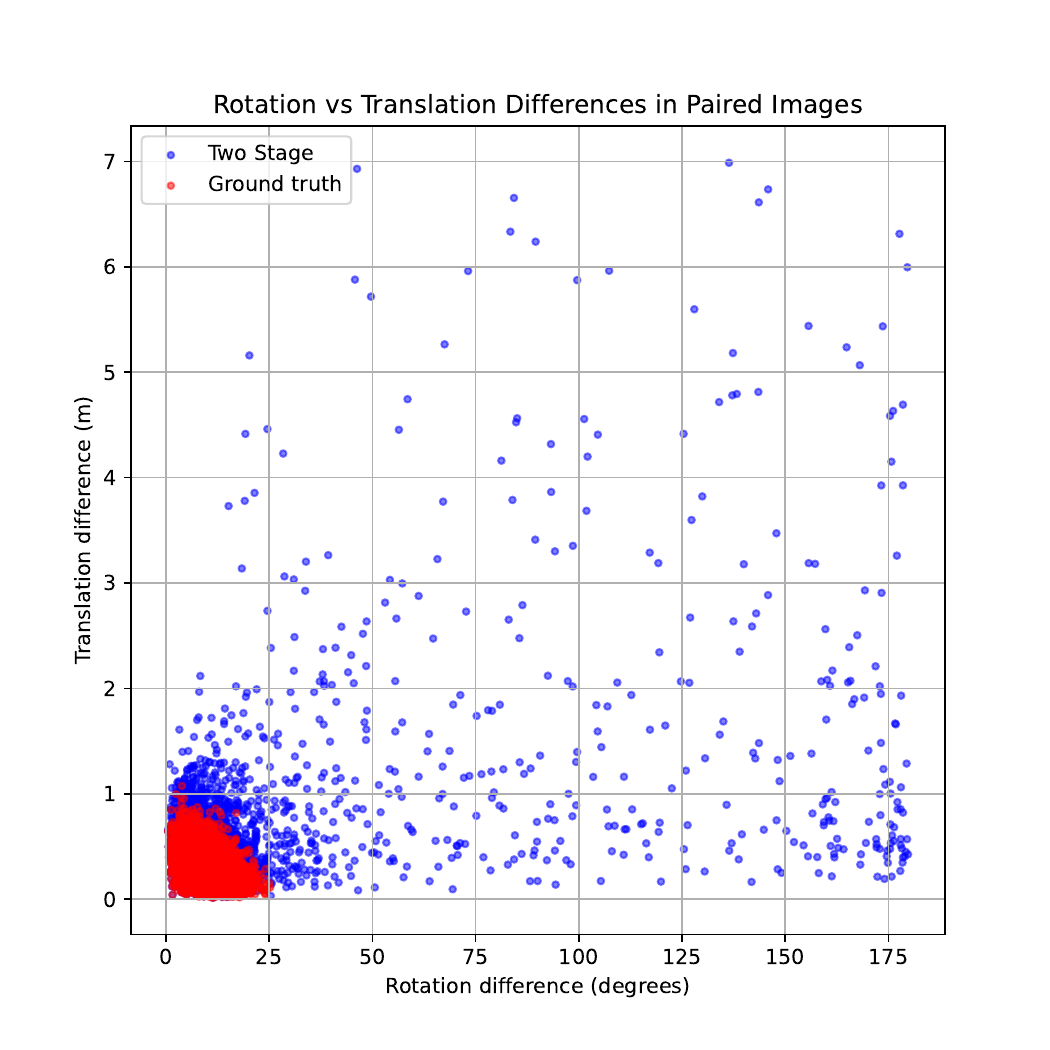}
        \caption{Lightbox}
    \end{subfigure}
    \hfill
    \begin{subfigure}[t]{0.49\linewidth}
        \centering
        \includegraphics[width=\linewidth]{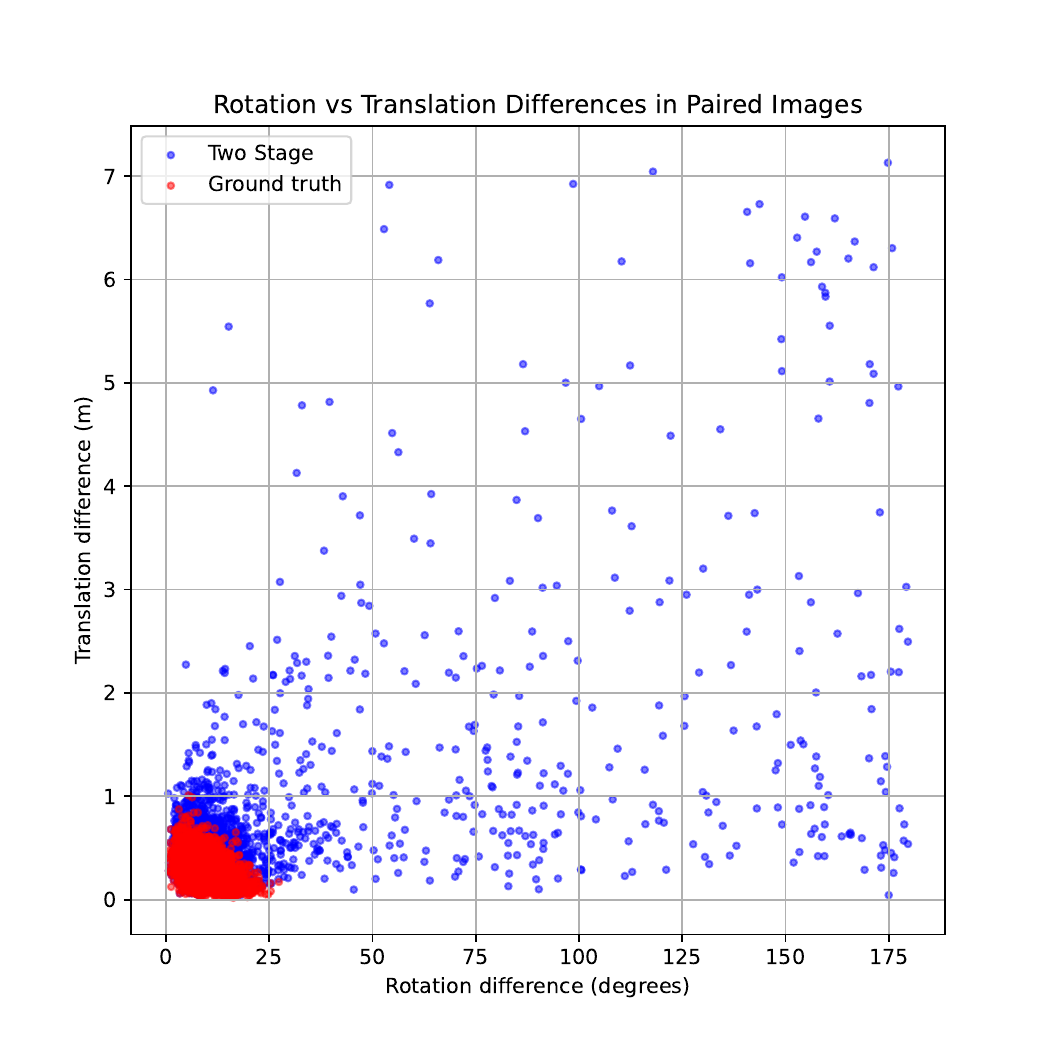}
        \caption{Sunlamp}
    \end{subfigure}
    \caption{Comparison of pairing performance in SPEED+. 
Pairs discovered by the CDP stage are shown in \textcolor{blue}{blue}, and ground-truth pairs are shown in \textcolor{red}{red}. 
In Lightbox, \textcolor{blue}{90.8\%} of CDP pairs fall within the low-error range \([0,1]\text{ m}, [0,25]^\circ\), 
while in Sunlamp, \textcolor{blue}{81.9\%} of CDP pairs fall in this range.}

    \label{fig:two_pairs_comparison}
\end{figure}

% \begin{figure}
%     \centering
%     \includegraphics[width=0.8\linewidth]{appendix_figures/pairs_analysis/lightbox_dist.pdf}
%     \caption{Lightbox-2stages}
%     \label{fig:placeholder}
% \end{figure}
% \begin{figure}
%     \centering
%     \includegraphics[width=0.8\linewidth]{appendix_figures/pairs_analysis/gt_lightbox_dist.pdf}
%     \caption{Lightbox-gt}
%     \label{fig:placeholder}
% \end{figure}
% \begin{figure}
%     \centering
%     \includegraphics[width=\linewidth]{appendix_figures/pairs_analysis/sunlamp_dist-2.pdf}
%     \caption{Sunlamp-2stage}
%     \label{fig:placeholder}
% \end{figure}
% \begin{figure}
%     \centering
%     \includegraphics[width=\linewidth]{appendix_figures/pairs_analysis/gt_sunlamp_dist.pdf}
%     \caption{Sunlamp-gt}
%     \label{fig:placeholder}
% \end{figure}
% \begin{figure}
%     \centering
%     \includegraphics[width=0.5\linewidth]{appendix_figures/pairs_analysis/light_two_pairs_scatter_comparison.pdf}
%     \caption{Lightbox}
%     \label{fig:placeholder}
% \end{figure}
% \begin{figure}
%     \centering
%     \includegraphics[width=0.5\linewidth]{appendix_figures/pairs_analysis/sun_two_pairs_scatter_comparison.pdf}
%     \caption{Sunlamp}
%     \label{fig:placeholder}
% \end{figure}
\input{appendix/pairs_failure}
In Figure \ref{fig:two_pairs_comparison}, we show the distributions of rotational and translational differences between paired samples from the Lightbox and Sunlamp domains of SPEED+. For each domain, we visualize the pairs identified by the CDP stage (\textcolor{blue}{blue}) alongside the pairs obtained directly from ground-truth labels (\textcolor{red}{red}).

On Lightbox, ground-truth pairs lie almost entirely within the region [0,1] m,[0,25]$^{\circ}$ . The CDP stage exhibits a highly similar structure: 90.8\% of CDP-generated pairs fall within this error range, indicating that CDP reliably discovers valid correspondences in this domain. A comparable trend is observed on Sunlamp, where 81.9\% of CDP pairs lie within  [0,1] m,[0,25]$^{\circ}$ . These results demonstrate that the CDP stage produces pairings that are strongly aligned with the ground-truth distribution, supporting its effectiveness and robustness across domains, even in real and more challenging illumination settings.

A small fraction of CDP pairs, however, exhibit larger differences in either rotation or translation. Understanding the conditions under which these higher-error pairs occur is important, as they can affect the overall framework performance. We analyze these cases in detail in the following section.

\subsubsection{Pairing Failure Analysis: }

In Figure \ref{fig:failrure_for_pairs}, we illustrate several of the worst-discovered pairs from the CDP stage in SPEED+, selected based on large rotational and translational differences. The first three rows correspond to the Sunlamp domain, while the last two rows correspond to the Lightbox domain. For each case, we show the real-synthetic pair, the prediction from the synthetic-only model, and the prediction after \acronym adaptation.

\textbf{Sunlamp:} In the first row, part of the object is occluded by shadows, and a bright lighting source appears at the bottom of the image, creating confusion for the model. As seen in the synthetic-only prediction, this leads to an inaccurate pose estimate. After \acronym, the prediction improves, which is likely because the framework can generalize from better-aligned pairs elsewhere in the dataset, allowing it to partially correct errors even for poorly paired examples. The second row presents a similarly challenging case with harsh lighting and partially hidden object features, resulting in only a modest improvement after \acronym. In the third row, the real image contains a lighting source that could be confused with the object itself, explaining the pairing failure; nevertheless, \acronym~still provides some correction, plausibly due to the same generalization mechanism.

\textbf{Lightbox:} In the fourth row, the real image is very dark, obscuring the object and making it difficult to find a good matching pair. Despite this, \acronym~improves the pose estimate significantly, almost matching the ground-truth, likely by leveraging knowledge learned from other correctly aligned pairs. The fifth row also contains occluded and shadowed objects with limited features, leading to poor initial predictions, but \acronym~still yields noticeable improvement.

Overall, failures occur mostly under occlusion, harsh shadows, or confusing light conditions, which reduce the visibility of key object features. Importantly, \acronym~consistently mitigates the impact of these failures, likely by generalizing from correctly aligned pairs, demonstrating its ability to improve predictions even when individual pairings are imperfect.

%% file: appendix/pairs_failure.tex
\begin{figure*}[t]
\begin{center}
\renewcommand{\arraystretch}{0.8}
\begin{tabular}{@{}>{\centering\arraybackslash}m{0.2\textwidth}@{\hspace{3pt}}
                >{\centering\arraybackslash}m{0.2\textwidth}@{\hspace{3pt}}
                >{\centering\arraybackslash}m{0.2\textwidth}@{\hspace{3pt}}
                >{\centering\arraybackslash}m{0.2\textwidth}@{}}

Real Image & Synthetic Pair & w/o UDA & \acronym \\

\begin{tikzpicture}
  \node[inner sep=0pt] (img) at (0,0) {\includegraphics[width=\linewidth]{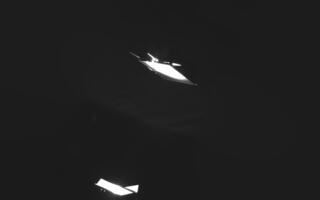}};
  \node[anchor=south west, xshift=-3pt, yshift=-3pt] at (img.south west)
  {\scalebox{0.5}{\colorbox{gray!30}{\begin{tabular}{@{}l@{}} {img000594}\end{tabular}}}};
  \node[anchor=south east, xshift=3pt, yshift=-3pt] at (img.south east)
      {\scalebox{0.5}{{\begin{tabular}{@{}l@{}} {} \\ {}\end{tabular}}}};
       \node[anchor=north east, xshift=-10pt, yshift=2pt] at (img.north east)
  {\scalebox{0.7}{\scriptsize{}}};
\end{tikzpicture} &
\begin{tikzpicture}
  \node[inner sep=0pt] (img) at (0,0) {\includegraphics[width=\linewidth]{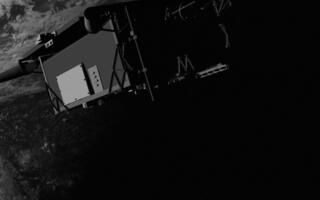}};
  \node[anchor=south west, xshift=-3pt, yshift=-3pt] at (img.south west)
 {\scalebox{0.5}{\colorbox{gray!30}{\begin{tabular}{@{}l@{}} {img032220}\end{tabular}}}};
  \node[anchor=south east, xshift=3pt, yshift=-3pt] at (img.south east)
      {\scalebox{0.5}{{\begin{tabular}{@{}l@{}} {} \\ {}\end{tabular}}}};
          \node[anchor=north east, xshift=-10pt, yshift=2pt] at (img.north east)
  {\scalebox{0.7}{\scriptsize\colorbox{gray!30}{\strut$\Delta R=166.7^\circ,\ \Delta t=6.37$$ m$}}};
\end{tikzpicture} &

\begin{tikzpicture}
  \node[inner sep=0pt] (img) at (0,0) {\includegraphics[width=\linewidth]{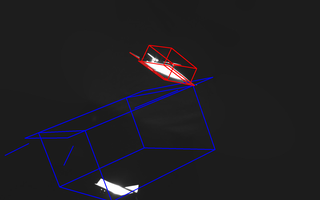}};
    \node[anchor=south west, xshift=-3pt, yshift=-3pt] at (img.south west)
  {\scalebox{0.5}{{\begin{tabular}{@{}l@{}} {}\end{tabular}}}};
  \node[anchor=south east, xshift=3pt, yshift=-3pt] at (img.south east)
      {\scalebox{0.5}{\colorbox{gray!30}{\begin{tabular}{@{}l@{}} {$E_r$: 158.0$^{\circ}$} \\ {$E_t$: 6.28\,$m$}\end{tabular}}}};
       \node[anchor=north east, xshift=-10pt, yshift=2pt] at (img.north east)
  {\scalebox{0.7}{\scriptsize{}}};
\end{tikzpicture} &
\begin{tikzpicture}
  \node[inner sep=0pt] (img) at (0,0) {\includegraphics[width=\linewidth]{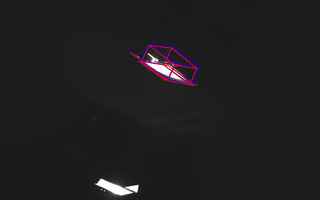}};
    \node[anchor=south west, xshift=-3pt, yshift=-3pt] at (img.south west)
 {\scalebox{0.5}{{\begin{tabular}{@{}l@{}} {}\end{tabular}}}};
  \node[anchor=south east, xshift=3pt, yshift=-3pt] at (img.south east)
      {\scalebox{0.5}{\colorbox{gray!30}{\begin{tabular}{@{}l@{}} {$E_r$: 0.583$^{\circ}$} \\ {$E_t$: 0.14\,$m$}\end{tabular}}}};
  \node[anchor=north east, xshift=-10pt, yshift=2pt] at (img.north east)
  {\scalebox{0.7}{\scriptsize{}}};
\end{tikzpicture} \\

\begin{tikzpicture}
  \node[inner sep=0pt] (img) at (0,0) {\includegraphics[width=\linewidth]{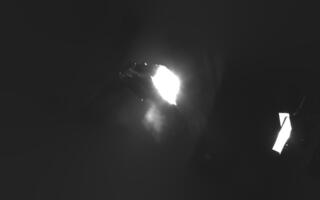}};
  \node[anchor=south west, xshift=-3pt, yshift=-3pt] at (img.south west)
  {\scalebox{0.5}{\colorbox{gray!30}{\begin{tabular}{@{}l@{}} {img001900}\end{tabular}}}};
  \node[anchor=south east, xshift=3pt, yshift=-3pt] at (img.south east)
      {\scalebox{0.5}{{\begin{tabular}{@{}l@{}} {} \\ {}\end{tabular}}}};
       \node[anchor=north east, xshift=-10pt, yshift=2pt] at (img.north east)
  {\scalebox{0.7}{\scriptsize{}}};
\end{tikzpicture} &
\begin{tikzpicture}
  \node[inner sep=0pt] (img) at (0,0) {\includegraphics[width=\linewidth]{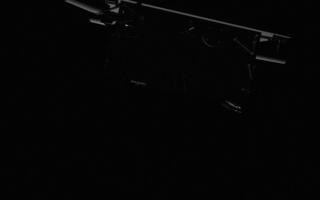}};
  \node[anchor=south west, xshift=-3pt, yshift=-3pt] at (img.south west)
 {\scalebox{0.5}{\colorbox{gray!30}{\begin{tabular}{@{}l@{}} {img002435}\end{tabular}}}};
  \node[anchor=south east, xshift=3pt, yshift=-3pt] at (img.south east)
      {\scalebox{0.5}{{\begin{tabular}{@{}l@{}} {} \\ {}\end{tabular}}}};
          \node[anchor=north east, xshift=-10pt, yshift=2pt] at (img.north east)
  {\scalebox{0.7}{\scriptsize\colorbox{gray!30}{\strut$\Delta R=161.9^\circ,\ \Delta t=6.59$$ m$}}};
\end{tikzpicture} &

\begin{tikzpicture}
  \node[inner sep=0pt] (img) at (0,0) {\includegraphics[width=\linewidth]{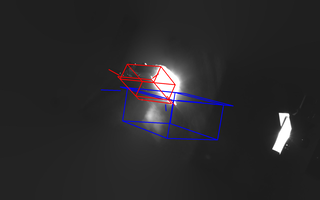}};
    \node[anchor=south west, xshift=-3pt, yshift=-3pt] at (img.south west)
   {\scalebox{0.5}{{\begin{tabular}{@{}l@{}} {}\end{tabular}}}};
  \node[anchor=south east, xshift=3pt, yshift=-3pt] at (img.south east)
      {\scalebox{0.5}{\colorbox{gray!30}{\begin{tabular}{@{}l@{}} {$E_r$: 64.7$^{\circ}$} \\ {$E_t$: 4.36\,$m$}\end{tabular}}}};  \node[anchor=north east, xshift=-10pt, yshift=2pt] at (img.north east)
  {\scalebox{0.7}{\scriptsize{}}};
\end{tikzpicture} &
\begin{tikzpicture}
  \node[inner sep=0pt] (img) at (0,0) {\includegraphics[width=\linewidth]{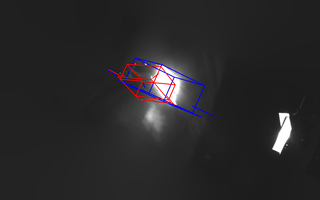}};
    \node[anchor=south west, xshift=-3pt, yshift=-3pt] at (img.south west)
 {\scalebox{0.5}{{\begin{tabular}{@{}l@{}} {}\end{tabular}}}};
  \node[anchor=south east, xshift=3pt, yshift=-3pt] at (img.south east)
      {\scalebox{0.5}{\colorbox{gray!30}{\begin{tabular}{@{}l@{}} {$E_r$: 32.5$^{\circ}$} \\ {$E_t$: 3.35\,$m$}\end{tabular}}}};  \node[anchor=north east, xshift=-10pt, yshift=2pt] at (img.north east)
  {\scalebox{0.7}{\scriptsize{}}};
\end{tikzpicture} \\

\begin{tikzpicture}
  \node[inner sep=0pt] (img) at (0,0) {\includegraphics[width=\linewidth]{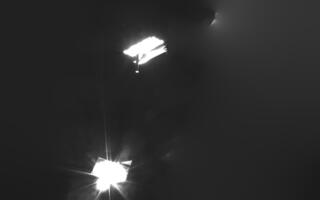}};
  \node[anchor=south west, xshift=-3pt, yshift=-3pt] at (img.south west)
  {\scalebox{0.5}{\colorbox{gray!30}{\begin{tabular}{@{}l@{}} {img001967}\end{tabular}}}};
  \node[anchor=south east, xshift=3pt, yshift=-3pt] at (img.south east)
      {\scalebox{0.5}{{\begin{tabular}{@{}l@{}} {} \\ {}\end{tabular}}}};
       \node[anchor=north east, xshift=-10pt, yshift=2pt] at (img.north east)
  {\scalebox{0.7}{\scriptsize{}}};
\end{tikzpicture} &
\begin{tikzpicture}
  \node[inner sep=0pt] (img) at (0,0) {\includegraphics[width=\linewidth]{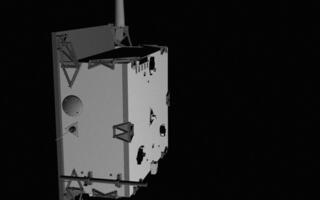}};
  \node[anchor=south west, xshift=-3pt, yshift=-3pt] at (img.south west)
 {\scalebox{0.5}{\colorbox{gray!30}{\begin{tabular}{@{}l@{}} {img011079}\end{tabular}}}};
  \node[anchor=south east, xshift=3pt, yshift=-3pt] at (img.south east)
      {\scalebox{0.5}{{\begin{tabular}{@{}l@{}} {} \\ {}\end{tabular}}}};
          \node[anchor=north east, xshift=-10pt, yshift=2pt] at (img.north east)
  {\scalebox{0.7}{\scriptsize\colorbox{gray!30}{\strut$\Delta R=171.3^\circ,\ \Delta t=6.12$$ m$}}};
\end{tikzpicture} &

\begin{tikzpicture}
  \node[inner sep=0pt] (img) at (0,0) {\includegraphics[width=\linewidth]{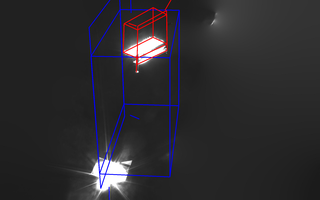}};
    \node[anchor=south west, xshift=-3pt, yshift=-3pt] at (img.south west)
   {\scalebox{0.5}{{\begin{tabular}{@{}l@{}} {}\end{tabular}}}};
  \node[anchor=south east, xshift=3pt, yshift=-3pt] at (img.south east)
      {\scalebox{0.5}{\colorbox{gray!30}{\begin{tabular}{@{}l@{}} {$E_r$: 164.4$^{\circ}$} \\ {$E_t$: 6.07\,$m$}\end{tabular}}}};  \node[anchor=north east, xshift=-10pt, yshift=2pt] at (img.north east)
  {\scalebox{0.7}{\scriptsize{}}};
\end{tikzpicture} &
\begin{tikzpicture}
  \node[inner sep=0pt] (img) at (0,0) {\includegraphics[width=\linewidth]{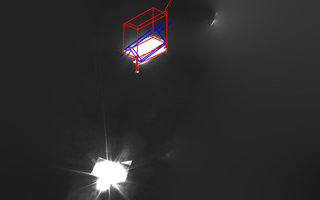}};
    \node[anchor=south west, xshift=-3pt, yshift=-3pt] at (img.south west)
 {\scalebox{0.5}{{\begin{tabular}{@{}l@{}} {}\end{tabular}}}};
  \node[anchor=south east, xshift=3pt, yshift=-3pt] at (img.south east)
      {\scalebox{0.5}{\colorbox{gray!30}{\begin{tabular}{@{}l@{}} {$E_r$: 147.6$^{\circ}$} \\ {$E_t$: 1.58\,$m$}\end{tabular}}}};  \node[anchor=north east, xshift=-10pt, yshift=2pt] at (img.north east)
  {\scalebox{0.7}{\scriptsize{}}};
\end{tikzpicture} \\
\begin{tikzpicture}
  \node[inner sep=0pt] (img) at (0,0) {\includegraphics[width=\linewidth]{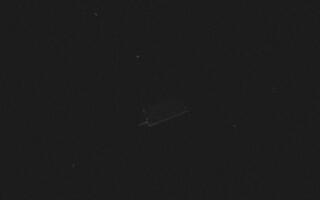}};
  \node[anchor=south west, xshift=-3pt, yshift=-3pt] at (img.south west)
  {\scalebox{0.5}{\colorbox{gray!30}{\begin{tabular}{@{}l@{}} {img000173}\end{tabular}}}};
  \node[anchor=south east, xshift=3pt, yshift=-3pt] at (img.south east)
      {\scalebox{0.5}{{\begin{tabular}{@{}l@{}} {} \\ {}\end{tabular}}}};  \node[anchor=north east, xshift=-10pt, yshift=2pt] at (img.north east)
  {\scalebox{0.7}{\scriptsize{}}};
\end{tikzpicture} &
\begin{tikzpicture}
  \node[inner sep=0pt] (img) at (0,0) {\includegraphics[width=\linewidth]{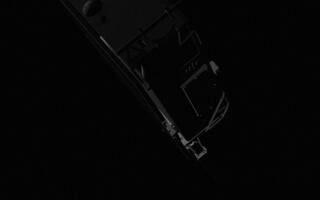}};
  \node[anchor=south west, xshift=-3pt, yshift=-3pt] at (img.south west)
 {\scalebox{0.5}{\colorbox{gray!30}{\begin{tabular}{@{}l@{}} {img002453}\end{tabular}}}};
  \node[anchor=south east, xshift=3pt, yshift=-3pt] at (img.south east)
      {\scalebox{0.5}{{\begin{tabular}{@{}l@{}} {} \\ {}\end{tabular}}}};
          \node[anchor=north east, xshift=-10pt, yshift=2pt] at (img.north east)
  {\scalebox{0.7}{\scriptsize\colorbox{gray!30}{\strut$\Delta R=136.4^\circ,\ \Delta t=6.99$$ m$}}};
\end{tikzpicture} &

\begin{tikzpicture}
  \node[inner sep=0pt] (img) at (0,0) {\includegraphics[width=\linewidth]{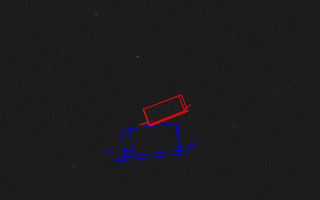}};
    \node[anchor=south west, xshift=-3pt, yshift=-3pt] at (img.south west)
   {\scalebox{0.5}{{\begin{tabular}{@{}l@{}} {}\end{tabular}}}};
  \node[anchor=south east, xshift=3pt, yshift=-3pt] at (img.south east)
      {\scalebox{0.5}{\colorbox{gray!30}{\begin{tabular}{@{}l@{}} {$E_r$: 96.3$^{\circ}$} \\ {$E_t$: 3.81\,$m$}\end{tabular}}}};  \node[anchor=north east, xshift=-10pt, yshift=2pt] at (img.north east)
  {\scalebox{0.7}{\scriptsize{}}};
\end{tikzpicture} &
\begin{tikzpicture}
  \node[inner sep=0pt] (img) at (0,0) {\includegraphics[width=\linewidth]{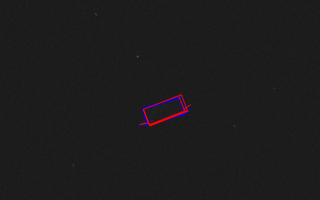}};
    \node[anchor=south west, xshift=-3pt, yshift=-3pt] at (img.south west)
   {\scalebox{0.5}{{\begin{tabular}{@{}l@{}} {}\end{tabular}}}};
  \node[anchor=south east, xshift=3pt, yshift=-3pt] at (img.south east)
      {\scalebox{0.5}{\colorbox{gray!30}{\begin{tabular}{@{}l@{}} {$E_r$: 4.5$^{\circ}$} \\ {$E_t$: 0.09\,$m$}\end{tabular}}}};  \node[anchor=north east, xshift=-10pt, yshift=2pt] at (img.north east)
  {\scalebox{0.7}{\scriptsize{}}};
\end{tikzpicture} \\

\begin{tikzpicture}
  \node[inner sep=0pt] (img) at (0,0) {\includegraphics[width=\linewidth]{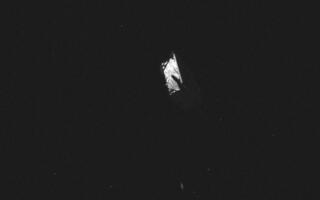}};
  \node[anchor=south west, xshift=-3pt, yshift=-3pt] at (img.south west)
  {\scalebox{0.5}{\colorbox{gray!30}{\begin{tabular}{@{}l@{}} {img000373}\end{tabular}}}};
  \node[anchor=south east, xshift=3pt, yshift=-3pt] at (img.south east)
      {\scalebox{0.5}{{\begin{tabular}{@{}l@{}} {} \\ {}\end{tabular}}}};
       \node[anchor=north east, xshift=-10pt, yshift=2pt] at (img.north east)
  {\scalebox{0.7}{\scriptsize{}}};
\end{tikzpicture} &
\begin{tikzpicture}
  \node[inner sep=0pt] (img) at (0,0) {\includegraphics[width=\linewidth]{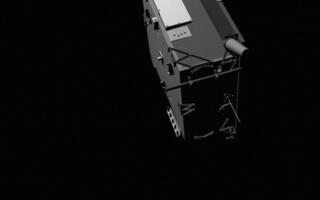}};
  \node[anchor=south west, xshift=-3pt, yshift=-3pt] at (img.south west)
  {\scalebox{0.5}{\colorbox{gray!30}{\begin{tabular}{@{}l@{}} {img012493}\end{tabular}}}};
  \node[anchor=south east, xshift=3pt, yshift=-3pt] at (img.south east)
      {\scalebox{0.5}{{\begin{tabular}{@{}l@{}} {} \\ {}\end{tabular}}}};
          \node[anchor=north east, xshift=-10pt, yshift=2pt] at (img.north east)
  {\scalebox{0.7}{\scriptsize\colorbox{gray!30}{\strut$\Delta R=179.6^\circ,\ \Delta t=6.00$$ m$}}};
\end{tikzpicture} &

\begin{tikzpicture}
  \node[inner sep=0pt] (img) at (0,0) {\includegraphics[width=\linewidth]{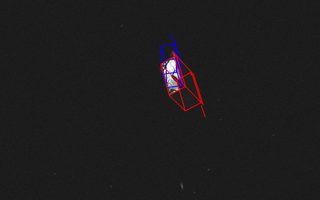}};
    \node[anchor=south west, xshift=-3pt, yshift=-3pt] at (img.south west)
  {\scalebox{0.5}{{\begin{tabular}{@{}l@{}} {}\end{tabular}}}};
  \node[anchor=south east, xshift=3pt, yshift=-3pt] at (img.south east)
      {\scalebox{0.5}{\colorbox{gray!30}{\begin{tabular}{@{}l@{}} {$E_r$: 174.1$^{\circ}$} \\ {$E_t$: 2.38\,$m$}\end{tabular}}}};
       \node[anchor=north east, xshift=-10pt, yshift=2pt] at (img.north east)
  {\scalebox{0.7}{\scriptsize{}}};
\end{tikzpicture} &
\begin{tikzpicture}
  \node[inner sep=0pt] (img) at (0,0) {\includegraphics[width=\linewidth]{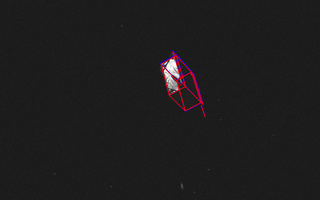}};
    \node[anchor=south west, xshift=-3pt, yshift=-3pt] at (img.south west)
   {\scalebox{0.5}{{\begin{tabular}{@{}l@{}} {}\end{tabular}}}};
  \node[anchor=south east, xshift=3pt, yshift=-3pt] at (img.south east)
      {\scalebox{0.5}{\colorbox{gray!30}{\begin{tabular}{@{}l@{}} {$E_r$: 0.43$^{\circ}$} \\ {$E_t$: 0.23\,$m$}\end{tabular}}}};
    \node[anchor=north east, xshift=-10pt, yshift=2pt] at (img.north east)
  {\scalebox{0.7}{\scriptsize{}}};
\end{tikzpicture} \\

\end{tabular}
% \vspace{-1em}
\vspace{-0.2cm}
\caption{Cross-Domain Pairing failure cases on SPEED+. The first two columns show the discovered real-synthetic pairs along with their rotational and translational differences.  The last two columns show the estimated pose on the real image before and after \acronym. \textcolor{red}{Ground-truth} poses and \textcolor{blue}{predictions} are indicated in red and blue, respectively.}

  \label{fig:failrure_for_pairs}
\end{center}
\vspace{-0.6cm}
\end{figure*}

%% file: appendix/failure.tex
\begin{figure*}[t]
\begin{center}
\renewcommand{\arraystretch}{0.8}
\begin{tabular}{@{}>{\centering\arraybackslash}m{0.23\textwidth}@{\hspace{3pt}}
                >{\centering\arraybackslash}m{0.23\textwidth}@{\hspace{3pt}}
                >{\centering\arraybackslash}m{0.23\textwidth}@{\hspace{3pt}}
                >{\centering\arraybackslash}m{0.23\textwidth}@{}}

w/o UDA & \acronym & \textbf{$Synthetic ~pair$} & \textbf{$w/o~UDA~heatmap$} \\

\begin{tikzpicture}
  \node[inner sep=0pt] (img) at (0,0) {\includegraphics[width=\linewidth]{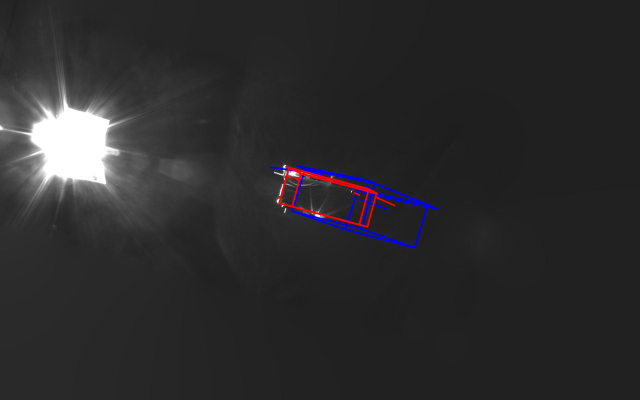}};
  \node[anchor=south west, xshift=-3pt, yshift=-3pt] at (img.south west)
  {\scalebox{0.5}{\colorbox{gray!30}{\begin{tabular}{@{}l@{}} {img001069}\end{tabular}}}};
  \node[anchor=south east, xshift=3pt, yshift=-3pt] at (img.south east)
      {\scalebox{0.5}{\colorbox{gray!30}{\begin{tabular}{@{}l@{}} {$E_r$: 142.3$^{\circ}$} \\ {$E_t$: 1.85\,$m$}\end{tabular}}}};
\end{tikzpicture} &
\begin{tikzpicture}
  \node[inner sep=0pt] (img) at (0,0) {\includegraphics[width=\linewidth]{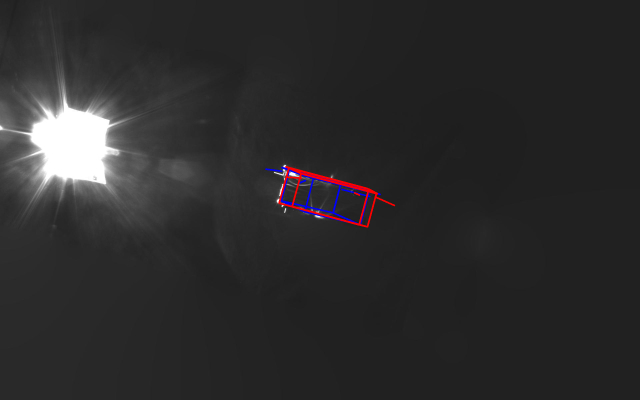}};
  \node[anchor=south west, xshift=-3pt, yshift=-3pt] at (img.south west)
  {\scalebox{0.5}{\begin{tabular}{@{}l@{}} {}\end{tabular}}};
  \node[anchor=south east, xshift=3pt, yshift=-3pt] at (img.south east)
      {\scalebox{0.5}{\colorbox{gray!30}{\begin{tabular}{@{}l@{}} {$E_r$: 121.3$^{\circ}$} \\ {$E_t$: 0.615\,$m$}\end{tabular}}}};
\end{tikzpicture} &

\begin{tikzpicture}
  \node[inner sep=0pt] (img) at (0,0) {\includegraphics[width=\linewidth]{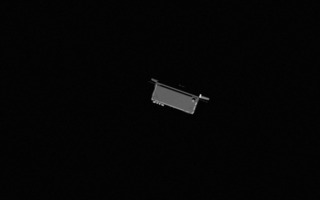}};
    \node[anchor=south west, xshift=-3pt, yshift=-3pt] at (img.south west)
  {\scalebox{0.5}{\colorbox{gray!30}{\begin{tabular}{@{}l@{}} {img021087}\end{tabular}}}};
\end{tikzpicture} &
\begin{tikzpicture}
  \node[inner sep=0pt] (img) at (0,0) {\includegraphics[width=\linewidth]{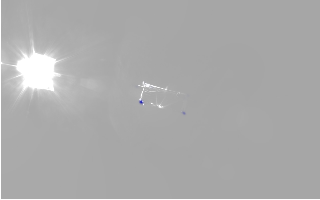}};
    \node[anchor=south west, xshift=-3pt, yshift=-3pt] at (img.south west)
  {\scalebox{0.5}{\begin{tabular}{@{}l@{}} {}\end{tabular}}};
 
\end{tikzpicture} \\

\begin{tikzpicture}
  \node[inner sep=0pt] (img) at (0,0) {\includegraphics[width=\linewidth]{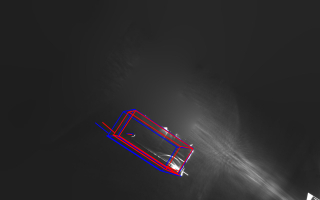}};
  \node[anchor=south west, xshift=-3pt, yshift=-3pt] at (img.south west)
  {\scalebox{0.5}{\colorbox{gray!30}{\begin{tabular}{@{}l@{}} {img001491}\end{tabular}}}};
  \node[anchor=south east, xshift=3pt, yshift=-3pt] at (img.south east)
      {\scalebox{0.5}{\colorbox{gray!30}{\begin{tabular}{@{}l@{}} {$E_r$: 4.55$^{\circ}$} \\ {$E_t$: 0.215\,$m$}\end{tabular}}}};
\end{tikzpicture} &
\begin{tikzpicture}
  \node[inner sep=0pt] (img) at (0,0) {\includegraphics[width=\linewidth]{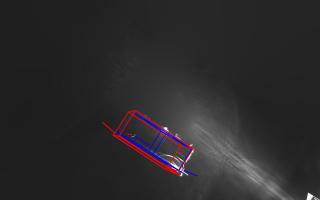}};
  \node[anchor=south west, xshift=-3pt, yshift=-3pt] at (img.south west)
  {\scalebox{0.5}{\begin{tabular}{@{}l@{}} {}\end{tabular}}};
  \node[anchor=south east, xshift=3pt, yshift=-3pt] at (img.south east)
      {\scalebox{0.5}{\colorbox{gray!30}{\begin{tabular}{@{}l@{}} {$E_r$: 133.2$^{\circ}$} \\ {$E_t$: 0.917\,$m$}\end{tabular}}}};
\end{tikzpicture} &

\begin{tikzpicture}
  \node[inner sep=0pt] (img) at (0,0) {\includegraphics[width=\linewidth]{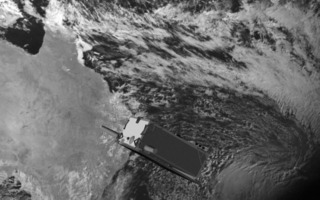}};
    \node[anchor=south west, xshift=-3pt, yshift=-3pt] at (img.south west)
  {\scalebox{0.5}{\colorbox{gray!30}{\begin{tabular}{@{}l@{}} {img042986}\end{tabular}}}};
\end{tikzpicture} &
\begin{tikzpicture}
  \node[inner sep=0pt] (img) at (0,0) {\includegraphics[width=\linewidth]{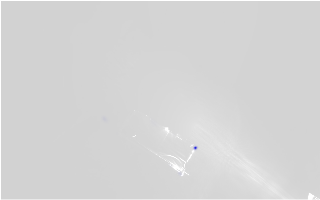}};
    \node[anchor=south west, xshift=-3pt, yshift=-3pt] at (img.south west)
  {\scalebox{0.5}{\begin{tabular}{@{}l@{}} {}\end{tabular}}};
\end{tikzpicture} \\

\begin{tikzpicture}
  \node[inner sep=0pt] (img) at (0,0) {\includegraphics[width=\linewidth]{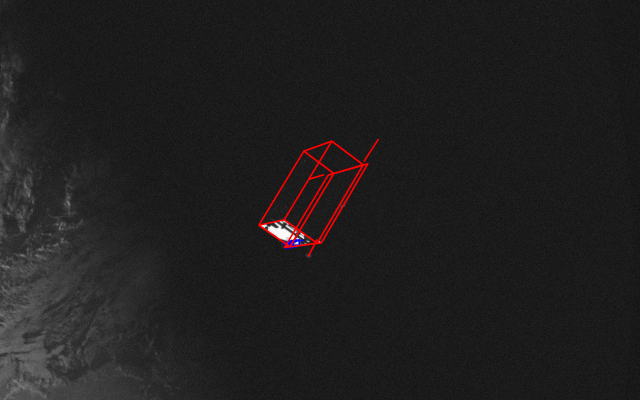}};
  \node[anchor=south west, xshift=-3pt, yshift=-3pt] at (img.south west)
  {\scalebox{0.5}{\colorbox{gray!30}{\begin{tabular}{@{}l@{}} {img000806}\end{tabular}}}};
  \node[anchor=south east, xshift=3pt, yshift=-3pt] at (img.south east)
      {\scalebox{0.5}{\colorbox{gray!30}{\begin{tabular}{@{}l@{}} {$E_r$: 160.6$^{\circ}$} \\ {$E_t$: 55.98\,$m$}\end{tabular}}}};
\end{tikzpicture} &
\begin{tikzpicture}
  \node[inner sep=0pt] (img) at (0,0) {\includegraphics[width=\linewidth]{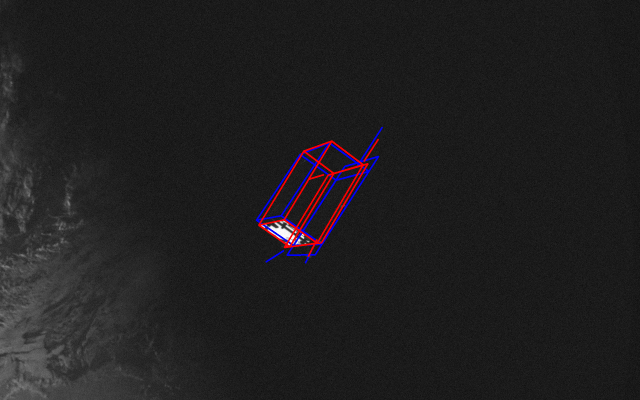}};
  \node[anchor=south west, xshift=-3pt, yshift=-3pt] at (img.south west)
  {\scalebox{0.5}{\begin{tabular}{@{}l@{}} {}\end{tabular}}};
  \node[anchor=south east, xshift=3pt, yshift=-3pt] at (img.south east)
      {\scalebox{0.5}{\colorbox{gray!30}{\begin{tabular}{@{}l@{}} {$E_r$: 96.2$^{\circ}$} \\ {$E_t$: 1.22\,$m$}\end{tabular}}}};
\end{tikzpicture} &

\begin{tikzpicture}
  \node[inner sep=0pt] (img) at (0,0) {\includegraphics[width=\linewidth]{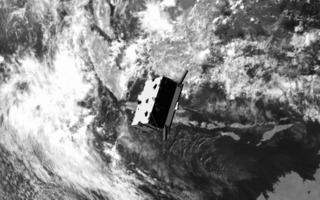}};
    \node[anchor=south west, xshift=-3pt, yshift=-3pt] at (img.south west)
  {\scalebox{0.5}{\colorbox{gray!30}{\begin{tabular}{@{}l@{}} {img038093}\end{tabular}}}};
\end{tikzpicture} &
\begin{tikzpicture}
  \node[inner sep=0pt] (img) at (0,0) {\includegraphics[width=\linewidth]{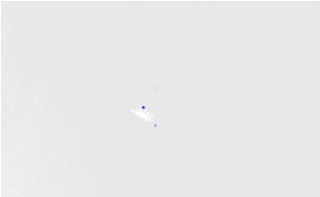}};
    \node[anchor=south west, xshift=-3pt, yshift=-3pt] at (img.south west)
  {\scalebox{0.5}{\begin{tabular}{@{}l@{}} {}\end{tabular}}};
\end{tikzpicture} \\

\begin{tikzpicture}
  \node[inner sep=0pt] (img) at (0,0) {\includegraphics[width=\linewidth]{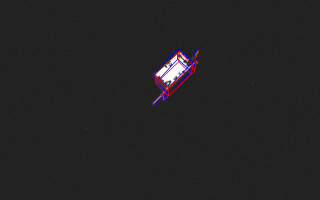}};
  \node[anchor=south west, xshift=-3pt, yshift=-3pt] at (img.south west)
  {\scalebox{0.5}{\colorbox{gray!30}{\begin{tabular}{@{}l@{}} {img004181}\end{tabular}}}};
  \node[anchor=south east, xshift=3pt, yshift=-3pt] at (img.south east)
      {\scalebox{0.5}{\colorbox{gray!30}{\begin{tabular}{@{}l@{}} {$E_r$: 3.96$^{\circ}$} \\ {$E_t$: 0.22\,$m$}\end{tabular}}}};
\end{tikzpicture} &
\begin{tikzpicture}
  \node[inner sep=0pt] (img) at (0,0) {\includegraphics[width=\linewidth]{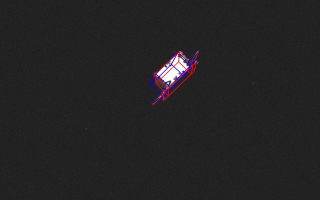}};
  \node[anchor=south west, xshift=-3pt, yshift=-3pt] at (img.south west)
  {\scalebox{0.5}{\begin{tabular}{@{}l@{}} {}\end{tabular}}};
  \node[anchor=south east, xshift=3pt, yshift=-3pt] at (img.south east)
      {\scalebox{0.5}{\colorbox{gray!30}{\begin{tabular}{@{}l@{}} {$E_r$: 170.8$^{\circ}$} \\ {$E_t$: 0.406\,$m$}\end{tabular}}}};
\end{tikzpicture} &

\begin{tikzpicture}
  \node[inner sep=0pt] (img) at (0,0) {\includegraphics[width=\linewidth]{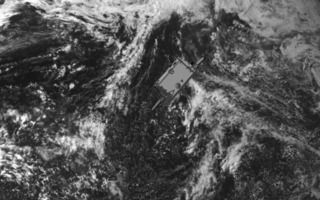}};
    \node[anchor=south west, xshift=-3pt, yshift=-3pt] at (img.south west)
  {\scalebox{0.5}{\colorbox{gray!30}{\begin{tabular}{@{}l@{}} {img038051}\end{tabular}}}};
\end{tikzpicture} &
\begin{tikzpicture}
  \node[inner sep=0pt] (img) at (0,0) {\includegraphics[width=\linewidth]{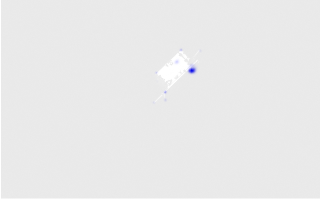}};
    \node[anchor=south west, xshift=-3pt, yshift=-3pt] at (img.south west)
  {\scalebox{0.5}{\begin{tabular}{@{}l@{}} {}\end{tabular}}};

\end{tikzpicture} \\

\end{tabular}
% \vspace{-1em}
% \vspace{-0.2cm}
\caption{Alignment failure cases on SPEED+. First two columns show pose estimates without UDA (left) and with \acronym (right). Ground-truth poses are shown in \textcolor{red}{red} and predictions in \textcolor{blue}{blue}. The third column presents the discovered synthetic pair used for alignment, while the last column shows the synethic-only model predictions that serve as the basis for patch extraction. \vspace{-0.5cm}}

  \label{fig:failrure}
\end{center}
\end{figure*}

%% file: appendix/3.1.BOP_pairs.tex
% \newpage
% \textit{\textbf{A. (Occluded)-LineMOD Pairs}}

% \twocolumn[
% \begin{center}
% \vspace{0.5em}
% {\Large \bfseries A. (Occluded)-LineMOD Pairsl}
% \end{center}
% ]
\begin{figure*}[h]
\begin{center}
{\Large \bfseries A. (Occluded)-LineMOD Pairs}\par\vspace{0.5em}
\renewcommand{\arraystretch}{0.8}
\begin{tabular}{@{}>{\centering\arraybackslash}m{0.23\textwidth}@{\hspace{3pt}}
                >{\centering\arraybackslash}m{0.23\textwidth}@{\hspace{8pt}}
                >{\centering\arraybackslash}m{0.012\textwidth}@{}
                >{\centering\arraybackslash}m{0.23\textwidth}@{\hspace{3pt}}
                >{\centering\arraybackslash}m{0.23\textwidth}@{}}

\textbf{Target} & \textbf{Source} & & \textbf{Target} & \textbf{Source} \\

\begin{tikzpicture}
  \node[inner sep=0pt] (img) at (0,0) {\includegraphics[width=\linewidth]{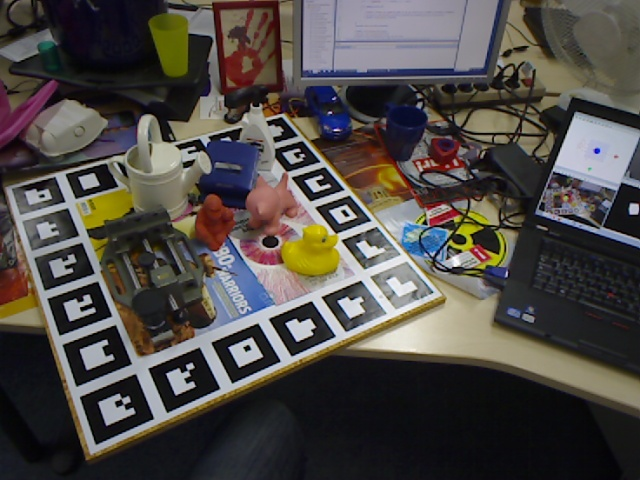}};
      \node[anchor=south west, xshift=-3pt, yshift=-3pt] at (img.south west)
  {\scalebox{0.5}{\colorbox{gray!30}{\begin{tabular}{@{}l@{}} {Ape000037}\end{tabular}}}};
\end{tikzpicture} &
\begin{tikzpicture}
  \node[inner sep=0pt] (img) at (0,0) {\includegraphics[width=\linewidth]{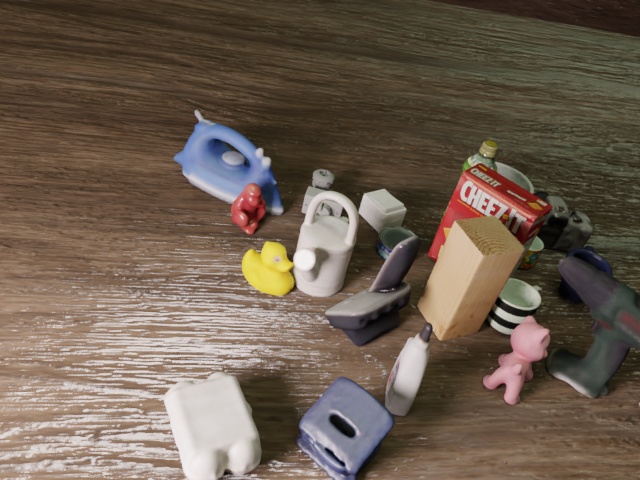}};
     \node[anchor=south west, xshift=-3pt, yshift=-3pt] at (img.south west)
  {\scalebox{0.5}{\begin{tabular}{@{}l@{}}\end{tabular}}};
       \node[anchor=north east, xshift=-10pt, yshift=2pt] at (img.north east)
  {\scalebox{0.7}{\scriptsize\colorbox{gray!30}{\strut$\Delta R=11.56^\circ,\ \Delta t=1.38$$ m$}}};
\end{tikzpicture} &
& 
\begin{tikzpicture}
  \node[inner sep=0pt] (img) at (0,0) {\includegraphics[width=\linewidth]{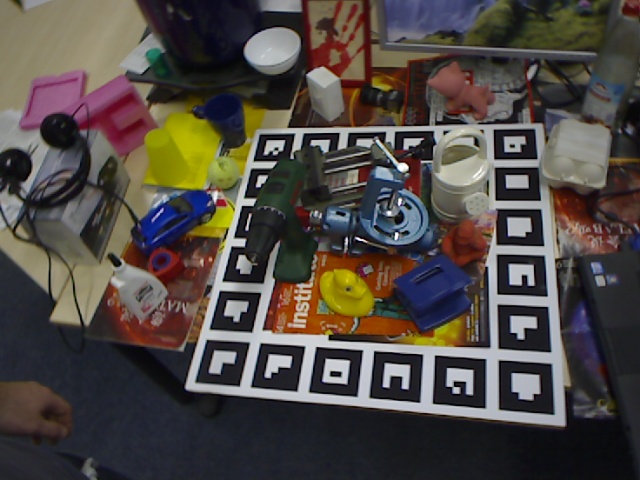}};
   \node[anchor=south west, xshift=-3pt, yshift=-3pt] at (img.south west)
  {\scalebox{0.5}{\colorbox{gray!30}{\begin{tabular}{@{}l@{}} {benchv000410}\end{tabular}}}};
\end{tikzpicture} &
\begin{tikzpicture}
  \node[inner sep=0pt] (img) at (0,0) {\includegraphics[width=\linewidth]{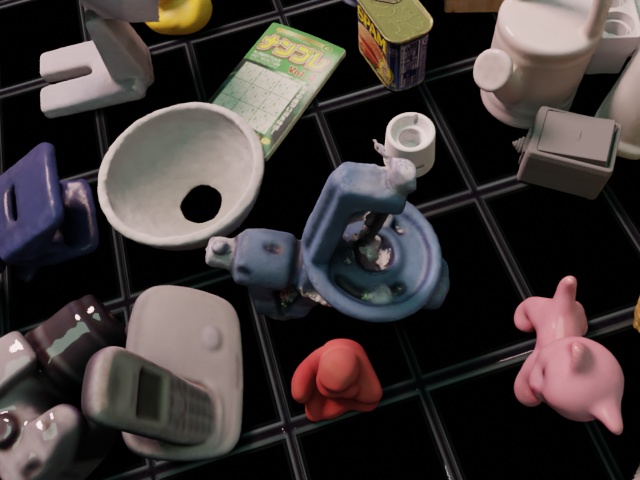}};
       \node[anchor=south west, xshift=-3pt, yshift=-3pt] at (img.south west)
  {\scalebox{0.5}{\begin{tabular}{@{}l@{}}\end{tabular}}};
       \node[anchor=north east, xshift=-10pt, yshift=2pt] at (img.north east)
  {\scalebox{0.7}{\scriptsize\colorbox{gray!30}{\strut$\Delta R=9.7^\circ,\ \Delta t=5.11$$ m$}}};
\end{tikzpicture}  \\

\begin{tikzpicture}
  \node[inner sep=0pt] (img) at (0,0) {\includegraphics[width=\linewidth]{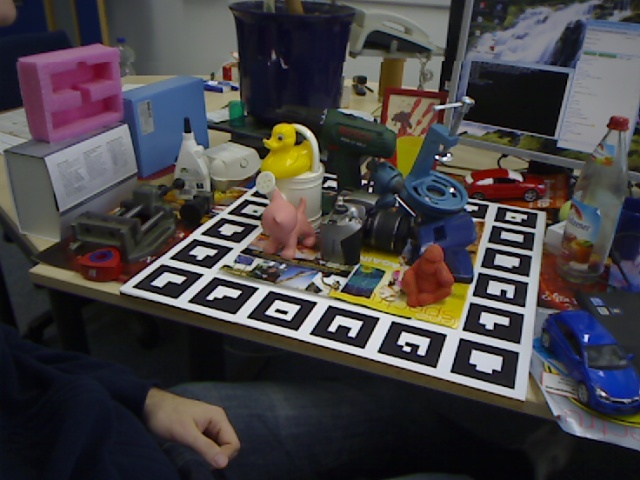}};
  \node[anchor=south west, xshift=-3pt, yshift=-3pt] at (img.south west)
  {\scalebox{0.5}{\colorbox{gray!30}{\begin{tabular}{@{}l@{}} {cam000518}\end{tabular}}}};
\end{tikzpicture} &
\begin{tikzpicture}
  \node[inner sep=0pt] (img) at (0,0) {\includegraphics[width=\linewidth]{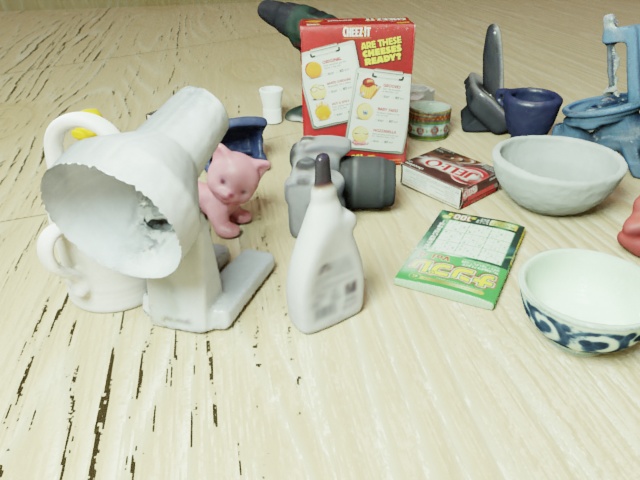}};
        \node[anchor=south west, xshift=-3pt, yshift=-3pt] at (img.south west)
  {\scalebox{0.5}{\begin{tabular}{@{}l@{}}\end{tabular}}};
       \node[anchor=north east, xshift=-10pt, yshift=2pt] at (img.north east)
  {\scalebox{0.7}{\scriptsize\colorbox{gray!30}{\strut$\Delta R=17.6^\circ,\ \Delta t=2.69$$ m$}}};
\end{tikzpicture} &
& 
\begin{tikzpicture}
  \node[inner sep=0pt] (img) at (0,0) {\includegraphics[width=\linewidth]{appendix_figures/target_bop/000951.png}};
  \node[anchor=south west, xshift=-3pt, yshift=-3pt] at (img.south west)
  {\scalebox{0.5}{\colorbox{gray!30}{\begin{tabular}{@{}l@{}} {can000951}\end{tabular}}}};
\end{tikzpicture} &
\begin{tikzpicture}
  \node[inner sep=0pt] (img) at (0,0) {\includegraphics[width=\linewidth]{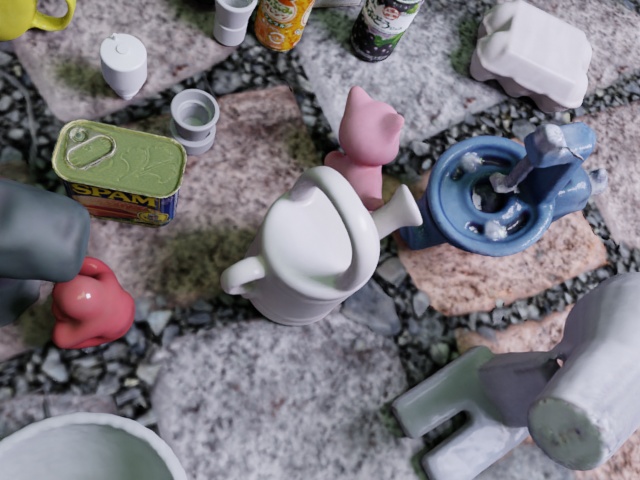}};
  \node[anchor=south west, xshift=-3pt, yshift=-3pt] at (img.south west)
  {\scalebox{0.5}{\begin{tabular}{@{}l@{}}\end{tabular}}};
       \node[anchor=north east, xshift=-10pt, yshift=2pt] at (img.north east)
  {\scalebox{0.7}{\scriptsize\colorbox{gray!30}{\strut$\Delta R=16.8^\circ,\ \Delta t=1.37$$ m$}}};
\end{tikzpicture} \\

\begin{tikzpicture}
  \node[inner sep=0pt] (img) at (0,0) {\includegraphics[width=\linewidth]{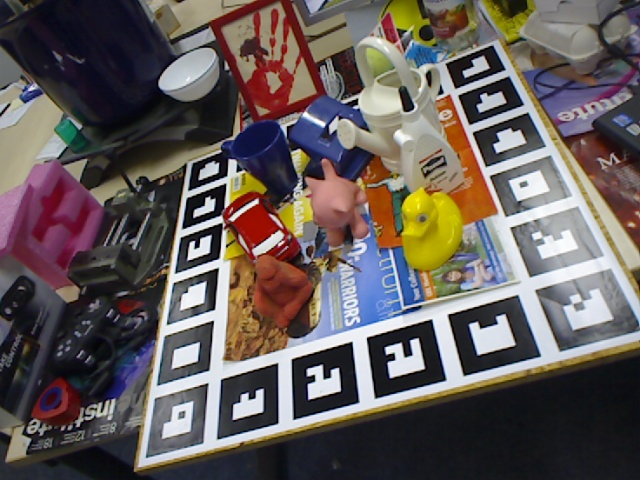}};
  \node[anchor=south west, xshift=-3pt, yshift=-3pt] at (img.south west)
  {\scalebox{0.5}{\colorbox{gray!30}{\begin{tabular}{@{}l@{}} {cat001126}\end{tabular}}}};
\end{tikzpicture} &
\begin{tikzpicture}
  \node[inner sep=0pt] (img) at (0,0) {\includegraphics[width=\linewidth]{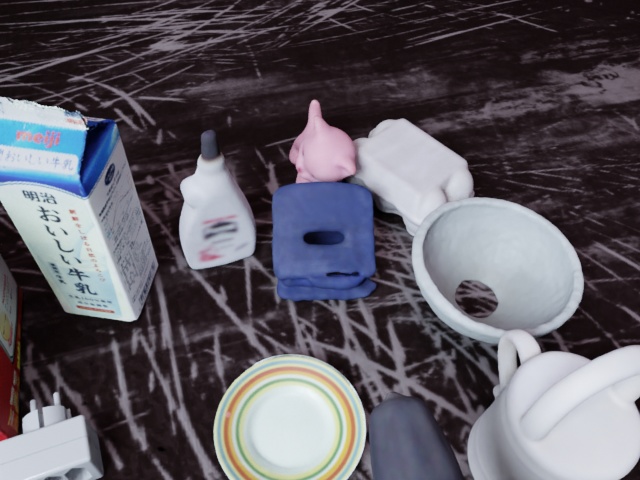}};
    \node[anchor=south west, xshift=-3pt, yshift=-3pt] at (img.south west)
  {\scalebox{0.5}{\begin{tabular}{@{}l@{}}\end{tabular}}};
       \node[anchor=north east, xshift=-10pt, yshift=2pt] at (img.north east)
  {\scalebox{0.7}{\scriptsize\colorbox{gray!30}{\strut$\Delta R=17.65^\circ,\ \Delta t=1.31$$ m$}}};
\end{tikzpicture} &
& 
\begin{tikzpicture}
  \node[inner sep=0pt] (img) at (0,0) {\includegraphics[width=\linewidth]{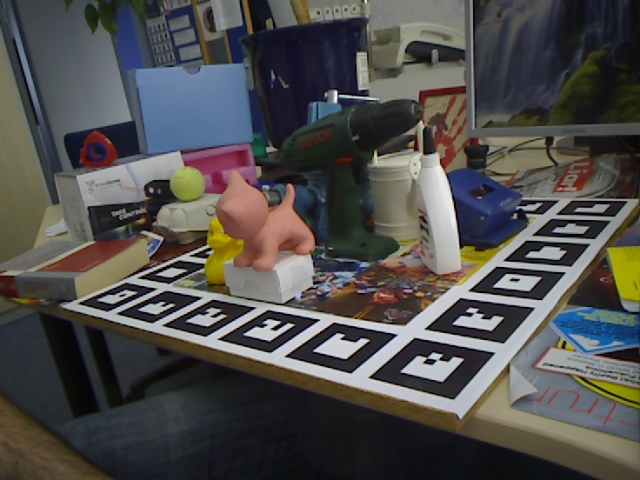}};
 \node[anchor=south west, xshift=-3pt, yshift=-3pt] at (img.south west)
  {\scalebox{0.5}{\colorbox{gray!30}{\begin{tabular}{@{}l@{}} {driller000340}\end{tabular}}}};
\end{tikzpicture} &
\begin{tikzpicture}
  \node[inner sep=0pt] (img) at (0,0) {\includegraphics[width=\linewidth]{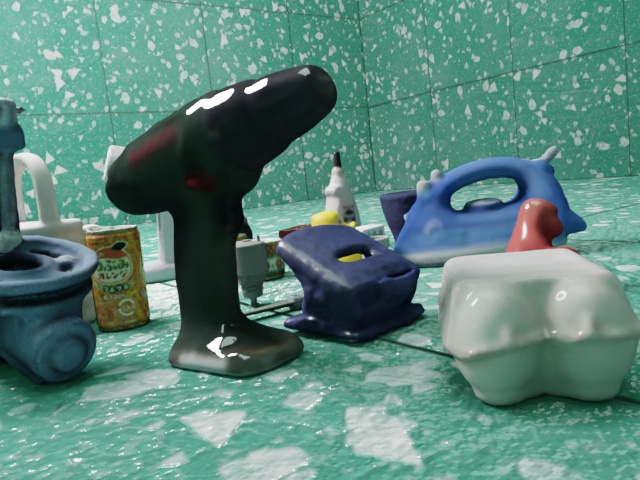}};
     \node[anchor=south west, xshift=-3pt, yshift=-3pt] at (img.south west)
  {\scalebox{0.5}{\begin{tabular}{@{}l@{}}\end{tabular}}};
       \node[anchor=north east, xshift=-10pt, yshift=2pt] at (img.north east)
  {\scalebox{0.7}{\scriptsize\colorbox{gray!30}{\strut$\Delta R=11.3^\circ,\ \Delta t=1.72$$ m$}}};
\end{tikzpicture} \\

\begin{tikzpicture}
  \node[inner sep=0pt] (img) at (0,0) {\includegraphics[width=\linewidth]{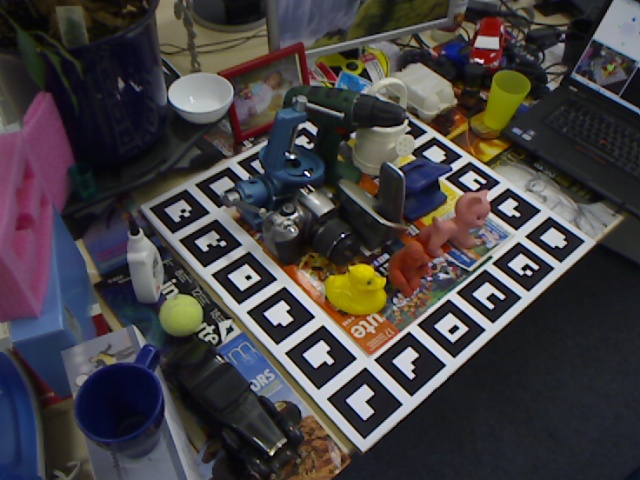}};
 \node[anchor=south west, xshift=-3pt, yshift=-3pt] at (img.south west)
  {\scalebox{0.5}{\colorbox{gray!30}{\begin{tabular}{@{}l@{}} {phone0000504}\end{tabular}}}};
\end{tikzpicture} &
\begin{tikzpicture}
  \node[inner sep=0pt] (img) at (0,0) {\includegraphics[width=\linewidth]{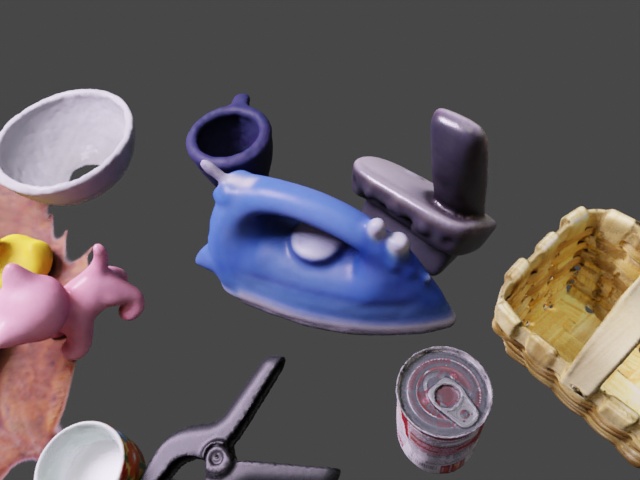}};
   \node[anchor=south west, xshift=-3pt, yshift=-3pt] at (img.south west)
  {\scalebox{0.5}{\begin{tabular}{@{}l@{}}\end{tabular}}};
       \node[anchor=north east, xshift=-10pt, yshift=2pt] at (img.north east)
  {\scalebox{0.7}{\scriptsize\colorbox{gray!30}{\strut$\Delta R=12.56^\circ,\ \Delta t=1.54$$ m$}}};
\end{tikzpicture} &
& 
\begin{tikzpicture}
  \node[inner sep=0pt] (img) at (0,0) {\includegraphics[width=\linewidth]{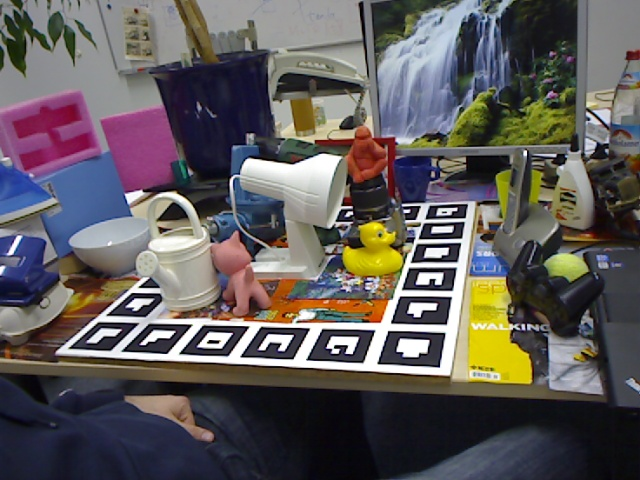}};
 \node[anchor=south west, xshift=-3pt, yshift=-3pt] at (img.south west)
  {\scalebox{0.5}{\colorbox{gray!30}{\begin{tabular}{@{}l@{}} {lamp000629}\end{tabular}}}};
\end{tikzpicture} &
\begin{tikzpicture}
  \node[inner sep=0pt] (img) at (0,0) {\includegraphics[width=\linewidth]{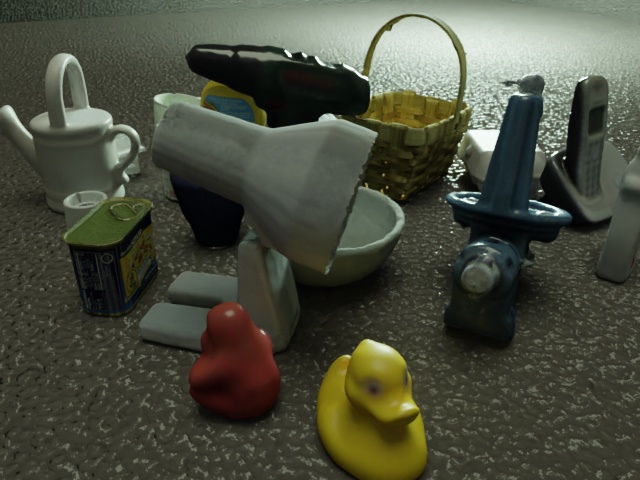}};
  \node[anchor=south west, xshift=-3pt, yshift=-3pt] at (img.south west)
  {\scalebox{0.5}{\begin{tabular}{@{}l@{}}\end{tabular}}};
       \node[anchor=north east, xshift=-10pt, yshift=2pt] at (img.north east)
  {\scalebox{0.7}{\scriptsize\colorbox{gray!30}{\strut$\Delta R=11.22^\circ,\ \Delta t=2.02$$ m$}}};
\end{tikzpicture}  \\

\begin{tikzpicture}
  \node[inner sep=0pt] (img) at (0,0) {\includegraphics[width=\linewidth]{appendix_figures/target_bop/000003.png}};
 \node[anchor=south west, xshift=-3pt, yshift=-3pt] at (img.south west)
  {\scalebox{0.5}{\colorbox{gray!30}{\begin{tabular}{@{}l@{}} {OLM-duck000003}\end{tabular}}}};
\end{tikzpicture} &
\begin{tikzpicture}
  \node[inner sep=0pt] (img) at (0,0) {\includegraphics[width=\linewidth]{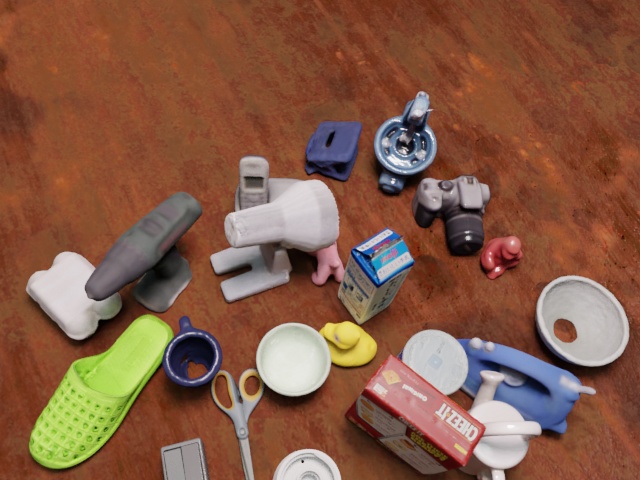}};
  \node[anchor=south west, xshift=-3pt, yshift=-3pt] at (img.south west)
  {\scalebox{0.5}{\begin{tabular}{@{}l@{}}\end{tabular}}};
       \node[anchor=north east, xshift=-10pt, yshift=2pt] at (img.north east)
  {\scalebox{0.7}{\scriptsize\colorbox{gray!30}{\strut$\Delta R=10.03^\circ,\ \Delta t=0.449$$ m$}}};
\end{tikzpicture} &
& 
\begin{tikzpicture}
  \node[inner sep=0pt] (img) at (0,0) {\includegraphics[width=\linewidth]{appendix_figures/target_bop/000003.png}};
\node[anchor=south west, xshift=-3pt, yshift=-3pt] at (img.south west)
  {\scalebox{0.5}{\colorbox{gray!30}{\begin{tabular}{@{}l@{}} {OLM-Holep000003}\end{tabular}}}};
\end{tikzpicture} &
\begin{tikzpicture}
  \node[inner sep=0pt] (img) at (0,0) {\includegraphics[width=\linewidth]{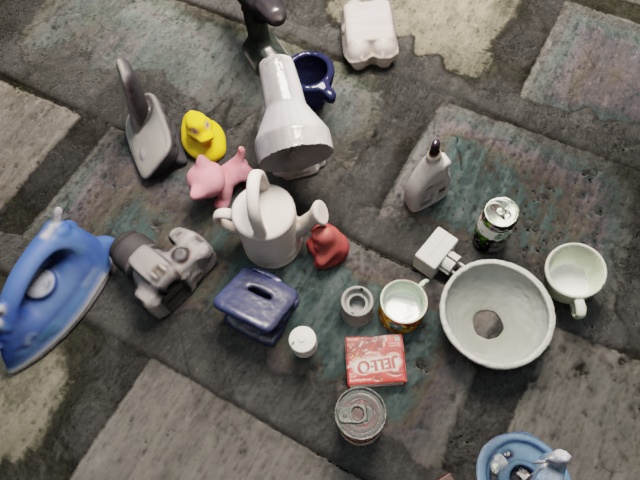}};
  \node[anchor=south west, xshift=-3pt, yshift=-3pt] at (img.south west)
  {\scalebox{0.5}{\begin{tabular}{@{}l@{}}\end{tabular}}};
       \node[anchor=north east, xshift=-10pt, yshift=2pt] at (img.north east)
  {\scalebox{0.7}{\scriptsize\colorbox{gray!30}{\strut$\Delta R=1.39^\circ,\ \Delta t=1.29$$ m$}}};
\end{tikzpicture}  \\

\begin{tikzpicture}
  \node[inner sep=0pt] (img) at (0,0) {\includegraphics[width=\linewidth]{appendix_figures/target_bop/000446.png}};
\node[anchor=south west, xshift=-3pt, yshift=-3pt] at (img.south west)
  {\scalebox{0.5}{\colorbox{gray!30}{\begin{tabular}{@{}l@{}} {OLM-EB000446}\end{tabular}}}};
\end{tikzpicture} &
\begin{tikzpicture}
  \node[inner sep=0pt] (img) at (0,0) {\includegraphics[width=\linewidth]{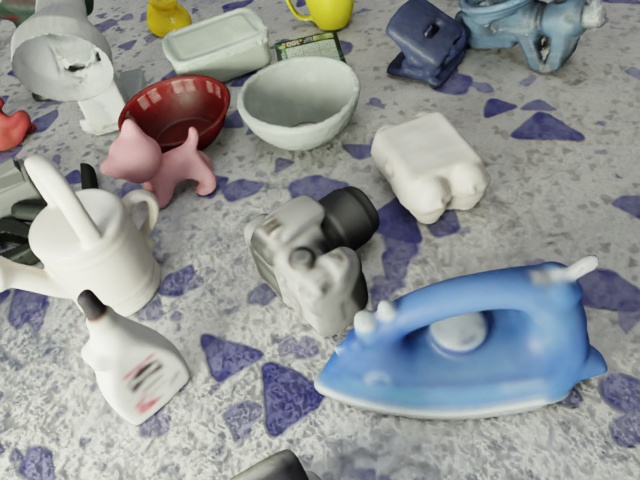}};
  \node[anchor=south west, xshift=-3pt, yshift=-3pt] at (img.south west)
  {\scalebox{0.5}{\begin{tabular}{@{}l@{}}\end{tabular}}};
       \node[anchor=north east, xshift=-10pt, yshift=2pt] at (img.north east)
  {\scalebox{0.7}{\scriptsize\colorbox{gray!30}{\strut$\Delta R=2.28^\circ,\ \Delta t=2.35$$ m$}}};
\end{tikzpicture} &
& 
\begin{tikzpicture}
  \node[inner sep=0pt] (img) at (0,0) {\includegraphics[width=\linewidth]{appendix_figures/target_bop/000178.png}};
 \node[anchor=south west, xshift=-3pt, yshift=-3pt] at (img.south west)
  {\scalebox{0.5}{\colorbox{gray!30}{\begin{tabular}{@{}l@{}} {OLM-glue000178}\end{tabular}}}};
\end{tikzpicture} &
\begin{tikzpicture}
  \node[inner sep=0pt] (img) at (0,0) {\includegraphics[width=\linewidth]{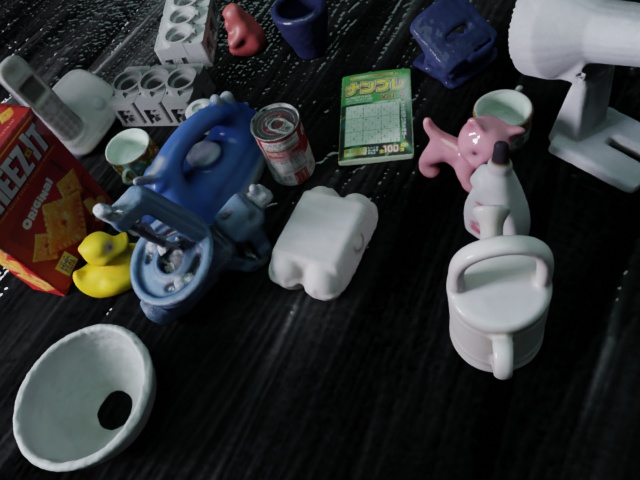}};
  \node[anchor=south west, xshift=-3pt, yshift=-3pt] at (img.south west)
  {\scalebox{0.5}{\begin{tabular}{@{}l@{}}\end{tabular}}};
       \node[anchor=north east, xshift=-10pt, yshift=2pt] at (img.north east)
  {\scalebox{0.7}{\scriptsize\colorbox{gray!30}{\strut$\Delta R=18.56^\circ,\ \Delta t=3.57$$ m$}}};
\end{tikzpicture}  \\

\end{tabular}
\vspace{-1em}
\caption{Examples of target-source image pairs in LineMOD (4 first rows) and Occluded-LineMOD (last 2 rows). Rotation and translation differences ($\Delta R$, $\Delta t$) between each pair are indicated.}
\end{center}
\end{figure*}

%% file: appendix/3.2.Light_pairs.tex
\begin{figure*}[h]
\begin{center}
{\Large \bfseries B. Lightbox Pairs}\par \vspace{0.2em}
\renewcommand{\arraystretch}{0.8}
\begin{tabular}{@{}>{\centering\arraybackslash}m{0.2\textwidth}@{\hspace{3pt}}
                >{\centering\arraybackslash}m{0.2\textwidth}@{\hspace{8pt}}
                >{\centering\arraybackslash}m{0.012\textwidth}@{}
                >{\centering\arraybackslash}m{0.2\textwidth}@{\hspace{3pt}}
                >{\centering\arraybackslash}m{0.2\textwidth}@{}}

\textbf{Target} & \textbf{Source} & & \textbf{Target} & \textbf{Source} \\

\begin{tikzpicture}
  \node[inner sep=0pt] (img) at (0,0) {\includegraphics[width=\linewidth]{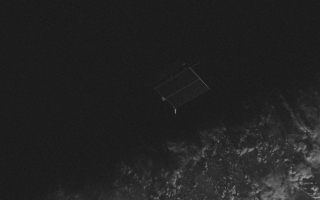}};
    \node[anchor=south west, xshift=-3pt, yshift=-3pt] at (img.south west)
      {\scalebox{0.5}{\colorbox{gray!30}{\begin{tabular}{@{}l@{}} {img000003}\end{tabular}}}};
\end{tikzpicture} &
\begin{tikzpicture}
  \node[inner sep=0pt] (img) at (0,0) {\includegraphics[width=\linewidth]{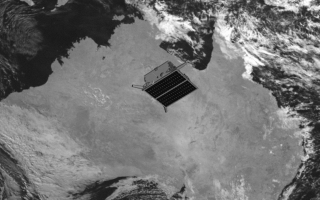}};
  \node[anchor=south west, xshift=-3pt, yshift=-3pt] at (img.south west)
  {\scalebox{0.5}{\colorbox{gray!30}{\begin{tabular}{@{}l@{}} {img039394}\end{tabular}}}};
  \node[anchor=north east, xshift=-10pt, yshift=2pt] at (img.north east)
  {\scalebox{0.7}{\scriptsize\colorbox{gray!30}{\strut$\Delta R=3.91^\circ,\ \Delta t=0.284$$ m$}}};
\end{tikzpicture} &
& 
\begin{tikzpicture}
  \node[inner sep=0pt] (img) at (0,0) {\includegraphics[width=\linewidth]{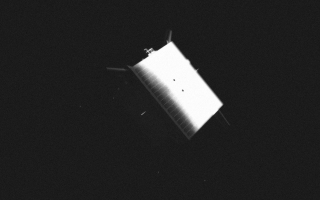}};
  \node[anchor=south west, xshift=-3pt, yshift=-3pt] at (img.south west)
  {\scalebox{0.5}{\colorbox{gray!30}{\begin{tabular}{@{}l@{}} {img000005}\end{tabular}}}};
 
\end{tikzpicture} &
\begin{tikzpicture}
  \node[inner sep=0pt] (img) at (0,0) {\includegraphics[width=\linewidth]{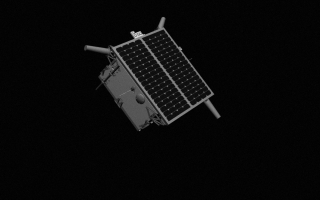}};
  \node[anchor=south west, xshift=-3pt, yshift=-3pt] at (img.south west)
  {\scalebox{0.5}{\colorbox{gray!30}{\begin{tabular}{@{}l@{}} {img007013}\end{tabular}}}};
    \node[anchor=north east, xshift=-10pt, yshift=2pt] at (img.north east)
  {\scalebox{0.7}{\scriptsize\colorbox{gray!30}{\strut$\Delta R=6.20^\circ,\ \Delta t=0.228$$ m$}}};
\end{tikzpicture}  \\

\begin{tikzpicture}
  \node[inner sep=0pt] (img) at (0,0) {\includegraphics[width=\linewidth]{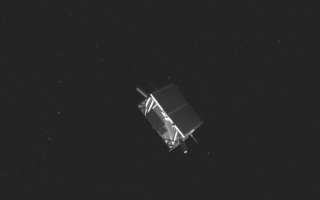}};
  \node[anchor=south west, xshift=-3pt, yshift=-3pt] at (img.south west)
  {\scalebox{0.5}{\colorbox{gray!30}{\begin{tabular}{@{}l@{}} {img000010}\end{tabular}}}};
\end{tikzpicture} &
\begin{tikzpicture}
  \node[inner sep=0pt] (img) at (0,0) {\includegraphics[width=\linewidth]{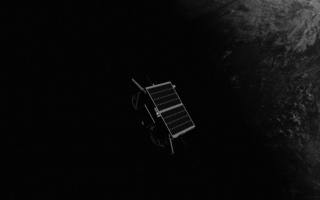}};
   \node[anchor=south west, xshift=-3pt, yshift=-3pt] at (img.south west)
  {\scalebox{0.5}{\colorbox{gray!30}{\begin{tabular}{@{}l@{}} {img055832}\end{tabular}}}};
     \node[anchor=north east, xshift=-10pt, yshift=2pt] at (img.north east)
  {\scalebox{0.7}{\scriptsize\colorbox{gray!30}{\strut$\Delta R=8.71^\circ,\ \Delta t=0.361$$ m$}}};
\end{tikzpicture} &
& 
\begin{tikzpicture}
  \node[inner sep=0pt] (img) at (0,0) {\includegraphics[width=\linewidth]{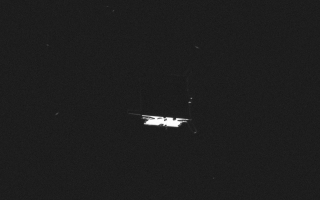}};
   \node[anchor=south west, xshift=-3pt, yshift=-3pt] at (img.south west)
  {\scalebox{0.5}{\colorbox{gray!30}{\begin{tabular}{@{}l@{}} {img000020}\end{tabular}}}};
 
\end{tikzpicture} &
\begin{tikzpicture}
  \node[inner sep=0pt] (img) at (0,0) {\includegraphics[width=\linewidth]{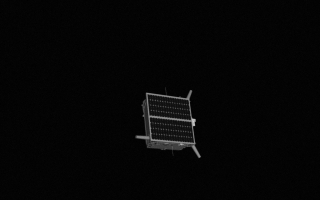}};
   \node[anchor=south west, xshift=-3pt, yshift=-3pt] at (img.south west)
  {\scalebox{0.5}{\colorbox{gray!30}{\begin{tabular}{@{}l@{}} {img004702}\end{tabular}}}};
     \node[anchor=north east, xshift=-10pt, yshift=2pt] at (img.north east)
  {\scalebox{0.7}{\scriptsize\colorbox{gray!30}{\strut$\Delta R=12.28^\circ,\ \Delta t=0.476$$ m$}}};
\end{tikzpicture} \\

\begin{tikzpicture}
  \node[inner sep=0pt] (img) at (0,0) {\includegraphics[width=\linewidth]{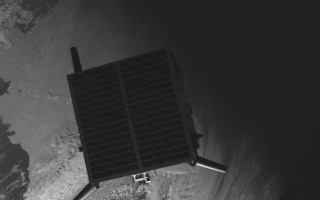}};
    \node[anchor=south west, xshift=-3pt, yshift=-3pt] at (img.south west)
  {\scalebox{0.5}{\colorbox{gray!30}{\begin{tabular}{@{}l@{}} {img000030}\end{tabular}}}};
\end{tikzpicture} &
\begin{tikzpicture}
  \node[inner sep=0pt] (img) at (0,0) {\includegraphics[width=\linewidth]{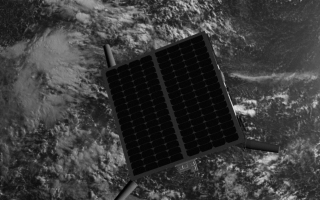}};
     \node[anchor=south west, xshift=-3pt, yshift=-3pt] at (img.south west)
  {\scalebox{0.5}{\colorbox{gray!30}{\begin{tabular}{@{}l@{}} {img036817}\end{tabular}}}};
     \node[anchor=north east, xshift=-10pt, yshift=2pt] at (img.north east)
  {\scalebox{0.7}{\scriptsize\colorbox{gray!30}{\strut$\Delta R=12.92^\circ,\ \Delta t=0.332$$ m$}}};
\end{tikzpicture} &
& 
\begin{tikzpicture}
  \node[inner sep=0pt] (img) at (0,0) {\includegraphics[width=\linewidth]{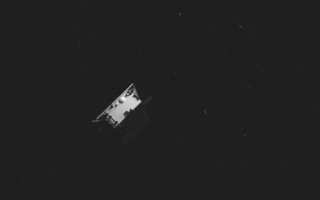}};
  \node[anchor=south west, xshift=-3pt, yshift=-3pt] at (img.south west)
  {\scalebox{0.5}{\colorbox{gray!30}{\begin{tabular}{@{}l@{}} {img000050}\end{tabular}}}};
\end{tikzpicture} &
\begin{tikzpicture}
  \node[inner sep=0pt] (img) at (0,0) {\includegraphics[width=\linewidth]{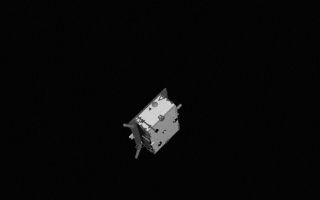}};
       \node[anchor=south west, xshift=-3pt, yshift=-3pt] at (img.south west)
  {\scalebox{0.5}{\colorbox{gray!30}{\begin{tabular}{@{}l@{}} {img017246}\end{tabular}}}};
     \node[anchor=north east, xshift=-10pt, yshift=2pt] at (img.north east)
  {\scalebox{0.7}{\scriptsize\colorbox{gray!30}{\strut$\Delta R=5.60^\circ,\ \Delta t=0.567$$ m$}}};
\end{tikzpicture} \\

\begin{tikzpicture}
  \node[inner sep=0pt] (img) at (0,0) {\includegraphics[width=\linewidth]{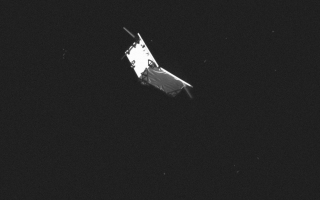}};
       \node[anchor=south west, xshift=-3pt, yshift=-3pt] at (img.south west)
  {\scalebox{0.5}{\colorbox{gray!30}{\begin{tabular}{@{}l@{}} {img000100}\end{tabular}}}};

\end{tikzpicture} &
\begin{tikzpicture}
  \node[inner sep=0pt] (img) at (0,0) {\includegraphics[width=\linewidth]{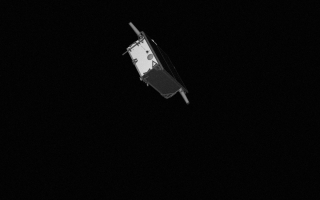}};
        \node[anchor=south west, xshift=-3pt, yshift=-3pt] at (img.south west)
  {\scalebox{0.5}{\colorbox{gray!30}{\begin{tabular}{@{}l@{}} {img011962}\end{tabular}}}};
     \node[anchor=north east, xshift=-10pt, yshift=2pt] at (img.north east)
  {\scalebox{0.7}{\scriptsize\colorbox{gray!30}{\strut$\Delta R=12.16^\circ,\ \Delta t=0.044$$ m$}}};
\end{tikzpicture} &
& 
\begin{tikzpicture}
  \node[inner sep=0pt] (img) at (0,0) {\includegraphics[width=\linewidth]{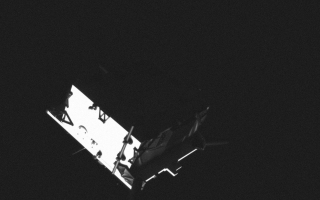}};
     \node[anchor=south west, xshift=-3pt, yshift=-3pt] at (img.south west)
  {\scalebox{0.5}{\colorbox{gray!30}{\begin{tabular}{@{}l@{}} {img000203}\end{tabular}}}};

\end{tikzpicture} &
\begin{tikzpicture}
  \node[inner sep=0pt] (img) at (0,0) {\includegraphics[width=\linewidth]{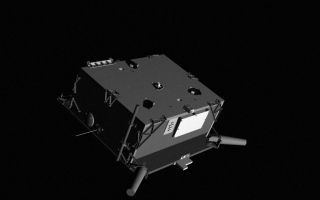}};
     \node[anchor=south west, xshift=-3pt, yshift=-3pt] at (img.south west)
  {\scalebox{0.5}{\colorbox{gray!30}{\begin{tabular}{@{}l@{}} {img010615}\end{tabular}}}};
     \node[anchor=north east, xshift=-10pt, yshift=2pt] at (img.north east)
  {\scalebox{0.7}{\scriptsize\colorbox{gray!30}{\strut$\Delta R=7.80^\circ,\ \Delta t=0.138$$ m$}}};
\end{tikzpicture}  \\

\end{tabular}
\vspace{-0.3cm}
\caption{Cross-domain pairs for lightbox domain.  Rotation and translation differences ($\Delta R$, $\Delta t$) between each pair are indicated.}
\end{center}
\end{figure*}

%% file: appendix/3.3.Sun_pairs.tex
\begin{figure*}[h]
\begin{center}
{\Large \bfseries C. Sunlamp Pairs}\par \vspace{0.4em}

\renewcommand{\arraystretch}{0.8}
\begin{tabular}{@{}>{\centering\arraybackslash}m{0.2\textwidth}@{\hspace{3pt}}
                >{\centering\arraybackslash}m{0.2\textwidth}@{\hspace{8pt}}
                >{\centering\arraybackslash}m{0.012\textwidth}@{}
                >{\centering\arraybackslash}m{0.2\textwidth}@{\hspace{3pt}}
                >{\centering\arraybackslash}m{0.2\textwidth}@{}}

\textbf{Target} & \textbf{Source} & & \textbf{Target} & \textbf{Source} \\

\begin{tikzpicture}
  \node[inner sep=0pt] (img) at (0,0) {\includegraphics[width=\linewidth]{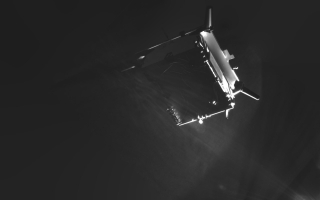}};
     \node[anchor=south west, xshift=-3pt, yshift=-3pt] at (img.south west)
  {\scalebox{0.5}{\colorbox{gray!30}{\begin{tabular}{@{}l@{}} {img000020}\end{tabular}}}};
\end{tikzpicture} &
\begin{tikzpicture}
  \node[inner sep=0pt] (img) at (0,0) {\includegraphics[width=\linewidth]{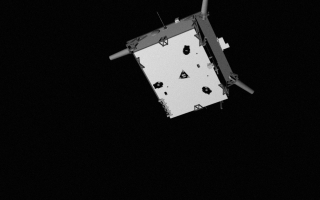}};
      \node[anchor=south west, xshift=-3pt, yshift=-3pt] at (img.south west)
  {\scalebox{0.5}{\colorbox{gray!30}{\begin{tabular}{@{}l@{}} {img015846}\end{tabular}}}};
     \node[anchor=north east, xshift=-10pt, yshift=2pt] at (img.north east)
  {\scalebox{0.7}{\scriptsize\colorbox{gray!30}{\strut$\Delta R=6.65^\circ,\ \Delta t=0.267$$ m$}}};
\end{tikzpicture} &
& 
\begin{tikzpicture}
  \node[inner sep=0pt] (img) at (0,0) {\includegraphics[width=\linewidth]{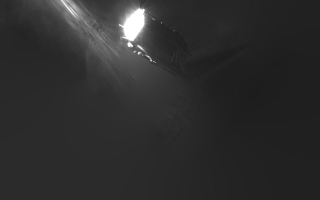}};
 \node[anchor=south west, xshift=-3pt, yshift=-3pt] at (img.south west)
  {\scalebox{0.5}{\colorbox{gray!30}{\begin{tabular}{@{}l@{}} {img000030}\end{tabular}}}};
\end{tikzpicture} &
\begin{tikzpicture}
  \node[inner sep=0pt] (img) at (0,0) {\includegraphics[width=\linewidth]{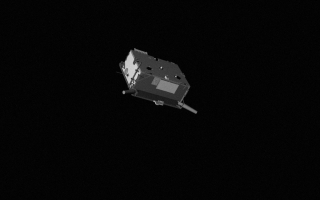}};
       \node[anchor=south west, xshift=-3pt, yshift=-3pt] at (img.south west)
  {\scalebox{0.5}{\colorbox{gray!30}{\begin{tabular}{@{}l@{}} {img014255}\end{tabular}}}};
     \node[anchor=north east, xshift=-10pt, yshift=2pt] at (img.north east)
  {\scalebox{0.7}{\scriptsize\colorbox{gray!30}{\strut$\Delta R=14.84^\circ,\ \Delta t=0.497$$ m$}}};
\end{tikzpicture}  \\

\begin{tikzpicture}
  \node[inner sep=0pt] (img) at (0,0) {\includegraphics[width=\linewidth]{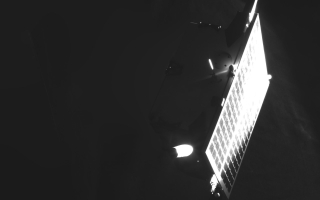}};
 \node[anchor=south west, xshift=-3pt, yshift=-3pt] at (img.south west)
  {\scalebox{0.5}{\colorbox{gray!30}{\begin{tabular}{@{}l@{}} {img000050}\end{tabular}}}};
\end{tikzpicture} &
\begin{tikzpicture}
  \node[inner sep=0pt] (img) at (0,0) {\includegraphics[width=\linewidth]{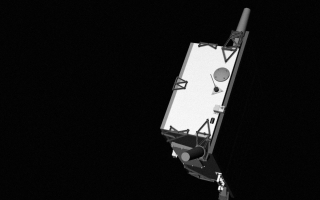}};
    \node[anchor=south west, xshift=-3pt, yshift=-3pt] at (img.south west)
  {\scalebox{0.5}{\colorbox{gray!30}{\begin{tabular}{@{}l@{}} {img024343}\end{tabular}}}};
     \node[anchor=north east, xshift=-10pt, yshift=2pt] at (img.north east)
  {\scalebox{0.7}{\scriptsize\colorbox{gray!30}{\strut$\Delta R=4.34^\circ,\ \Delta t=0.387$$ m$}}};
\end{tikzpicture} &
& 
\begin{tikzpicture}
  \node[inner sep=0pt] (img) at (0,0) {\includegraphics[width=\linewidth]{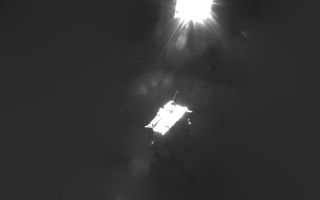}};
    \node[anchor=south west, xshift=-3pt, yshift=-3pt] at (img.south west)
  {\scalebox{0.5}{\colorbox{gray!30}{\begin{tabular}{@{}l@{}} {img001400}\end{tabular}}}};
\end{tikzpicture} &
\begin{tikzpicture}
  \node[inner sep=0pt] (img) at (0,0) {\includegraphics[width=\linewidth]{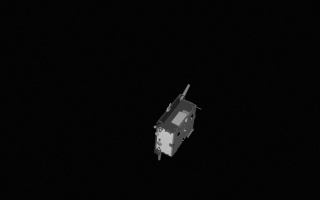}};
   \node[anchor=south west, xshift=-3pt, yshift=-3pt] at (img.south west)
  {\scalebox{0.5}{\colorbox{gray!30}{\begin{tabular}{@{}l@{}} {img029105}\end{tabular}}}};
     \node[anchor=north east, xshift=-10pt, yshift=2pt] at (img.north east)
  {\scalebox{0.7}{\scriptsize\colorbox{gray!30}{\strut$\Delta R=18.37^\circ,\ \Delta t=0.678$$ m$}}};
\end{tikzpicture} \\

\begin{tikzpicture}
  \node[inner sep=0pt] (img) at (0,0) {\includegraphics[width=\linewidth]{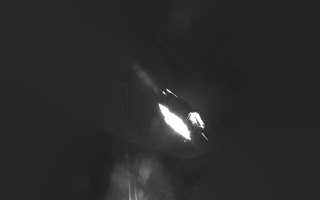}};
 \node[anchor=south west, xshift=-3pt, yshift=-3pt] at (img.south west)
  {\scalebox{0.5}{\colorbox{gray!30}{\begin{tabular}{@{}l@{}} {img001518}\end{tabular}}}};
\end{tikzpicture} &
\begin{tikzpicture}
  \node[inner sep=0pt] (img) at (0,0) {\includegraphics[width=\linewidth]{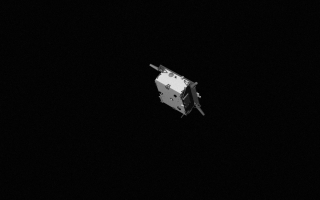}};
 \node[anchor=south west, xshift=-3pt, yshift=-3pt] at (img.south west)
  {\scalebox{0.5}{\colorbox{gray!30}{\begin{tabular}{@{}l@{}} {img020529}\end{tabular}}}};
     \node[anchor=north east, xshift=-10pt, yshift=2pt] at (img.north east)
  {\scalebox{0.7}{\scriptsize\colorbox{gray!30}{\strut$\Delta R=8.67^\circ,\ \Delta t=0.439$$ m$}}};
\end{tikzpicture} &
& 
\begin{tikzpicture}
  \node[inner sep=0pt] (img) at (0,0) {\includegraphics[width=\linewidth]{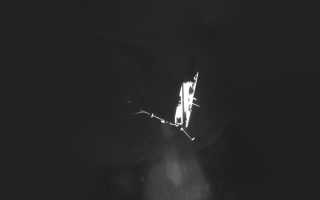}};
   \node[anchor=south west, xshift=-3pt, yshift=-3pt] at (img.south west)
  {\scalebox{0.5}{\colorbox{gray!30}{\begin{tabular}{@{}l@{}} {img000001}\end{tabular}}}};
\end{tikzpicture} &
\begin{tikzpicture}
  \node[inner sep=0pt] (img) at (0,0) {\includegraphics[width=\linewidth]{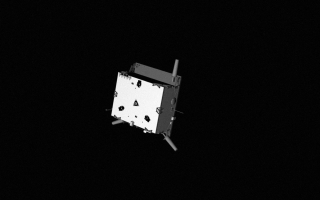}};
   \node[anchor=south west, xshift=-3pt, yshift=-3pt] at (img.south west)
  {\scalebox{0.5}{\colorbox{gray!30}{\begin{tabular}{@{}l@{}} {img007660}\end{tabular}}}};
     \node[anchor=north east, xshift=-10pt, yshift=2pt] at (img.north east)
  {\scalebox{0.7}{\scriptsize\colorbox{gray!30}{\strut$\Delta R=3.98^\circ,\ \Delta t=0.573$$ m$}}};
\end{tikzpicture} \\

\begin{tikzpicture}
  \node[inner sep=0pt] (img) at (0,0) {\includegraphics[width=\linewidth]{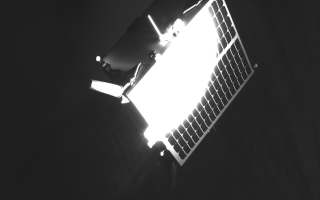}};
 \node[anchor=south west, xshift=-3pt, yshift=-3pt] at (img.south west)
  {\scalebox{0.5}{\colorbox{gray!30}{\begin{tabular}{@{}l@{}} {img000090}\end{tabular}}}};
\end{tikzpicture} &
\begin{tikzpicture}
  \node[inner sep=0pt] (img) at (0,0) {\includegraphics[width=\linewidth]{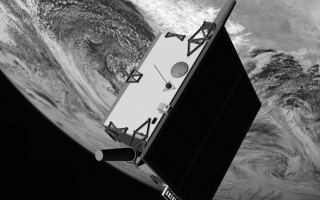}};
   \node[anchor=south west, xshift=-3pt, yshift=-3pt] at (img.south west)
  {\scalebox{0.5}{\colorbox{gray!30}{\begin{tabular}{@{}l@{}} {img039064}\end{tabular}}}};
     \node[anchor=north east, xshift=-10pt, yshift=2pt] at (img.north east)
  {\scalebox{0.7}{\scriptsize\colorbox{gray!30}{\strut$\Delta R=9.11^\circ,\ \Delta t=0.402$$ m$}}};
\end{tikzpicture} &
& 
\begin{tikzpicture}
  \node[inner sep=0pt] (img) at (0,0) {\includegraphics[width=\linewidth]{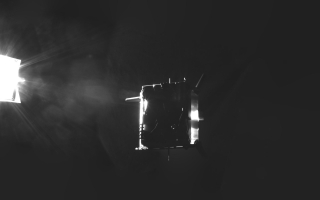}};
 \node[anchor=south west, xshift=-3pt, yshift=-3pt] at (img.south west)
  {\scalebox{0.5}{\colorbox{gray!30}{\begin{tabular}{@{}l@{}} {img000101}\end{tabular}}}};
\end{tikzpicture} &
\begin{tikzpicture}
  \node[inner sep=0pt] (img) at (0,0) {\includegraphics[width=\linewidth]{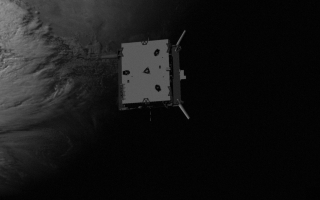}};
    \node[anchor=south west, xshift=-3pt, yshift=-3pt] at (img.south west)
  {\scalebox{0.5}{\colorbox{gray!30}{\begin{tabular}{@{}l@{}} {img040702}\end{tabular}}}};
     \node[anchor=north east, xshift=-10pt, yshift=2pt] at (img.north east)
  {\scalebox{0.7}{\scriptsize\colorbox{gray!30}{\strut$\Delta R=11.09^\circ,\ \Delta t=0.599$$ m$}}};
\end{tikzpicture}  \\

\end{tabular}
\vspace{-0.3cm}
\caption{Cross-domain pairs from Sunlamp. Rotation and translation differences ($\Delta R$, $\Delta t$) between each pair are indicated.}
\end{center}
\end{figure*}

%% file: appendix/3.6.viz_bop.tex
\begin{figure*}[h]
\begin{center}
{\Large \bfseries D. Improvements on LineMOD}\par \vspace{0.4em}

\renewcommand{\arraystretch}{0.8}
\begin{tabular}{@{}>{\centering\arraybackslash}m{0.225\textwidth}@{\hspace{3pt}}
                >{\centering\arraybackslash}m{0.225\textwidth}@{\hspace{8pt}}
                >{\centering\arraybackslash}m{0.012\textwidth}@{}
                >{\centering\arraybackslash}m{0.225\textwidth}@{\hspace{3pt}}
                >{\centering\arraybackslash}m{0.225\textwidth}@{}}

w/o UDA & \acronym & & w/o UDA & \acronym \\

\begin{tikzpicture}
  \node[inner sep=0pt] (img) at (0,0) {\includegraphics[width=\linewidth]{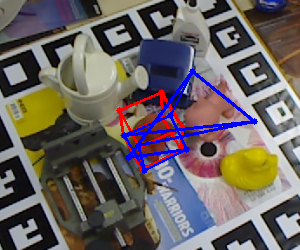}};
  \node[anchor=south west, xshift=-3pt, yshift=-3pt] at (img.south west)
  {\scalebox{0.6}{\colorbox{gray!30}{\begin{tabular}{@{}l@{}} {Ape000033}\end{tabular}}}};
\end{tikzpicture} &
\begin{tikzpicture}
  \node[inner sep=0pt] (img) at (0,0) {\includegraphics[width=\linewidth]{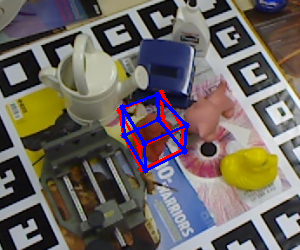}};
  \node[anchor=south west, xshift=-3pt, yshift=-3pt] at (img.south west)
  {\scalebox{0.6}{\begin{tabular}{@{}l@{}} {}\end{tabular}}};
\end{tikzpicture} &
& 
\begin{tikzpicture}
  \node[inner sep=0pt] (img) at (0,0) {\includegraphics[width=\linewidth]{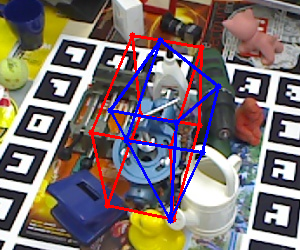}};
  \node[anchor=south west, xshift=-3pt, yshift=-3pt] at (img.south west)
  {\scalebox{0.6}{\colorbox{gray!30}{\begin{tabular}{@{}l@{}} {benchv000012}\end{tabular}}}};
\end{tikzpicture} &
\begin{tikzpicture}
  \node[inner sep=0pt] (img) at (0,0) {\includegraphics[width=\linewidth]{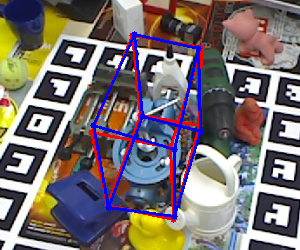}};
 \node[anchor=south west, xshift=-3pt, yshift=-3pt] at (img.south west)
  {\scalebox{0.6}{\begin{tabular}{@{}l@{}} {}\end{tabular}}};
\end{tikzpicture}  \\

\begin{tikzpicture}
  \node[inner sep=0pt] (img) at (0,0) {\includegraphics[width=\linewidth]{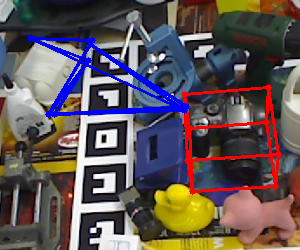}};
 \node[anchor=south west, xshift=-3pt, yshift=-3pt] at (img.south west)
  {\scalebox{0.6}{\colorbox{gray!30}{\begin{tabular}{@{}l@{}} {cam000001}\end{tabular}}}};
\end{tikzpicture} &
\begin{tikzpicture}
  \node[inner sep=0pt] (img) at (0,0) {\includegraphics[width=\linewidth]{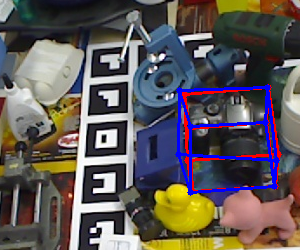}};
 \node[anchor=south west, xshift=-3pt, yshift=-3pt] at (img.south west)
  {\scalebox{0.6}{\begin{tabular}{@{}l@{}} {}\end{tabular}}};
\end{tikzpicture} &
& 
\begin{tikzpicture}
  \node[inner sep=0pt] (img) at (0,0) {\includegraphics[width=\linewidth]{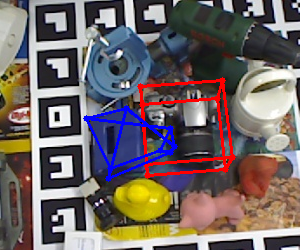}};
  \node[anchor=south west, xshift=-3pt, yshift=-3pt] at (img.south west)
  {\scalebox{0.6}{\colorbox{gray!30}{\begin{tabular}{@{}l@{}} {cam000006}\end{tabular}}}};
 
\end{tikzpicture} &
\begin{tikzpicture}
  \node[inner sep=0pt] (img) at (0,0) {\includegraphics[width=\linewidth]{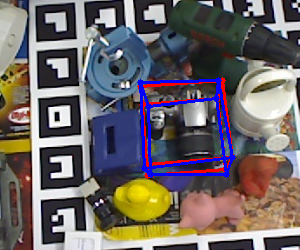}};
\node[anchor=south west, xshift=-3pt, yshift=-3pt] at (img.south west)
  {\scalebox{0.6}{\begin{tabular}{@{}l@{}} {}\end{tabular}}};
\end{tikzpicture} \\

\begin{tikzpicture}
  \node[inner sep=0pt] (img) at (0,0) {\includegraphics[width=\linewidth]{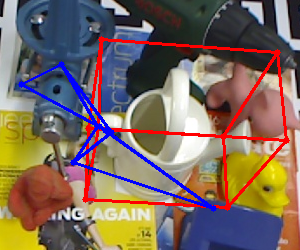}};
 \node[anchor=south west, xshift=-3pt, yshift=-3pt] at (img.south west)
  {\scalebox{0.6}{\colorbox{gray!30}{\begin{tabular}{@{}l@{}} {can001028}\end{tabular}}}};
\end{tikzpicture} &
\begin{tikzpicture}
  \node[inner sep=0pt] (img) at (0,0) {\includegraphics[width=\linewidth]{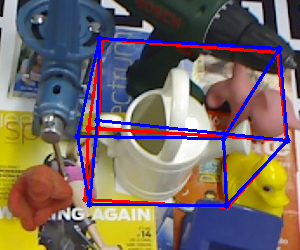}};
\node[anchor=south west, xshift=-3pt, yshift=-3pt] at (img.south west)
  {\scalebox{0.6}{\begin{tabular}{@{}l@{}} {}\end{tabular}}};
\end{tikzpicture} &
& 
\begin{tikzpicture}
  \node[inner sep=0pt] (img) at (0,0) {\includegraphics[width=\linewidth]{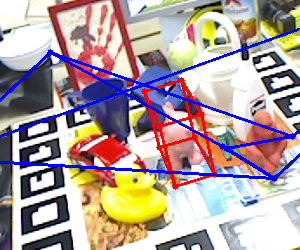}};
  \node[anchor=south west, xshift=-3pt, yshift=-3pt] at (img.south west)
  {\scalebox{0.6}{\colorbox{gray!30}{\begin{tabular}{@{}l@{}} {cat001059}\end{tabular}}}};
\end{tikzpicture} &
\begin{tikzpicture}
  \node[inner sep=0pt] (img) at (0,0) {\includegraphics[width=\linewidth]{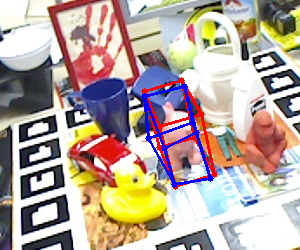}};
\node[anchor=south west, xshift=-3pt, yshift=-3pt] at (img.south west)
  {\scalebox{0.6}{\begin{tabular}{@{}l@{}} {}\end{tabular}}};
\end{tikzpicture} \\

\begin{tikzpicture}
  \node[inner sep=0pt] (img) at (0,0) {\includegraphics[width=\linewidth]{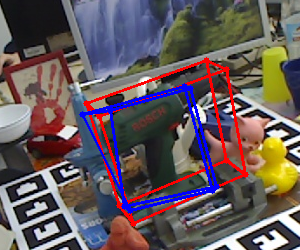}};
 \node[anchor=south west, xshift=-3pt, yshift=-3pt] at (img.south west)
  {\scalebox{0.6}{\colorbox{gray!30}{\begin{tabular}{@{}l@{}} {driller000096}\end{tabular}}}};
\end{tikzpicture} &
\begin{tikzpicture}
  \node[inner sep=0pt] (img) at (0,0) {\includegraphics[width=\linewidth]{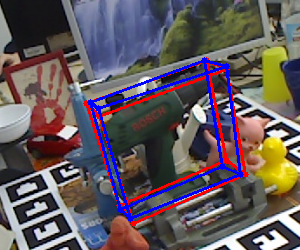}};
\node[anchor=south west, xshift=-3pt, yshift=-3pt] at (img.south west)
  {\scalebox{0.6}{\begin{tabular}{@{}l@{}} {}\end{tabular}}};
\end{tikzpicture} &
& 
\begin{tikzpicture}
  \node[inner sep=0pt] (img) at (0,0) {\includegraphics[width=\linewidth]{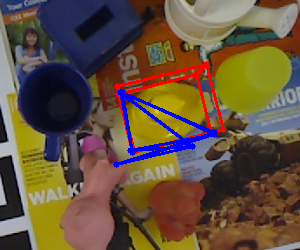}};
  \node[anchor=south west, xshift=-3pt, yshift=-3pt] at (img.south west)
  {\scalebox{0.6}{\colorbox{gray!30}{\begin{tabular}{@{}l@{}} {duck001027}\end{tabular}}}};
\end{tikzpicture} &
\begin{tikzpicture}
  \node[inner sep=0pt] (img) at (0,0) {\includegraphics[width=\linewidth]{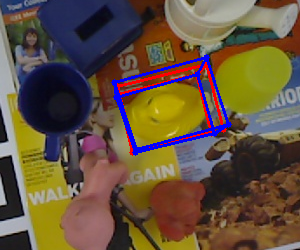}};
 \node[anchor=south west, xshift=-3pt, yshift=-3pt] at (img.south west)
  {\scalebox{0.6}{\begin{tabular}{@{}l@{}} {}\end{tabular}}};
\end{tikzpicture}  \\

\begin{tikzpicture}
  \node[inner sep=0pt] (img) at (0,0) {\includegraphics[width=\linewidth]{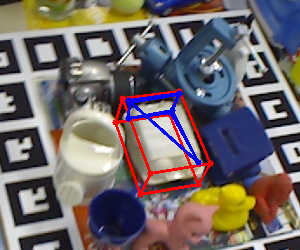}};
  \node[anchor=south west, xshift=-3pt, yshift=-3pt] at (img.south west)
  {\scalebox{0.6}{\colorbox{gray!30}{\begin{tabular}{@{}l@{}} {eggbox000101}\end{tabular}}}};

\end{tikzpicture} &
\begin{tikzpicture}
  \node[inner sep=0pt] (img) at (0,0) {\includegraphics[width=\linewidth]{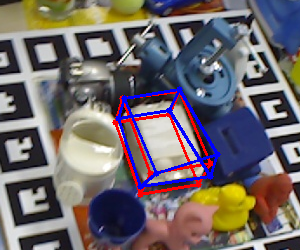}};
 \node[anchor=south west, xshift=-3pt, yshift=-3pt] at (img.south west)
  {\scalebox{0.6}{\begin{tabular}{@{}l@{}} {}\end{tabular}}};
\end{tikzpicture} &
& 
\begin{tikzpicture}
  \node[inner sep=0pt] (img) at (0,0) {\includegraphics[width=\linewidth]{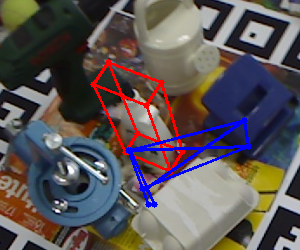}};
  \node[anchor=south west, xshift=-3pt, yshift=-3pt] at (img.south west)
  {\scalebox{0.6}{\colorbox{gray!30}{\begin{tabular}{@{}l@{}} {glue000066}\end{tabular}}}};

\end{tikzpicture} &
\begin{tikzpicture}
  \node[inner sep=0pt] (img) at (0,0) {\includegraphics[width=\linewidth]{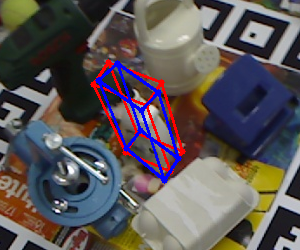}};
\node[anchor=south west, xshift=-3pt, yshift=-3pt] at (img.south west)
  {\scalebox{0.6}{\begin{tabular}{@{}l@{}} {}\end{tabular}}};
\end{tikzpicture}  \\

% \begin{tikzpicture}
%   \node[inner sep=0pt] (img) at (0,0) {\includegraphics[width=\linewidth]{appendix_figures/bop_base/holep_000352.png}};
%   \node[anchor=south west, xshift=-3pt, yshift=-3pt] at (img.south west)
%   {\scalebox{0.6}{\colorbox{gray!30}{\begin{tabular}{@{}l@{}} {holep000352}\end{tabular}}}};
% \end{tikzpicture} &
% \begin{tikzpicture}
%   \node[inner sep=0pt] (img) at (0,0) {\includegraphics[width=\linewidth]{appendix_figures/bop_caplr/holep_000352.png}};
%  \node[anchor=south west, xshift=-3pt, yshift=-3pt] at (img.south west)
%   {\scalebox{0.6}{\begin{tabular}{@{}l@{}} {}\end{tabular}
%   }};
% \end{tikzpicture} &
% & 
% \begin{tikzpicture}
%   \node[inner sep=0pt] (img) at (0,0) {\includegraphics[width=\linewidth]{appendix_figures/bop_base/iron_000122.png}};
% \node[anchor=south west, xshift=-3pt, yshift=-3pt] at (img.south west)
%   {\scalebox{0.6}{\colorbox{gray!30}{\begin{tabular}{@{}l@{}} {iron000122}\end{tabular}}}};
% \end{tikzpicture} &
% \begin{tikzpicture}
%   \node[inner sep=0pt] (img) at (0,0) {\includegraphics[width=\linewidth]{appendix_figures/bop_caplr/iron_000122.png}};
%  \node[anchor=south west, xshift=-3pt, yshift=-3pt] at (img.south west)
%   {\scalebox{0.6}{\begin{tabular}{@{}l@{}} {}\end{tabular}}};
% \end{tikzpicture}  \\

\begin{tikzpicture}
  \node[inner sep=0pt] (img) at (0,0) {\includegraphics[width=\linewidth]{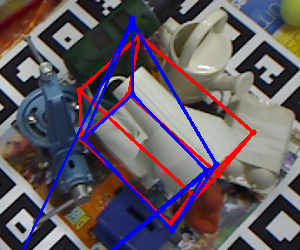}};
\node[anchor=south west, xshift=-3pt, yshift=-3pt] at (img.south west)
  {\scalebox{0.6}{\colorbox{gray!30}{\begin{tabular}{@{}l@{}} {lamp000095}\end{tabular}}}};
\end{tikzpicture} &
\begin{tikzpicture}
  \node[inner sep=0pt] (img) at (0,0) {\includegraphics[width=\linewidth]{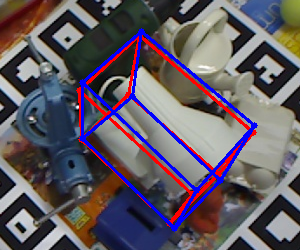}};
 \node[anchor=south west, xshift=-3pt, yshift=-3pt] at (img.south west)
  {\scalebox{0.6}{\begin{tabular}{@{}l@{}} {}\end{tabular}}};
\end{tikzpicture} &
& 
\begin{tikzpicture}
  \node[inner sep=0pt] (img) at (0,0) {\includegraphics[width=\linewidth]{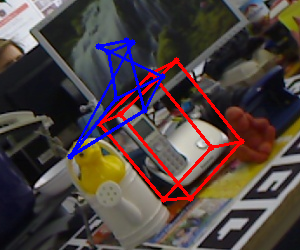}};
\node[anchor=south west, xshift=-3pt, yshift=-3pt] at (img.south west)
  {\scalebox{0.6}{\colorbox{gray!30}{\begin{tabular}{@{}l@{}} {phone000279}\end{tabular}}}};
\end{tikzpicture} &
\begin{tikzpicture}
  \node[inner sep=0pt] (img) at (0,0) {\includegraphics[width=\linewidth]{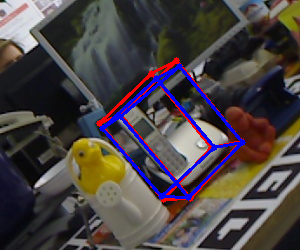}};
\node[anchor=south west, xshift=-3pt, yshift=-3pt] at (img.south west)
  {\scalebox{0.6}{\begin{tabular}{@{}l@{}} {}\end{tabular}}};
\end{tikzpicture}  \\
\end{tabular}
% \vspace{-1.1em}
\caption{Qualitative comparison of pose predictions on LineMOD images. Each pair shows results without UDA (left) and with \acronym (right). Ground-truth poses are shown in \textcolor{red}{red}, and predictions in \textcolor{blue}{blue}.}
\end{center}
\end{figure*}

%% file: appendix/3.4.viz_light.tex
% \textit{\textbf{}}
\begin{figure*}[h]
\begin{center}
{\Large \bfseries E. Improvements on Lightbox}\par \vspace{0.4em}

\renewcommand{\arraystretch}{0.8}
\begin{tabular}{@{}>{\centering\arraybackslash}m{0.23\textwidth}@{\hspace{3pt}}
                >{\centering\arraybackslash}m{0.23\textwidth}@{\hspace{8pt}}
                >{\centering\arraybackslash}m{0.012\textwidth}@{}
                >{\centering\arraybackslash}m{0.23\textwidth}@{\hspace{3pt}}
                >{\centering\arraybackslash}m{0.23\textwidth}@{}}

w/o UDA & \acronym & & w/o UDA & \acronym \\

\begin{tikzpicture}
  \node[inner sep=0pt] (img) at (0,0) {\includegraphics[width=\linewidth]{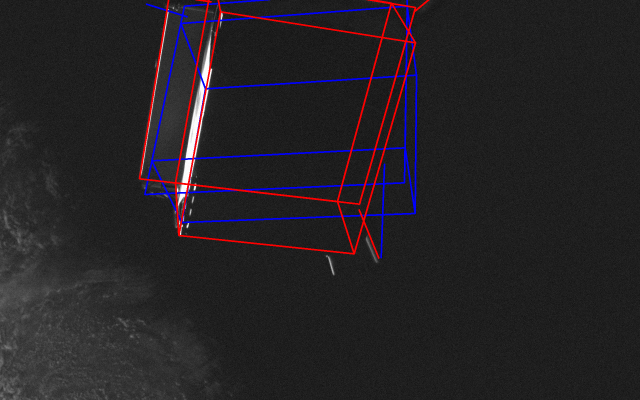}};
  \node[anchor=south west, xshift=-3pt, yshift=-3pt] at (img.south west)
  {\scalebox{0.5}{\colorbox{gray!30}{\begin{tabular}{@{}l@{}} {img000019}\end{tabular}}}};
  \node[anchor=south east, xshift=3pt, yshift=-3pt] at (img.south east)
      {\scalebox{0.5}{\colorbox{gray!30}{\begin{tabular}{@{}l@{}} {$E_r$: 116.1$^{\circ}$} \\ {$E_t$: 0.295\,$m$}\end{tabular}}}};
\end{tikzpicture} &
\begin{tikzpicture}
  \node[inner sep=0pt] (img) at (0,0) {\includegraphics[width=\linewidth]{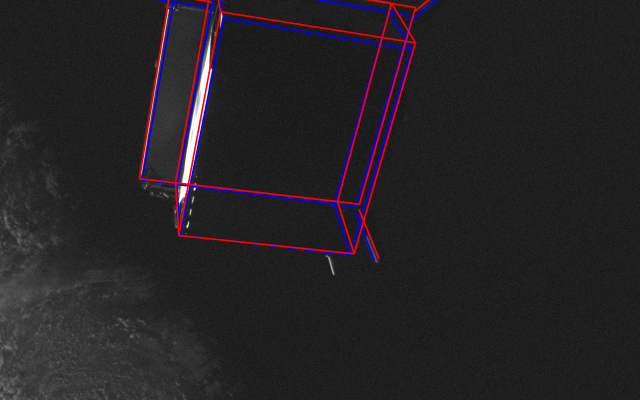}};
    \node[anchor=south west, xshift=-3pt, yshift=-3pt] at (img.south west)
  {\scalebox{0.5}{\begin{tabular}{@{}l@{}} {}\end{tabular}}};
  \node[anchor=south east, xshift=3pt, yshift=-3pt] at (img.south east)
      {\scalebox{0.5}{\colorbox{gray!30}{\begin{tabular}{@{}l@{}} {$E_r$: 0.75$^{\circ}$} \\ {$E_t$: 0.092\,$m$}\end{tabular}}}};
\end{tikzpicture} &
& 
\begin{tikzpicture}
  \node[inner sep=0pt] (img) at (0,0) {\includegraphics[width=\linewidth]{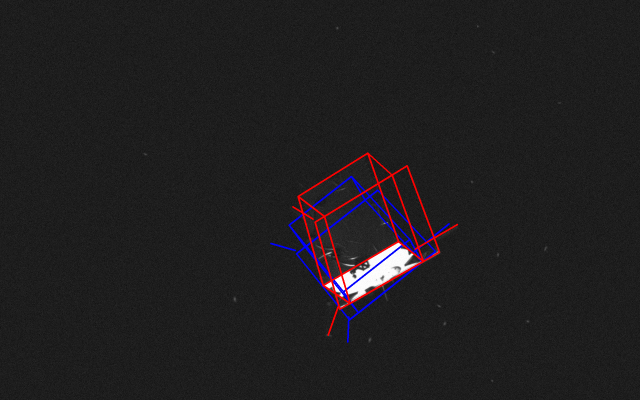}};
    \node[anchor=south west, xshift=-3pt, yshift=-3pt] at (img.south west)
  {\scalebox{0.5}{\colorbox{gray!30}{\begin{tabular}{@{}l@{}} {img000067}\end{tabular}}}};
  \node[anchor=south east, xshift=3pt, yshift=-3pt] at (img.south east)
      {\scalebox{0.5}{\colorbox{gray!30}{\begin{tabular}{@{}l@{}} {$E_r$: 23.4$^{\circ}$} \\ {$E_t$: 0.569\,$m$}\end{tabular}}}};
\end{tikzpicture} &
\begin{tikzpicture}
  \node[inner sep=0pt] (img) at (0,0) {\includegraphics[width=\linewidth]{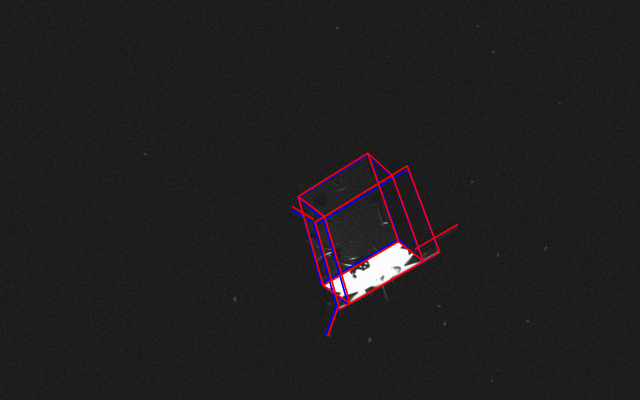}};
    \node[anchor=south west, xshift=-3pt, yshift=-3pt] at (img.south west)
  {\scalebox{0.5}{\begin{tabular}{@{}l@{}} {}\end{tabular}}};
  \node[anchor=south east, xshift=3pt, yshift=-3pt] at (img.south east)
      {\scalebox{0.5}{\colorbox{gray!30}{\begin{tabular}{@{}l@{}} {$E_r$: 3.32$^{\circ}$} \\ {$E_t$: 0.038\,$m$}\end{tabular}}}};
\end{tikzpicture}  \\

\begin{tikzpicture}
  \node[inner sep=0pt] (img) at (0,0) {\includegraphics[width=\linewidth]{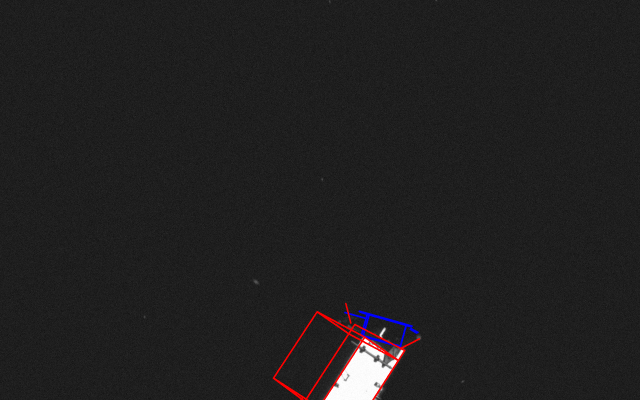}};
    \node[anchor=south west, xshift=-3pt, yshift=-3pt] at (img.south west)
  {\scalebox{0.5}{\colorbox{gray!30}{\begin{tabular}{@{}l@{}} {img000501}\end{tabular}}}};
  \node[anchor=south east, xshift=3pt, yshift=-3pt] at (img.south east)
      {\scalebox{0.5}{\colorbox{gray!30}{\begin{tabular}{@{}l@{}} {$E_r$: 155.9$^{\circ}$} \\ {$E_t$: 7.55\,$m$}\end{tabular}}}};
\end{tikzpicture} &
\begin{tikzpicture}
  \node[inner sep=0pt] (img) at (0,0) {\includegraphics[width=\linewidth]{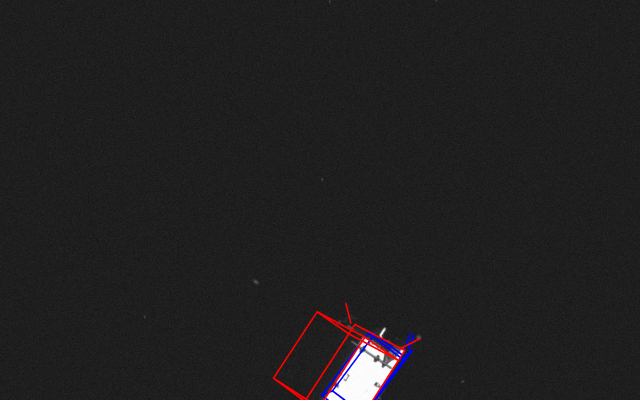}};
    \node[anchor=south west, xshift=-3pt, yshift=-3pt] at (img.south west)
  {\scalebox{0.5}{\begin{tabular}{@{}l@{}} {}\end{tabular}}};
  \node[anchor=south east, xshift=3pt, yshift=-3pt] at (img.south east)
      {\scalebox{0.5}{\colorbox{gray!30}{\begin{tabular}{@{}l@{}} {$E_r$: 35.49$^{\circ}$} \\ {$E_t$: 0.890\,$m$}\end{tabular}}}};
\end{tikzpicture} &
& 
\begin{tikzpicture}
  \node[inner sep=0pt] (img) at (0,0) {\includegraphics[width=\linewidth]{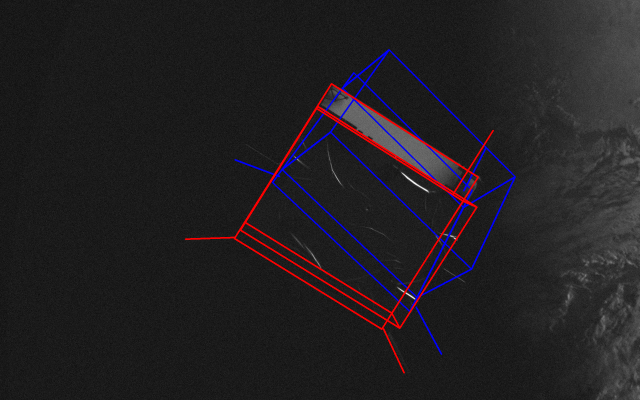}};
    \node[anchor=south west, xshift=-3pt, yshift=-3pt] at (img.south west)
  {\scalebox{0.5}{\colorbox{gray!30}{\begin{tabular}{@{}l@{}} {img006481}\end{tabular}}}};
  \node[anchor=south east, xshift=3pt, yshift=-3pt] at (img.south east)
      {\scalebox{0.5}{\colorbox{gray!30}{\begin{tabular}{@{}l@{}} {$E_r$: 45.64$^{\circ}$} \\ {$E_t$: 0.362\,$m$}\end{tabular}}}};
\end{tikzpicture} &
\begin{tikzpicture}
  \node[inner sep=0pt] (img) at (0,0) {\includegraphics[width=\linewidth]{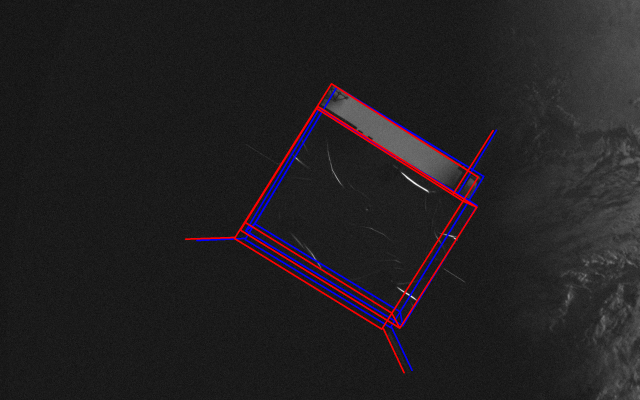}};
    \node[anchor=south west, xshift=-3pt, yshift=-3pt] at (img.south west)
  {\scalebox{0.5}{\begin{tabular}{@{}l@{}} {}\end{tabular}}};
  \node[anchor=south east, xshift=3pt, yshift=-3pt] at (img.south east)
      {\scalebox{0.5}{\colorbox{gray!30}{\begin{tabular}{@{}l@{}} {$E_r$: 4.28$^{\circ}$} \\ {$E_t$: 0.081\,$m$}\end{tabular}}}};
\end{tikzpicture}  \\

\begin{tikzpicture}
  \node[inner sep=0pt] (img) at (0,0) {\includegraphics[width=\linewidth]{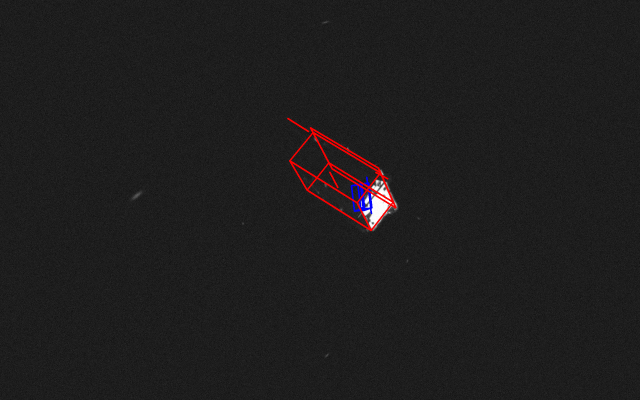}};
    \node[anchor=south west, xshift=-3pt, yshift=-3pt] at (img.south west)
  {\scalebox{0.5}{\colorbox{gray!30}{\begin{tabular}{@{}l@{}} {img000977}\end{tabular}}}};
  \node[anchor=south east, xshift=3pt, yshift=-3pt] at (img.south east)
      {\scalebox{0.5}{\colorbox{gray!30}{\begin{tabular}{@{}l@{}} {$E_r$: 157.0$^{\circ}$} \\ {$E_t$: 20.67\,$m$}\end{tabular}}}};
\end{tikzpicture} &
\begin{tikzpicture}
  \node[inner sep=0pt] (img) at (0,0) {\includegraphics[width=\linewidth]{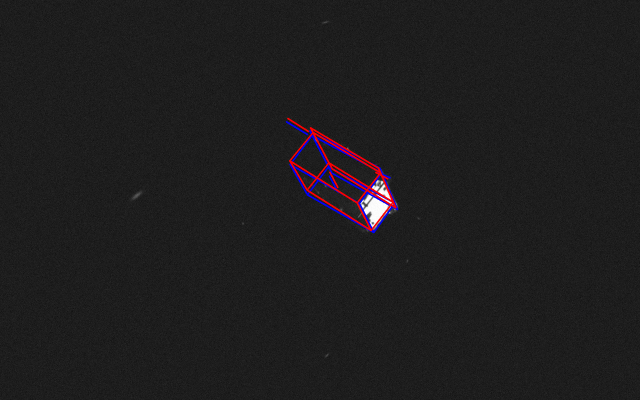}};
    \node[anchor=south west, xshift=-3pt, yshift=-3pt] at (img.south west)
  {\scalebox{0.5}{\begin{tabular}{@{}l@{}} {}\end{tabular}}};
  \node[anchor=south east, xshift=3pt, yshift=-3pt] at (img.south east)
      {\scalebox{0.5}{\colorbox{gray!30}{\begin{tabular}{@{}l@{}} {$E_r$: 2.59$^{\circ}$} \\ {$E_t$: 0.064\,$m$}\end{tabular}}}};
\end{tikzpicture} &
& 
\begin{tikzpicture}
  \node[inner sep=0pt] (img) at (0,0) {\includegraphics[width=\linewidth]{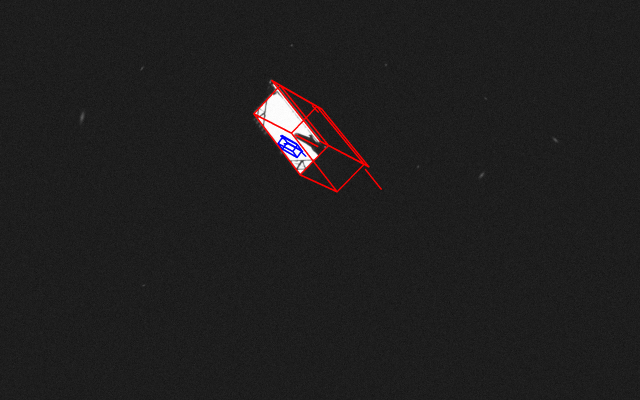}};
    \node[anchor=south west, xshift=-3pt, yshift=-3pt] at (img.south west)
  {\scalebox{0.5}{\colorbox{gray!30}{\begin{tabular}{@{}l@{}} {img001781}\end{tabular}}}};
  \node[anchor=south east, xshift=3pt, yshift=-3pt] at (img.south east)
      {\scalebox{0.5}{\colorbox{gray!30}{\begin{tabular}{@{}l@{}} {$E_r$: 114.9$^{\circ}$} \\ {$E_t$: 30.82\,$m$}\end{tabular}}}};
\end{tikzpicture} &
\begin{tikzpicture}
  \node[inner sep=0pt] (img) at (0,0) {\includegraphics[width=\linewidth]{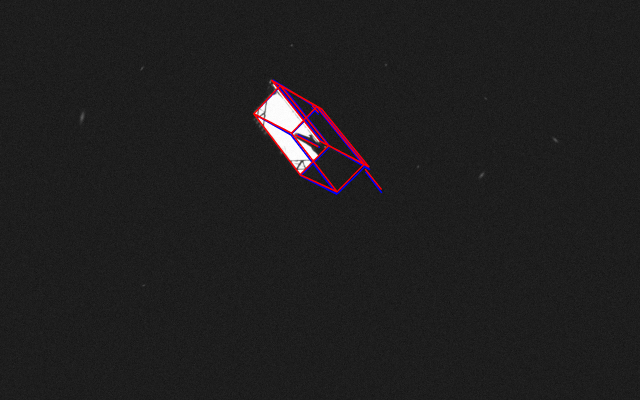}};
    \node[anchor=south west, xshift=-3pt, yshift=-3pt] at (img.south west)
  {\scalebox{0.5}{\begin{tabular}{@{}l@{}} {}\end{tabular}}};
  \node[anchor=south east, xshift=3pt, yshift=-3pt] at (img.south east)
      {\scalebox{0.5}{\colorbox{gray!30}{\begin{tabular}{@{}l@{}} {$E_r$: 1.74$^{\circ}$} \\ {$E_t$: 0.111\,$m$}\end{tabular}}}};
\end{tikzpicture} \\

\begin{tikzpicture}
  \node[inner sep=0pt] (img) at (0,0) {\includegraphics[width=\linewidth]{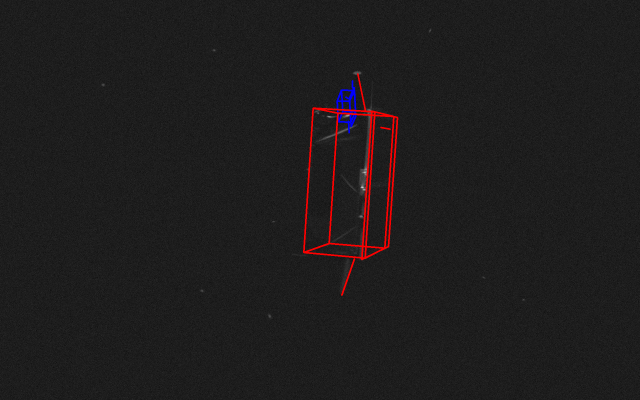}};
    \node[anchor=south west, xshift=-3pt, yshift=-3pt] at (img.south west)
  {\scalebox{0.5}{\colorbox{gray!30}{\begin{tabular}{@{}l@{}} {img002460}\end{tabular}}}};
  \node[anchor=south east, xshift=3pt, yshift=-3pt] at (img.south east)
      {\scalebox{0.5}{\colorbox{gray!30}{\begin{tabular}{@{}l@{}} {$E_r$: 110.56$^{\circ}$} \\ {$E_t$: 19.76\,$m$}\end{tabular}}}};
\end{tikzpicture} &
\begin{tikzpicture}
  \node[inner sep=0pt] (img) at (0,0) {\includegraphics[width=\linewidth]{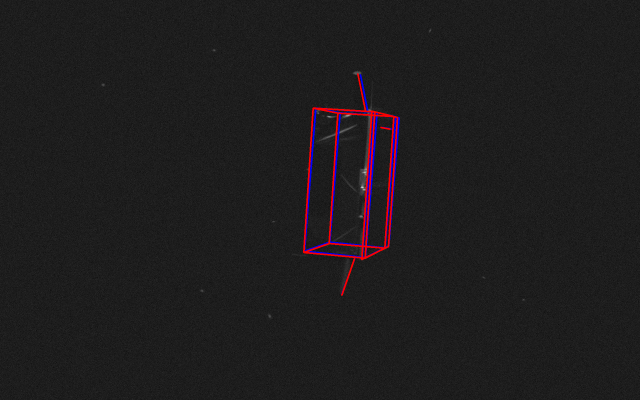}};
    \node[anchor=south west, xshift=-3pt, yshift=-3pt] at (img.south west)
  {\scalebox{0.5}{\begin{tabular}{@{}l@{}} {}\end{tabular}}};
  \node[anchor=south east, xshift=3pt, yshift=-3pt] at (img.south east)
      {\scalebox{0.5}{\colorbox{gray!30}{\begin{tabular}{@{}l@{}} {$E_r$: 0.75$^{\circ}$} \\ {$E_t$: 0.031\,$m$}\end{tabular}}}};
\end{tikzpicture} &
& 
\begin{tikzpicture}
  \node[inner sep=0pt] (img) at (0,0) {\includegraphics[width=\linewidth]{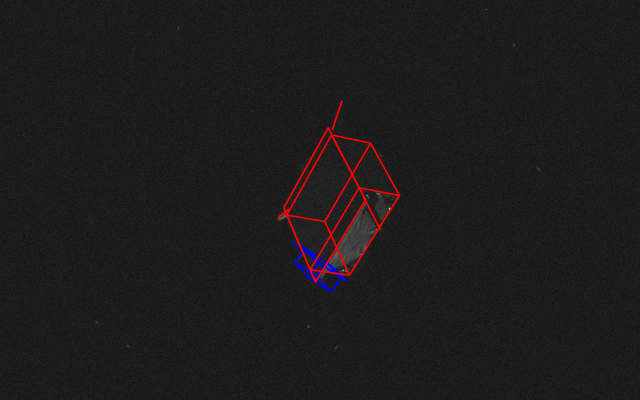}};
    \node[anchor=south west, xshift=-3pt, yshift=-3pt] at (img.south west)
  {\scalebox{0.5}{\colorbox{gray!30}{\begin{tabular}{@{}l@{}} {img003100}\end{tabular}}}};
  \node[anchor=south east, xshift=3pt, yshift=-3pt] at (img.south east)
      {\scalebox{0.5}{\colorbox{gray!30}{\begin{tabular}{@{}l@{}} {$E_r$: 138.8$^{\circ}$} \\ {$E_t$: 11.65\,$m$}\end{tabular}}}};
\end{tikzpicture} &
\begin{tikzpicture}
  \node[inner sep=0pt] (img) at (0,0) {\includegraphics[width=\linewidth]{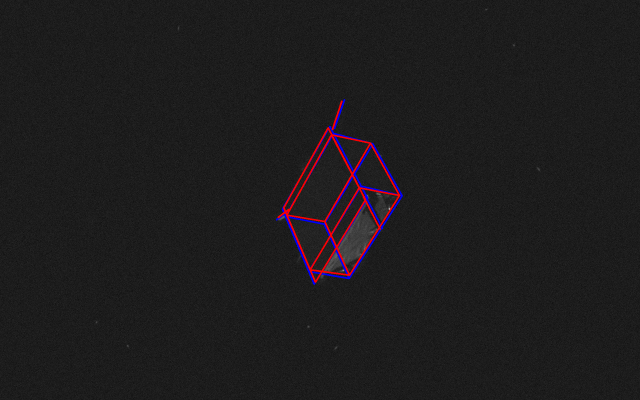}};
    \node[anchor=south west, xshift=-3pt, yshift=-3pt] at (img.south west)
  {\scalebox{0.5}{\begin{tabular}{@{}l@{}} {}\end{tabular}}};
  \node[anchor=south east, xshift=3pt, yshift=-3pt] at (img.south east)
      {\scalebox{0.5}{\colorbox{gray!30}{\begin{tabular}{@{}l@{}} {$E_r$: 2.15$^{\circ}$} \\ {$E_t$: 0.173\,$m$}\end{tabular}}}};
\end{tikzpicture}  \\

\begin{tikzpicture}
  \node[inner sep=0pt] (img) at (0,0) {\includegraphics[width=\linewidth]{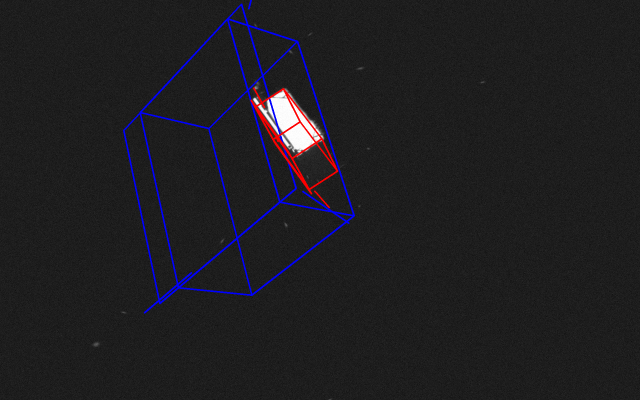}};
    \node[anchor=south west, xshift=-3pt, yshift=-3pt] at (img.south west)
  {\scalebox{0.5}{\colorbox{gray!30}{\begin{tabular}{@{}l@{}} {img003221}\end{tabular}}}};
  \node[anchor=south east, xshift=3pt, yshift=-3pt] at (img.south east)
      {\scalebox{0.5}{\colorbox{gray!30}{\begin{tabular}{@{}l@{}} {$E_r$: 167.8$^{\circ}$} \\ {$E_t$: 5.85\,$m$}\end{tabular}}}};
\end{tikzpicture} &
\begin{tikzpicture}
  \node[inner sep=0pt] (img) at (0,0) {\includegraphics[width=\linewidth]{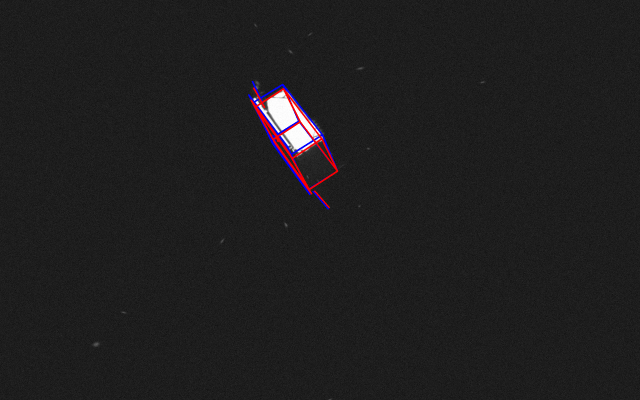}};
    \node[anchor=south west, xshift=-3pt, yshift=-3pt] at (img.south west)
  {\scalebox{0.5}{\begin{tabular}{@{}l@{}} {}\end{tabular}}};
  \node[anchor=south east, xshift=3pt, yshift=-3pt] at (img.south east)
      {\scalebox{0.5}{\colorbox{gray!30}{\begin{tabular}{@{}l@{}} {$E_r$: 2.98$^{\circ}$} \\ {$E_t$: 0.381\,$m$}\end{tabular}}}};
\end{tikzpicture} &
& 
\begin{tikzpicture}
  \node[inner sep=0pt] (img) at (0,0) {\includegraphics[width=\linewidth]{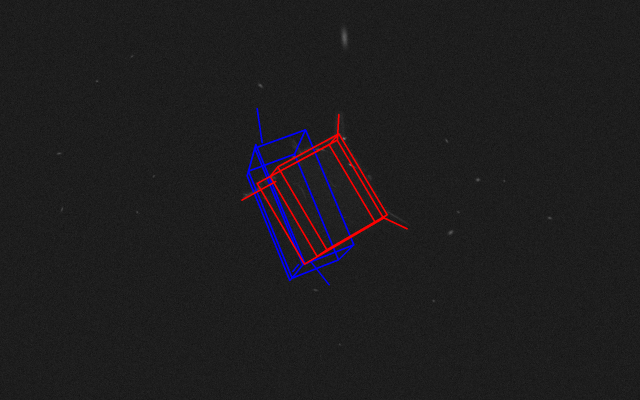}};
    \node[anchor=south west, xshift=-3pt, yshift=-3pt] at (img.south west)
  {\scalebox{0.5}{\colorbox{gray!30}{\begin{tabular}{@{}l@{}} {img004521}\end{tabular}}}};
  \node[anchor=south east, xshift=3pt, yshift=-3pt] at (img.south east)
      {\scalebox{0.5}{\colorbox{gray!30}{\begin{tabular}{@{}l@{}} {$E_r$: 172.3$^{\circ}$} \\ {$E_t$: 1.57\,$m$}\end{tabular}}}};
\end{tikzpicture} &
\begin{tikzpicture}
  \node[inner sep=0pt] (img) at (0,0) {\includegraphics[width=\linewidth]{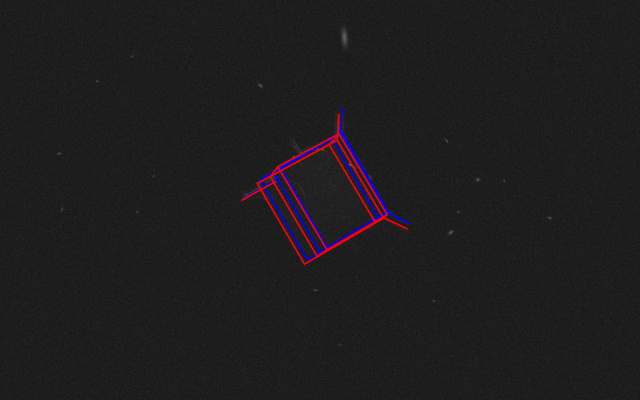}};
    \node[anchor=south west, xshift=-3pt, yshift=-3pt] at (img.south west)
  {\scalebox{0.5}{\begin{tabular}{@{}l@{}} {}\end{tabular}}};
  \node[anchor=south east, xshift=3pt, yshift=-3pt] at (img.south east)
      {\scalebox{0.5}{\colorbox{gray!30}{\begin{tabular}{@{}l@{}} {$E_r$: 6.10$^{\circ}$} \\ {$E_t$: 0.044\,$m$}\end{tabular}}}};
\end{tikzpicture}  \\

\begin{tikzpicture}
  \node[inner sep=0pt] (img) at (0,0) {\includegraphics[width=\linewidth]{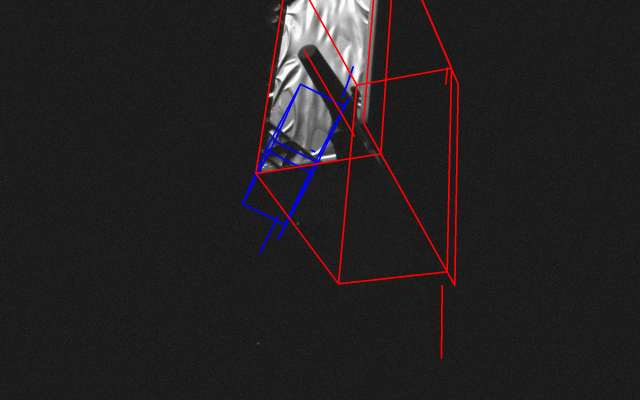}};
    \node[anchor=south west, xshift=-3pt, yshift=-3pt] at (img.south west)
  {\scalebox{0.5}{\colorbox{gray!30}{\begin{tabular}{@{}l@{}} {img0004650}\end{tabular}}}};
  \node[anchor=south east, xshift=3pt, yshift=-3pt] at (img.south east)
      {\scalebox{0.5}{\colorbox{gray!30}{\begin{tabular}{@{}l@{}} {$E_r$: 95.89$^{\circ}$} \\ {$E_t$: 3.75\,$m$}\end{tabular}}}};
\end{tikzpicture} &
\begin{tikzpicture}
  \node[inner sep=0pt] (img) at (0,0) {\includegraphics[width=\linewidth]{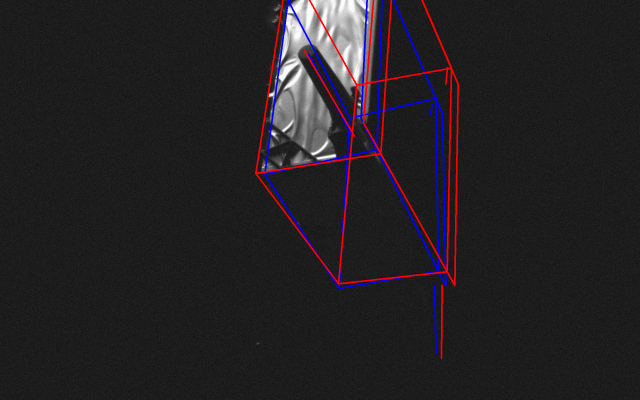}};
    \node[anchor=south west, xshift=-3pt, yshift=-3pt] at (img.south west)
  {\scalebox{0.5}{\begin{tabular}{@{}l@{}} {}\end{tabular}}};
  \node[anchor=south east, xshift=3pt, yshift=-3pt] at (img.south east)
      {\scalebox{0.5}{\colorbox{gray!30}{\begin{tabular}{@{}l@{}} {$E_r$: 6.47$^{\circ}$} \\ {$E_t$: 0.302\,$m$}\end{tabular}}}};
\end{tikzpicture} &
& 
\begin{tikzpicture}
  \node[inner sep=0pt] (img) at (0,0) {\includegraphics[width=\linewidth]{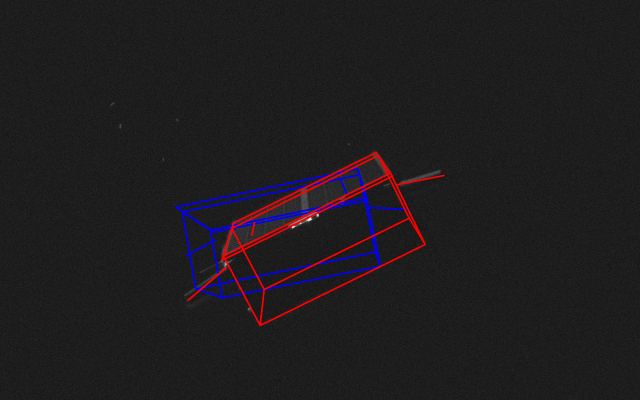}};
    \node[anchor=south west, xshift=-3pt, yshift=-3pt] at (img.south west)
  {\scalebox{0.5}{\colorbox{gray!30}{\begin{tabular}{@{}l@{}} {img004974}\end{tabular}}}};
  \node[anchor=south east, xshift=3pt, yshift=-3pt] at (img.south east)
      {\scalebox{0.5}{\colorbox{gray!30}{\begin{tabular}{@{}l@{}} {$E_r$: 179.0$^{\circ}$} \\ {$E_t$: 0.205\,$m$}\end{tabular}}}};
\end{tikzpicture} &
\begin{tikzpicture}
  \node[inner sep=0pt] (img) at (0,0) {\includegraphics[width=\linewidth]{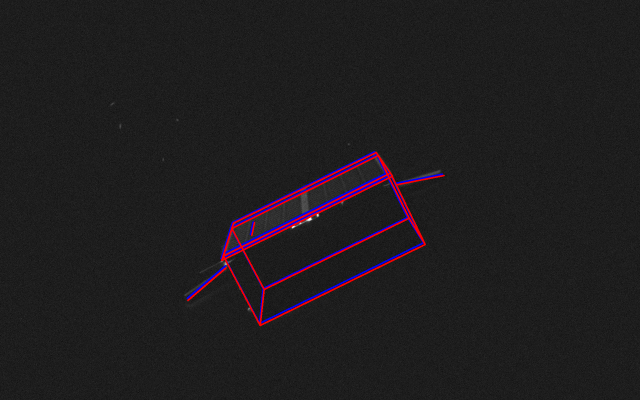}};
    \node[anchor=south west, xshift=-3pt, yshift=-3pt] at (img.south west)
  {\scalebox{0.5}{\begin{tabular}{@{}l@{}} {}\end{tabular}}};
  \node[anchor=south east, xshift=3pt, yshift=-3pt] at (img.south east)
      {\scalebox{0.5}{\colorbox{gray!30}{\begin{tabular}{@{}l@{}} {$E_r$: 0.25$^{\circ}$} \\ {$E_t$: 0.021\,$m$}\end{tabular}}}};
\end{tikzpicture}  \\

\begin{tikzpicture}
  \node[inner sep=0pt] (img) at (0,0) {\includegraphics[width=\linewidth]{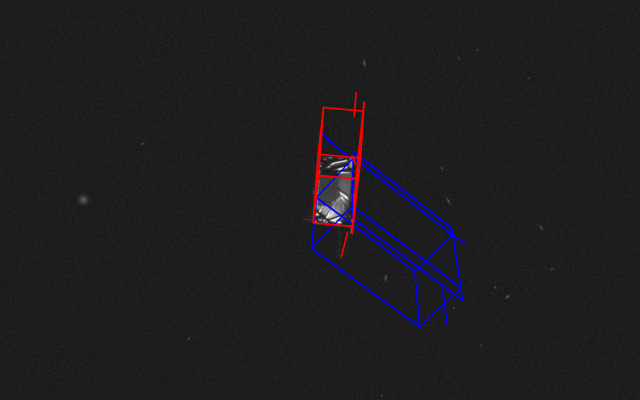}};
    \node[anchor=south west, xshift=-3pt, yshift=-3pt] at (img.south west)
  {\scalebox{0.5}{\colorbox{gray!30}{\begin{tabular}{@{}l@{}} {img005218}\end{tabular}}}};
  \node[anchor=south east, xshift=3pt, yshift=-3pt] at (img.south east)
      {\scalebox{0.5}{\colorbox{gray!30}{\begin{tabular}{@{}l@{}} {$E_r$: 165.7$^{\circ}$} \\ {$E_t$: 2.69\,$m$}\end{tabular}}}};
\end{tikzpicture} &
\begin{tikzpicture}
  \node[inner sep=0pt] (img) at (0,0) {\includegraphics[width=\linewidth]{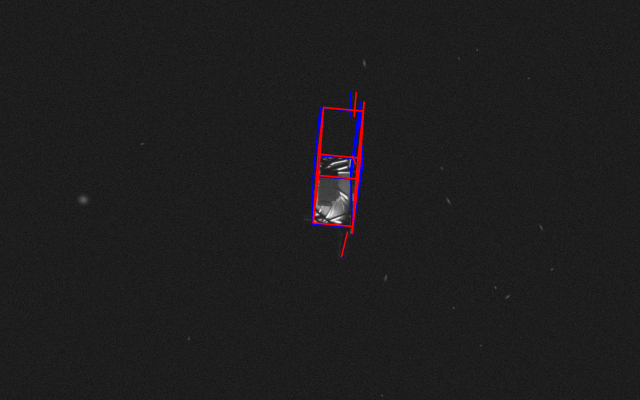}};
    \node[anchor=south west, xshift=-3pt, yshift=-3pt] at (img.south west)
  {\scalebox{0.5}{\begin{tabular}{@{}l@{}} {}\end{tabular}}};
  \node[anchor=south east, xshift=3pt, yshift=-3pt] at (img.south east)
      {\scalebox{0.5}{\colorbox{gray!30}{\begin{tabular}{@{}l@{}} {$E_r$: 4.30$^{\circ}$} \\ {$E_t$: 0.043\,$m$}\end{tabular}}}};
\end{tikzpicture} &
& 
\begin{tikzpicture}
  \node[inner sep=0pt] (img) at (0,0) {\includegraphics[width=\linewidth]{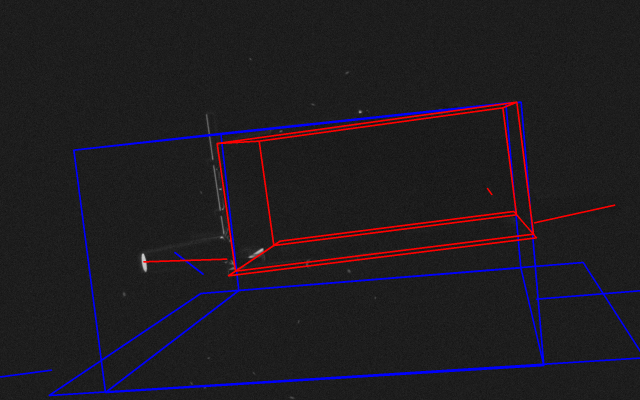}};
    \node[anchor=south west, xshift=-3pt, yshift=-3pt] at (img.south west)
  {\scalebox{0.5}{\colorbox{gray!30}{\begin{tabular}{@{}l@{}} {img005601}\end{tabular}}}};
  \node[anchor=south east, xshift=3pt, yshift=-3pt] at (img.south east)
      {\scalebox{0.5}{\colorbox{gray!30}{\begin{tabular}{@{}l@{}} {$E_r$: 92.32$^{\circ}$} \\ {$E_t$: 1.14\,$m$}\end{tabular}}}};
\end{tikzpicture} &
\begin{tikzpicture}
  \node[inner sep=0pt] (img) at (0,0) {\includegraphics[width=\linewidth]{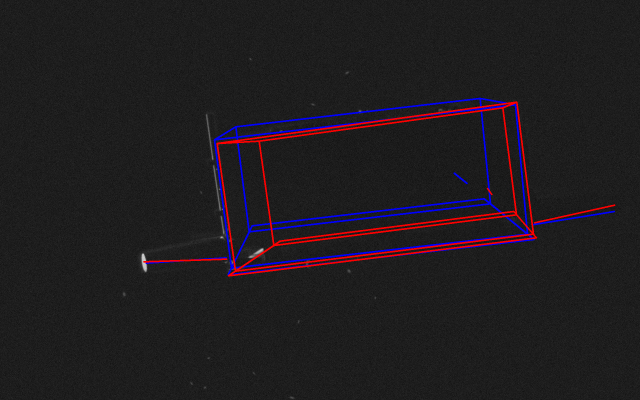}};
    \node[anchor=south west, xshift=-3pt, yshift=-3pt] at (img.south west)
  {\scalebox{0.5}{\begin{tabular}{@{}l@{}} {}\end{tabular}}};
  \node[anchor=south east, xshift=3pt, yshift=-3pt] at (img.south east)
      {\scalebox{0.5}{\colorbox{gray!30}{\begin{tabular}{@{}l@{}} {$E_r$: 7.62$^{\circ}$} \\ {$E_t$: 0.039\,$m$}\end{tabular}}}};
\end{tikzpicture}  \\

\begin{tikzpicture}
  \node[inner sep=0pt] (img) at (0,0) {\includegraphics[width=\linewidth]{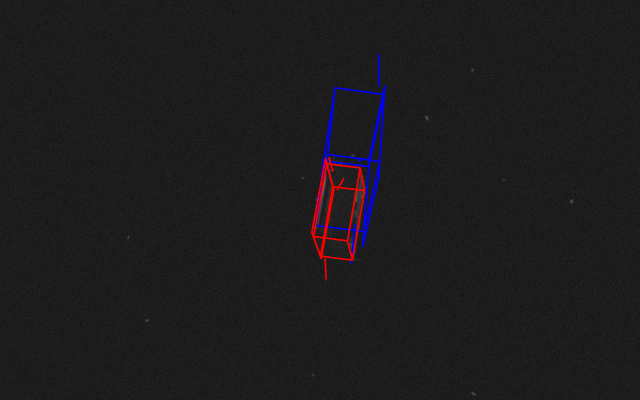}};
    \node[anchor=south west, xshift=-3pt, yshift=-3pt] at (img.south west)
  {\scalebox{0.5}{\colorbox{gray!30}{\begin{tabular}{@{}l@{}} {img005785}\end{tabular}}}};
  \node[anchor=south east, xshift=3pt, yshift=-3pt] at (img.south east)
      {\scalebox{0.5}{\colorbox{gray!30}{\begin{tabular}{@{}l@{}} {$E_r$: 176.4$^{\circ}$} \\ {$E_t$: 3.06\,$m$}\end{tabular}}}};
\end{tikzpicture} &
\begin{tikzpicture}
  \node[inner sep=0pt] (img) at (0,0) {\includegraphics[width=\linewidth]{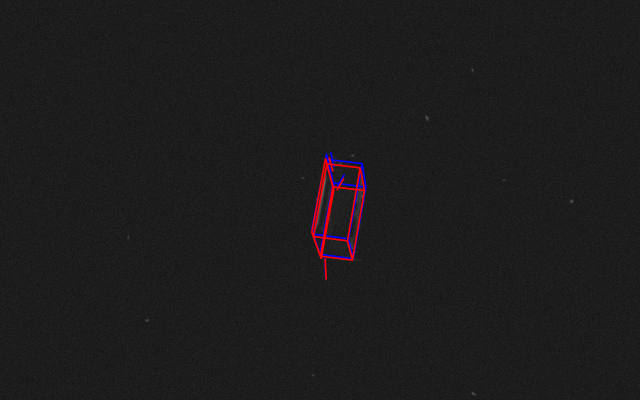}};
    \node[anchor=south west, xshift=-3pt, yshift=-3pt] at (img.south west)
  {\scalebox{0.5}{\begin{tabular}{@{}l@{}} {}\end{tabular}}};
  \node[anchor=south east, xshift=3pt, yshift=-3pt] at (img.south east)
      {\scalebox{0.5}{\colorbox{gray!30}{\begin{tabular}{@{}l@{}} {$E_r$: 0.74$^{\circ}$} \\ {$E_t$: 0.297\,$m$}\end{tabular}}}};
\end{tikzpicture} &
& 
\begin{tikzpicture}
  \node[inner sep=0pt] (img) at (0,0) {\includegraphics[width=\linewidth]{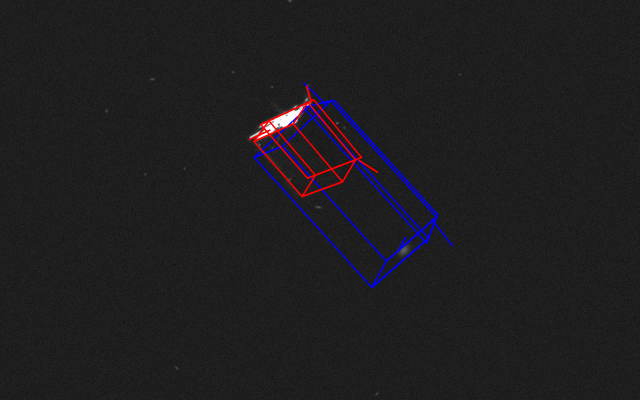}};
    \node[anchor=south west, xshift=-3pt, yshift=-3pt] at (img.south west)
  {\scalebox{0.5}{\colorbox{gray!30}{\begin{tabular}{@{}l@{}} {img006307}\end{tabular}}}};
  \node[anchor=south east, xshift=3pt, yshift=-3pt] at (img.south east)
      {\scalebox{0.5}{\colorbox{gray!30}{\begin{tabular}{@{}l@{}} {$E_r$: 158.9$^{\circ}$} \\ {$E_t$: 4.97\,$m$}\end{tabular}}}};
\end{tikzpicture} &
\begin{tikzpicture}
  \node[inner sep=0pt] (img) at (0,0) {\includegraphics[width=\linewidth]{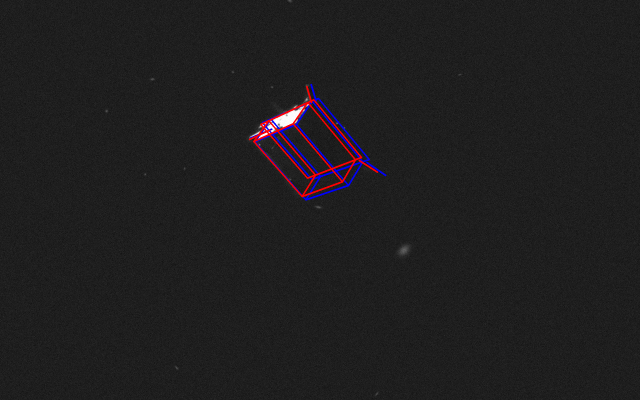}};
    \node[anchor=south west, xshift=-3pt, yshift=-3pt] at (img.south west)
  {\scalebox{0.5}{\begin{tabular}{@{}l@{}} {}\end{tabular}}};
  \node[anchor=south east, xshift=3pt, yshift=-3pt] at (img.south east)
      {\scalebox{0.5}{\colorbox{gray!30}{\begin{tabular}{@{}l@{}} {$E_r$: 1.73$^{\circ}$} \\ {$E_t$: 0.571\,$m$}\end{tabular}}}};
\end{tikzpicture}  \\

\end{tabular}
% \vspace{-1em}
\caption{Qualitative comparison of pose predictions on Lightbox. Each pair shows results without UDA (left) and with \acronym (right). Rotation and translation errors ($E_r$, $E_t$) are reported. Ground-truth poses are shown in \textcolor{red}{red}, and predictions in \textcolor{blue}{blue}.}
\end{center}
\end{figure*}

%% file: appendix/3.5.viz_sun.tex
% \textit{\textbf{C. Sunlamp}}
% \vspace{-1em}
\begin{figure*}[h]
\begin{center}
{\Large \bfseries F. Improvements on Sunlamp}\par \vspace{0.4em}
\renewcommand{\arraystretch}{1}
\begin{tabular}{@{}>{\centering\arraybackslash}m{0.23\textwidth}@{\hspace{3pt}}
                >{\centering\arraybackslash}m{0.23\textwidth}@{\hspace{8pt}}
                >{\centering\arraybackslash}m{0.012\textwidth}@{}
                >{\centering\arraybackslash}m{0.23\textwidth}@{\hspace{3pt}}
                >{\centering\arraybackslash}m{0.23\textwidth}@{}}

w/o UDA & \acronym & & w/o UDA & \acronym \\

\begin{tikzpicture}
  \node[inner sep=0pt] (img) at (0,0) {\includegraphics[width=\linewidth]{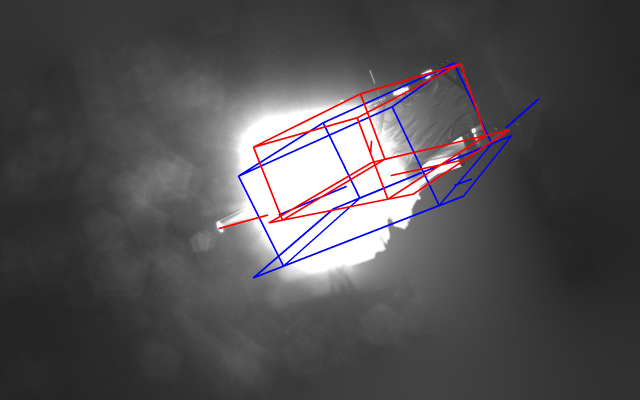}};
  \node[anchor=south west, xshift=-3pt, yshift=-3pt] at (img.south west)
  {\scalebox{0.5}{\colorbox{gray!30}{\begin{tabular}{@{}l@{}} {img000026}\end{tabular}}}};
  \node[anchor=south east, xshift=3pt, yshift=-3pt] at (img.south east)
      {\scalebox{0.5}{\colorbox{gray!30}{\begin{tabular}{@{}l@{}} {$E_r$: 162.1$^{\circ}$} \\ {$E_t$: 0.718\,$m$}\end{tabular}}}};
\end{tikzpicture} &
\begin{tikzpicture}
  \node[inner sep=0pt] (img) at (0,0) {\includegraphics[width=\linewidth]{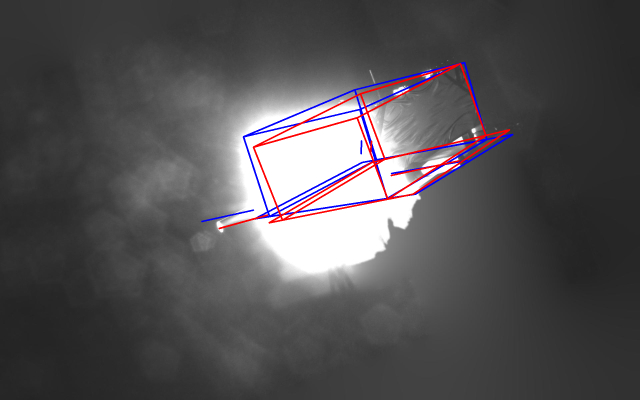}};
  \node[anchor=south west, xshift=-3pt, yshift=-3pt] at (img.south west)
  {\scalebox{0.5}{\begin{tabular}{@{}l@{}} {}\end{tabular}}};
  \node[anchor=south east, xshift=3pt, yshift=-3pt] at (img.south east)
      {\scalebox{0.5}{\colorbox{gray!30}{\begin{tabular}{@{}l@{}} {$E_r$: 4.16$^{\circ}$} \\ {$E_t$: 0.222\,$m$}\end{tabular}}}};
\end{tikzpicture} &
& 
\begin{tikzpicture}
  \node[inner sep=0pt] (img) at (0,0) {\includegraphics[width=\linewidth]{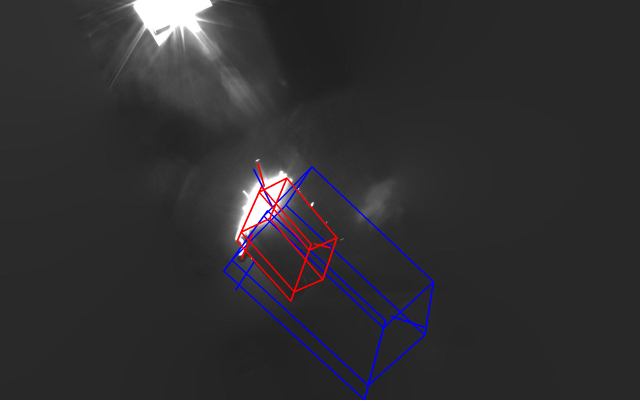}};
  \node[anchor=south west, xshift=-3pt, yshift=-3pt] at (img.south west)
  {\scalebox{0.5}{\colorbox{gray!30}{\begin{tabular}{@{}l@{}} {img000048}\end{tabular}}}};
  \node[anchor=south east, xshift=3pt, yshift=-3pt] at (img.south east)
      {\scalebox{0.5}{\colorbox{gray!30}{\begin{tabular}{@{}l@{}} {$E_r$: 24.97$^{\circ}$} \\ {$E_t$: 4.48\,$m$}\end{tabular}}}};
\end{tikzpicture} &
\begin{tikzpicture}
  \node[inner sep=0pt] (img) at (0,0) {\includegraphics[width=\linewidth]{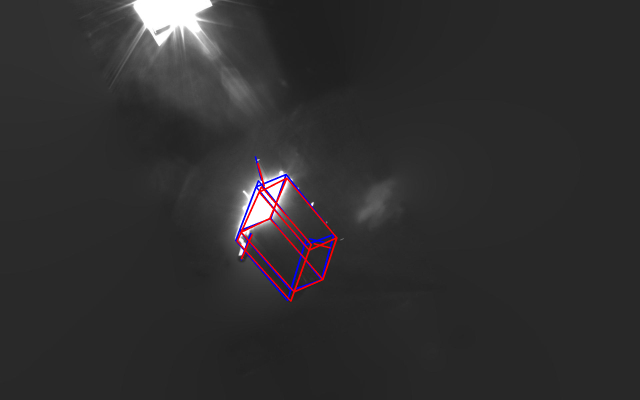}};
  \node[anchor=south west, xshift=-3pt, yshift=-3pt] at (img.south west)
  {\scalebox{0.5}{\begin{tabular}{@{}l@{}} {}\end{tabular}}};
  \node[anchor=south east, xshift=3pt, yshift=-3pt] at (img.south east)
      {\scalebox{0.5}{\colorbox{gray!30}{\begin{tabular}{@{}l@{}} {$E_r$: 5.13$^{\circ}$} \\ {$E_t$: 0.215\,$m$}\end{tabular}}}};
\end{tikzpicture}  \\

\begin{tikzpicture}
  \node[inner sep=0pt] (img) at (0,0) {\includegraphics[width=\linewidth]{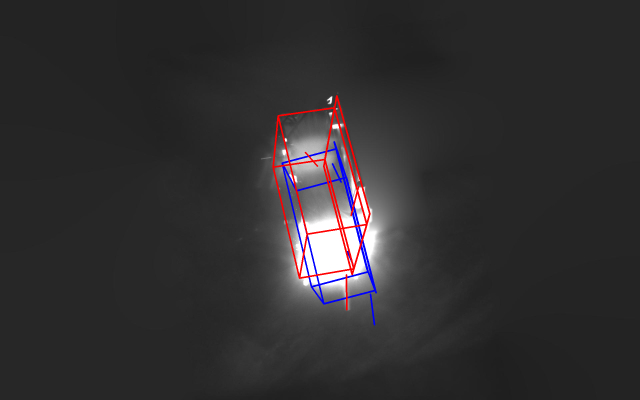}};
  \node[anchor=south west, xshift=-3pt, yshift=-3pt] at (img.south west)
  {\scalebox{0.5}{\colorbox{gray!30}{\begin{tabular}{@{}l@{}} {img000328}\end{tabular}}}};
  \node[anchor=south east, xshift=3pt, yshift=-3pt] at (img.south east)
      {\scalebox{0.5}{\colorbox{gray!30}{\begin{tabular}{@{}l@{}} {$E_r$: 17.6$^{\circ}$} \\ {$E_t$: 0.385\,$m$}\end{tabular}}}};
\end{tikzpicture} &
\begin{tikzpicture}
  \node[inner sep=0pt] (img) at (0,0) {\includegraphics[width=\linewidth]{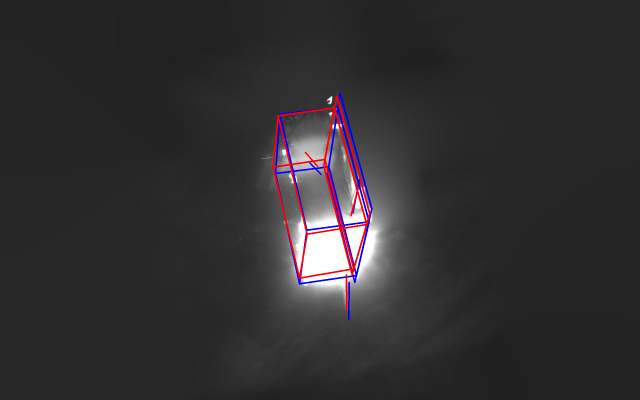}};
  \node[anchor=south west, xshift=-3pt, yshift=-3pt] at (img.south west)
  {\scalebox{0.5}{\begin{tabular}{@{}l@{}} {}\end{tabular}}};
  \node[anchor=south east, xshift=3pt, yshift=-3pt] at (img.south east)
      {\scalebox{0.5}{\colorbox{gray!30}{\begin{tabular}{@{}l@{}} {$E_r$: 5.02$^{\circ}$} \\ {$E_t$: 0.176\,$m$}\end{tabular}}}};
\end{tikzpicture} &
& 
\begin{tikzpicture}
  \node[inner sep=0pt] (img) at (0,0) {\includegraphics[width=\linewidth]{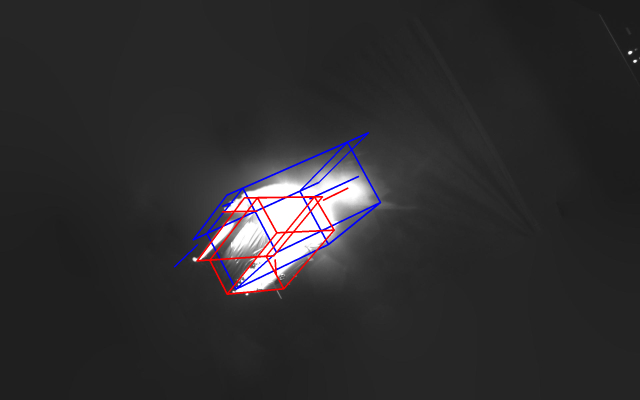}};
  \node[anchor=south west, xshift=-3pt, yshift=-3pt] at (img.south west)
  {\scalebox{0.5}{\colorbox{gray!30}{\begin{tabular}{@{}l@{}} {img000358}\end{tabular}}}};
  \node[anchor=south east, xshift=3pt, yshift=-3pt] at (img.south east)
      {\scalebox{0.5}{\colorbox{gray!30}{\begin{tabular}{@{}l@{}} {$E_r$: 154.8$^{\circ}$} \\ {$E_t$: 2.61\,$m$}\end{tabular}}}};
\end{tikzpicture} &
\begin{tikzpicture}
  \node[inner sep=0pt] (img) at (0,0) {\includegraphics[width=\linewidth]{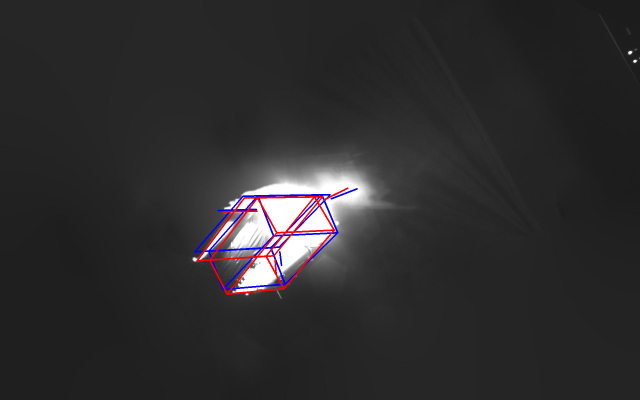}};
  \node[anchor=south west, xshift=-3pt, yshift=-3pt] at (img.south west)
  {\scalebox{0.5}{\begin{tabular}{@{}l@{}} {}\end{tabular}}};
  \node[anchor=south east, xshift=3pt, yshift=-3pt] at (img.south east)
      {\scalebox{0.5}{\colorbox{gray!30}{\begin{tabular}{@{}l@{}} {$E_r$: 7.96$^{\circ}$} \\ {$E_t$: 0.305\,$m$}\end{tabular}}}};
\end{tikzpicture} \\

\begin{tikzpicture}
  \node[inner sep=0pt] (img) at (0,0) {\includegraphics[width=\linewidth]{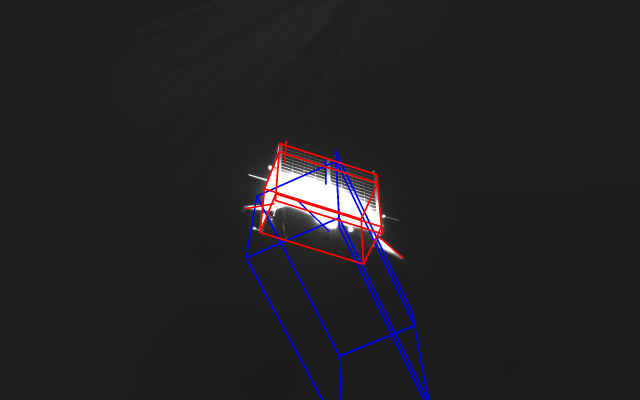}};
  \node[anchor=south west, xshift=-3pt, yshift=-3pt] at (img.south west)
  {\scalebox{0.5}{\colorbox{gray!30}{\begin{tabular}{@{}l@{}} {img000955}\end{tabular}}}};
  \node[anchor=south east, xshift=3pt, yshift=-3pt] at (img.south east)
      {\scalebox{0.5}{\colorbox{gray!30}{\begin{tabular}{@{}l@{}} {$E_r$: 46.33$^{\circ}$} \\ {$E_t$: 3.41\,$m$}\end{tabular}}}};
\end{tikzpicture} &
\begin{tikzpicture}
  \node[inner sep=0pt] (img) at (0,0) {\includegraphics[width=\linewidth]{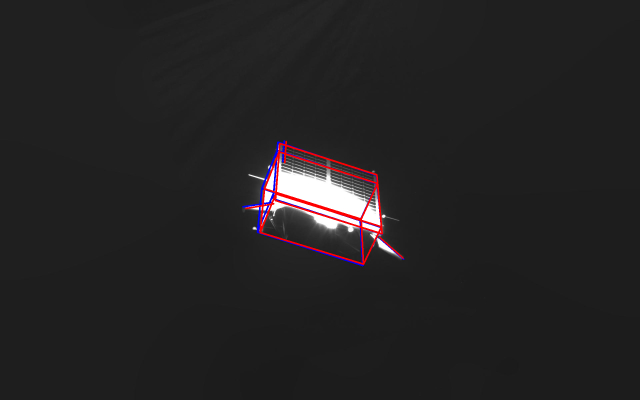}};
  \node[anchor=south west, xshift=-3pt, yshift=-3pt] at (img.south west)
  {\scalebox{0.5}{\begin{tabular}{@{}l@{}} {}\end{tabular}}};
  \node[anchor=south east, xshift=3pt, yshift=-3pt] at (img.south east)
      {\scalebox{0.5}{\colorbox{gray!30}{\begin{tabular}{@{}l@{}} {$E_r$: 0.98$^{\circ}$} \\ {$E_t$: 0.152\,$m$}\end{tabular}}}};
\end{tikzpicture} &
& 
\begin{tikzpicture}
  \node[inner sep=0pt] (img) at (0,0) {\includegraphics[width=\linewidth]{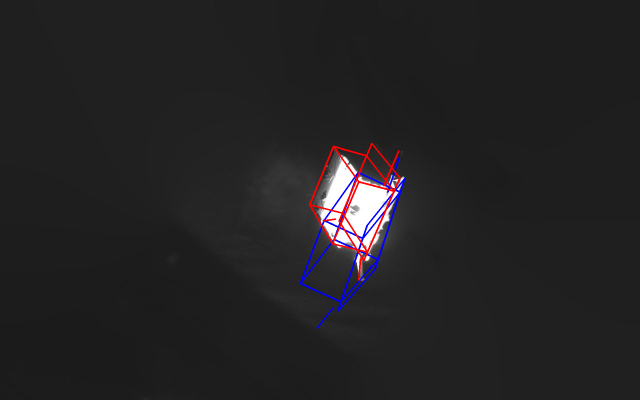}};
  \node[anchor=south west, xshift=-3pt, yshift=-3pt] at (img.south west)
  {\scalebox{0.5}{\colorbox{gray!30}{\begin{tabular}{@{}l@{}} {img000961}\end{tabular}}}};
  \node[anchor=south east, xshift=3pt, yshift=-3pt] at (img.south east)
      {\scalebox{0.5}{\colorbox{gray!30}{\begin{tabular}{@{}l@{}} {$E_r$: 54.03$^{\circ}$} \\ {$E_t$: 1.32\,$m$}\end{tabular}}}};
\end{tikzpicture} &
\begin{tikzpicture}
  \node[inner sep=0pt] (img) at (0,0) {\includegraphics[width=\linewidth]{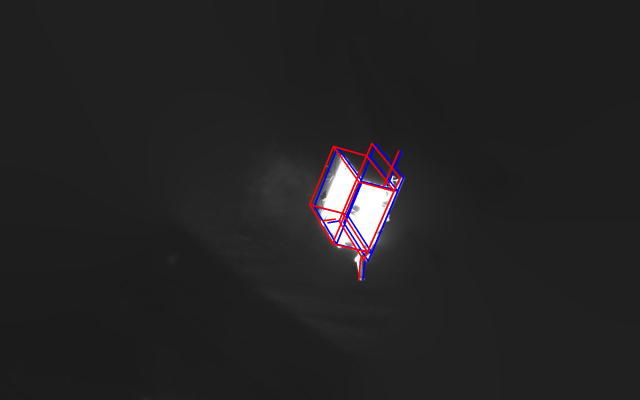}};
  \node[anchor=south west, xshift=-3pt, yshift=-3pt] at (img.south west)
  {\scalebox{0.5}{\begin{tabular}{@{}l@{}} {}\end{tabular}}};
  \node[anchor=south east, xshift=3pt, yshift=-3pt] at (img.south east)
      {\scalebox{0.5}{\colorbox{gray!30}{\begin{tabular}{@{}l@{}} {$E_r$: 1.22$^{\circ}$} \\ {$E_t$: 0.044\,$m$}\end{tabular}}}};
\end{tikzpicture} \\

\begin{tikzpicture}
  \node[inner sep=0pt] (img) at (0,0) {\includegraphics[width=\linewidth]{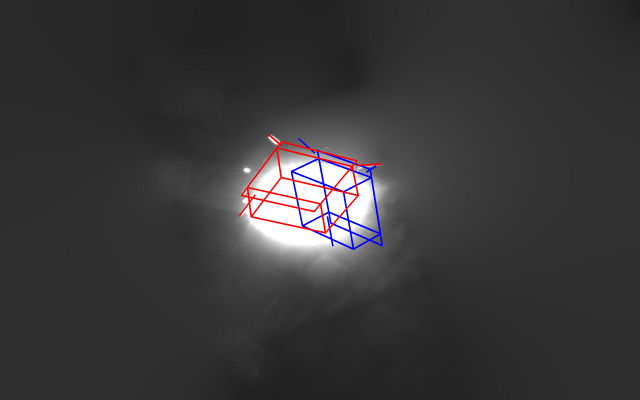}};
  \node[anchor=south west, xshift=-3pt, yshift=-3pt] at (img.south west)
  {\scalebox{0.5}{\colorbox{gray!30}{\begin{tabular}{@{}l@{}} {img000972}\end{tabular}}}};
  \node[anchor=south east, xshift=3pt, yshift=-3pt] at (img.south east)
      {\scalebox{0.5}{\colorbox{gray!30}{\begin{tabular}{@{}l@{}} {$E_r$: 76.31$^{\circ}$} \\ {$E_t$: 0.541\,$m$}\end{tabular}}}};
\end{tikzpicture} &
\begin{tikzpicture}
  \node[inner sep=0pt] (img) at (0,0) {\includegraphics[width=\linewidth]{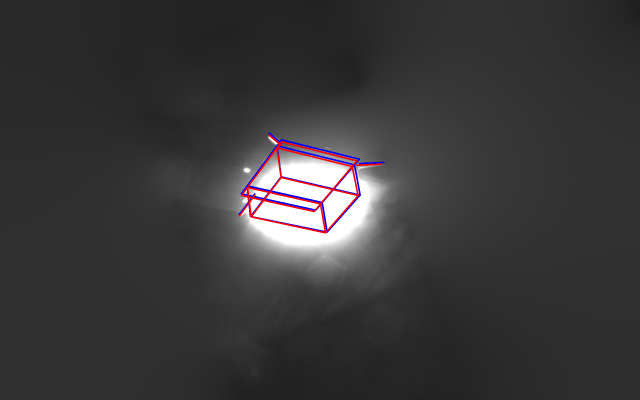}};
  \node[anchor=south west, xshift=-3pt, yshift=-3pt] at (img.south west)
  {\scalebox{0.5}{\begin{tabular}{@{}l@{}} {}\end{tabular}}};
  \node[anchor=south east, xshift=3pt, yshift=-3pt] at (img.south east)
      {\scalebox{0.5}{\colorbox{gray!30}{\begin{tabular}{@{}l@{}} {$E_r$: 1.11$^{\circ}$} \\ {$E_t$: 0.217\,$m$}\end{tabular}}}};
\end{tikzpicture} &
& 
\begin{tikzpicture}
  \node[inner sep=0pt] (img) at (0,0) {\includegraphics[width=\linewidth]{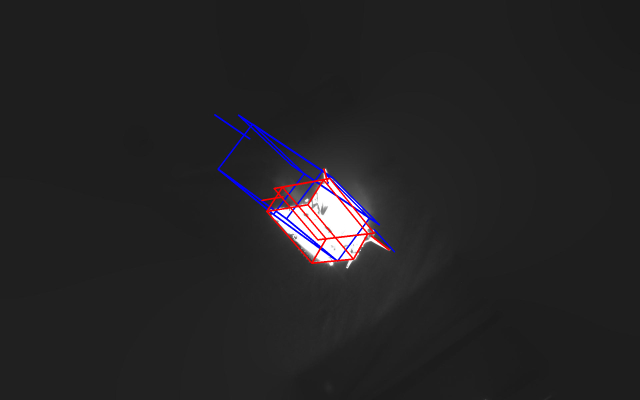}};
  \node[anchor=south west, xshift=-3pt, yshift=-3pt] at (img.south west)
  {\scalebox{0.5}{\colorbox{gray!30}{\begin{tabular}{@{}l@{}} {img001033}\end{tabular}}}};
  \node[anchor=south east, xshift=3pt, yshift=-3pt] at (img.south east)
      {\scalebox{0.5}{\colorbox{gray!30}{\begin{tabular}{@{}l@{}} {$E_r$: 40.16$^{\circ}$} \\ {$E_t$: 3.55\,$m$}\end{tabular}}}};
\end{tikzpicture} &
\begin{tikzpicture}
  \node[inner sep=0pt] (img) at (0,0) {\includegraphics[width=\linewidth]{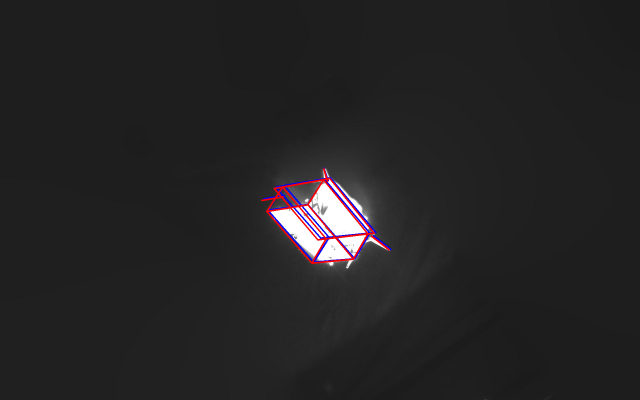}};
  \node[anchor=south west, xshift=-3pt, yshift=-3pt] at (img.south west)
  {\colorbox{gray!30}{\begin{tabular}{@{}l@{}} {}\end{tabular}}};
  \node[anchor=south east, xshift=3pt, yshift=-3pt] at (img.south east)
      {\scalebox{0.5}{\colorbox{gray!30}{\begin{tabular}{@{}l@{}} {$E_r$: 1.83$^{\circ}$} \\ {$E_t$: 0.073\,$m$}\end{tabular}}}};
\end{tikzpicture}  \\

\begin{tikzpicture}
  \node[inner sep=0pt] (img) at (0,0) {\includegraphics[width=\linewidth]{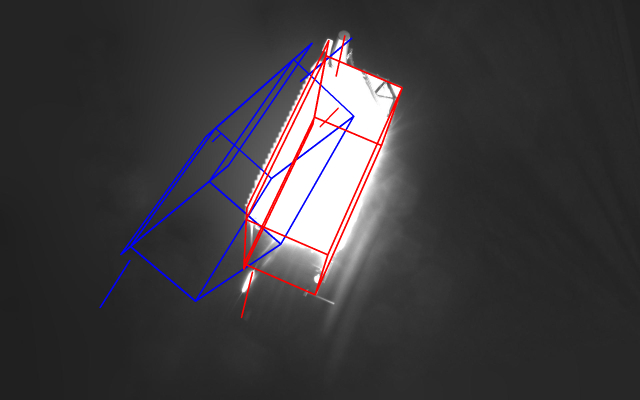}};
  \node[anchor=south west, xshift=-3pt, yshift=-3pt] at (img.south west)
  {\scalebox{0.5}{\colorbox{gray!30}{\begin{tabular}{@{}l@{}} {img001038}\end{tabular}}}};
  \node[anchor=south east, xshift=3pt, yshift=-3pt] at (img.south east)
      {\scalebox{0.5}{\colorbox{gray!30}{\begin{tabular}{@{}l@{}} {$E_r$: 33.65$^{\circ}$} \\ {$E_t$: 0.359\,$m$}\end{tabular}}}};
\end{tikzpicture} &
\begin{tikzpicture}
  \node[inner sep=0pt] (img) at (0,0) {\includegraphics[width=\linewidth]{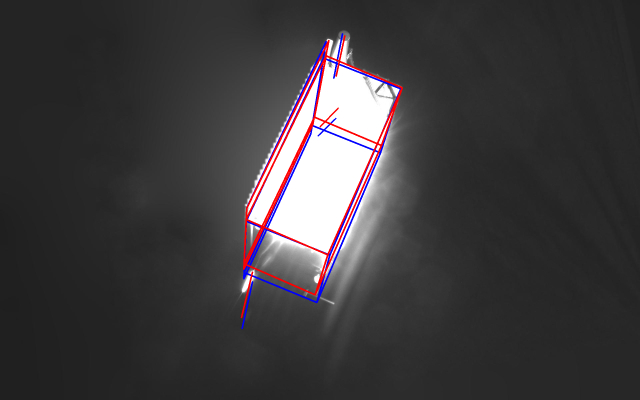}};
  \node[anchor=south west, xshift=-3pt, yshift=-3pt] at (img.south west)
  {\scalebox{0.5}{\begin{tabular}{@{}l@{}} {}\end{tabular}}};
  \node[anchor=south east, xshift=3pt, yshift=-3pt] at (img.south east)
      {\scalebox{0.5}{\colorbox{gray!30}{\begin{tabular}{@{}l@{}} {$E_r$: 3.13$^{\circ}$} \\ {$E_t$: 0.049\,$m$}\end{tabular}}}};
\end{tikzpicture} &
& 
\begin{tikzpicture}
  \node[inner sep=0pt] (img) at (0,0) {\includegraphics[width=\linewidth]{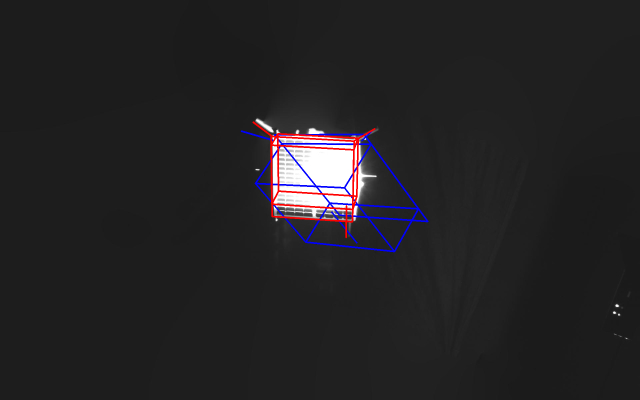}};
  \node[anchor=south west, xshift=-3pt, yshift=-3pt] at (img.south west)
  {\scalebox{0.5}{\colorbox{gray!30}{\begin{tabular}{@{}l@{}} {img001040}\end{tabular}}}};
  \node[anchor=south east, xshift=3pt, yshift=-3pt] at (img.south east)
      {\scalebox{0.5}{\colorbox{gray!30}{\begin{tabular}{@{}l@{}} {$E_r$: 150.7$^{\circ}$} \\ {$E_t$: 3.34\,$m$}\end{tabular}}}};
\end{tikzpicture} &
\begin{tikzpicture}
  \node[inner sep=0pt] (img) at (0,0) {\includegraphics[width=\linewidth]{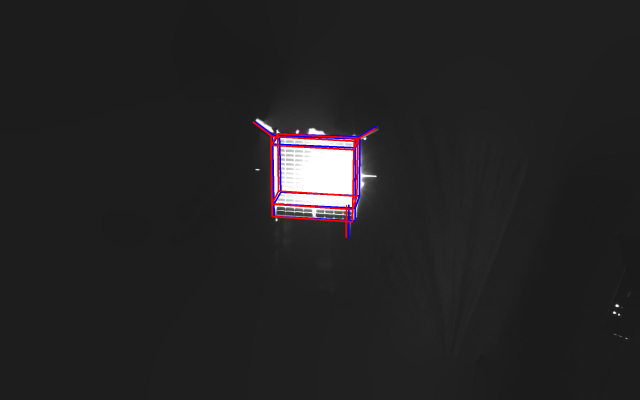}};
  \node[anchor=south west, xshift=-3pt, yshift=-3pt] at (img.south west)
  {\scalebox{0.5}{\begin{tabular}{@{}l@{}} {}\end{tabular}}};
  \node[anchor=south east, xshift=3pt, yshift=-3pt] at (img.south east)
      {\scalebox{0.5}{\colorbox{gray!30}{\begin{tabular}{@{}l@{}} {$E_r$: 2.04$^{\circ}$} \\ {$E_t$: 0.036\,$m$}\end{tabular}}}};
\end{tikzpicture}  \\

\begin{tikzpicture}
  \node[inner sep=0pt] (img) at (0,0) {\includegraphics[width=\linewidth]{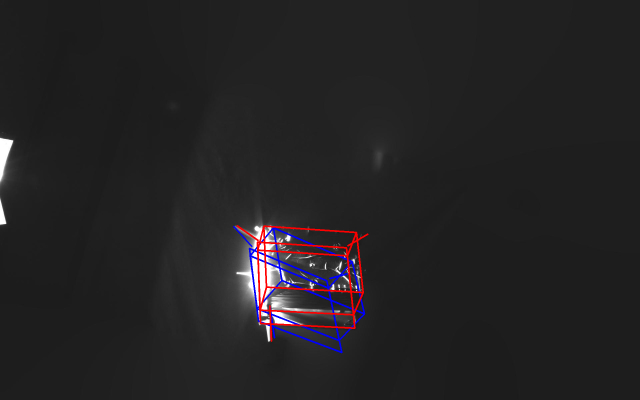}};
  \node[anchor=south west, xshift=-3pt, yshift=-3pt] at (img.south west)
  {\scalebox{0.5}{\colorbox{gray!30}{\begin{tabular}{@{}l@{}} {img002781}\end{tabular}}}};
  \node[anchor=south east, xshift=3pt, yshift=-3pt] at (img.south east)
      {\scalebox{0.5}{\colorbox{gray!30}{\begin{tabular}{@{}l@{}} {$E_r$: 27.9$^{\circ}$} \\ {$E_t$: 0.268\,$m$}\end{tabular}}}};
\end{tikzpicture} & 
\begin{tikzpicture}
  \node[inner sep=0pt] (img) at (0,0) {\includegraphics[width=\linewidth]{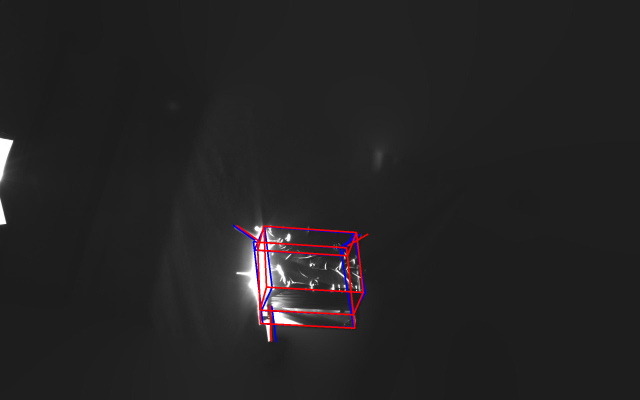}};
  \node[anchor=south west, xshift=-3pt, yshift=-3pt] at (img.south west)
  {\scalebox{0.5}{\begin{tabular}{@{}l@{}} {}\end{tabular}}};
  \node[anchor=south east, xshift=3pt, yshift=-3pt] at (img.south east)
      {\scalebox{0.5}{\colorbox{gray!30}{\begin{tabular}{@{}l@{}} {$E_r$: 6.81$^{\circ}$} \\ {$E_t$: 0.041\,$m$}\end{tabular}}}};
\end{tikzpicture} &
& 
\begin{tikzpicture}
  \node[inner sep=0pt] (img) at (0,0) {\includegraphics[width=\linewidth]{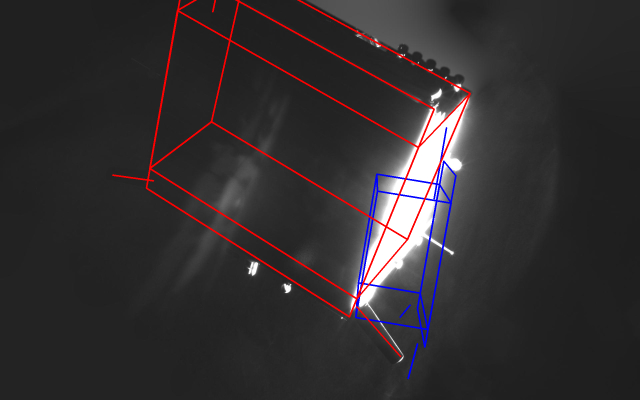}};
  \node[anchor=south west, xshift=-3pt, yshift=-3pt] at (img.south west)
  {\scalebox{0.5}{\colorbox{gray!30}{\begin{tabular}{@{}l@{}} {img001104}\end{tabular}}}};
  \node[anchor=south east, xshift=3pt, yshift=-3pt] at (img.south east)
      {\scalebox{0.5}{\colorbox{gray!30}{\begin{tabular}{@{}l@{}} {$E_r$: 114.5$^{\circ}$} \\ {$E_t$: 1.85\,$m$}\end{tabular}}}};
\end{tikzpicture} &
\begin{tikzpicture}
  \node[inner sep=0pt] (img) at (0,0) {\includegraphics[width=\linewidth]{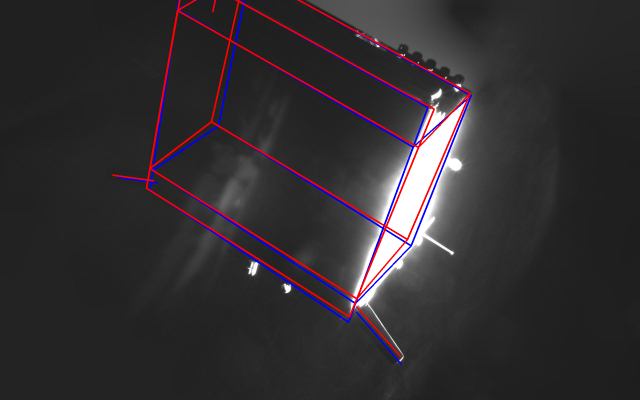}};
  \node[anchor=south west, xshift=-3pt, yshift=-3pt] at (img.south west)
  {\scalebox{0.5}{\begin{tabular}{@{}l@{}} {}\end{tabular}}};
  \node[anchor=south east, xshift=3pt, yshift=-3pt] at (img.south east)
      {\scalebox{0.5}{\colorbox{gray!30}{\begin{tabular}{@{}l@{}} {$E_r$: 3.21$^{\circ}$} \\ {$E_t$: 0.027\,$m$}\end{tabular}}}};
\end{tikzpicture}  \\

\begin{tikzpicture}
  \node[inner sep=0pt] (img) at (0,0) {\includegraphics[width=\linewidth]{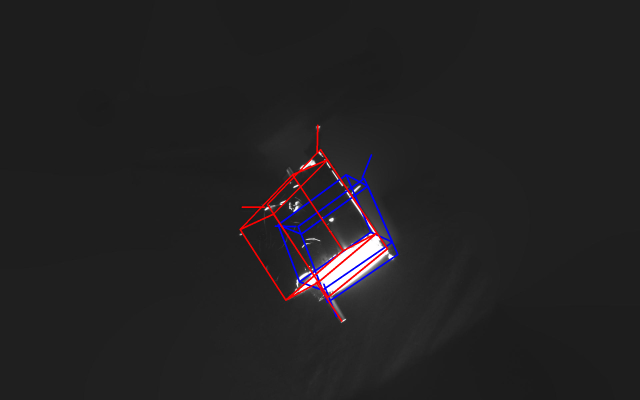}};
  \node[anchor=south west, xshift=-3pt, yshift=-3pt] at (img.south west)
  {\scalebox{0.5}{\colorbox{gray!30}{\begin{tabular}{@{}l@{}} {img001124}\end{tabular}}}};
  \node[anchor=south east, xshift=3pt, yshift=-3pt] at (img.south east)
      {\scalebox{0.5}{\colorbox{gray!30}{\begin{tabular}{@{}l@{}} {$E_r$: 35.2$^{\circ}$} \\ {$E_t$: 1.59\,$m$}\end{tabular}}}};
\end{tikzpicture} &
\begin{tikzpicture}
  \node[inner sep=0pt] (img) at (0,0) {\includegraphics[width=\linewidth]{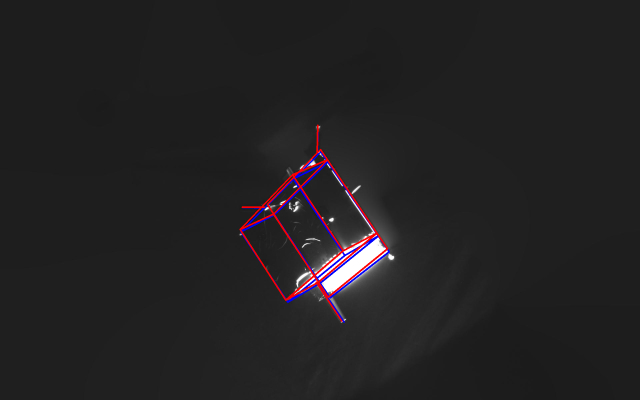}};
  \node[anchor=south west, xshift=-3pt, yshift=-3pt] at (img.south west)
  {\scalebox{0.5}{\begin{tabular}{@{}l@{}} {}\end{tabular}}};
  \node[anchor=south east, xshift=3pt, yshift=-3pt] at (img.south east)
      {\scalebox{0.5}{\colorbox{gray!30}{\begin{tabular}{@{}l@{}} {$E_r$: 1.96$^{\circ}$} \\ {$E_t$: 0.039\,$m$}\end{tabular}}}};
\end{tikzpicture} &
& 
\begin{tikzpicture}
  \node[inner sep=0pt] (img) at (0,0) {\includegraphics[width=\linewidth]{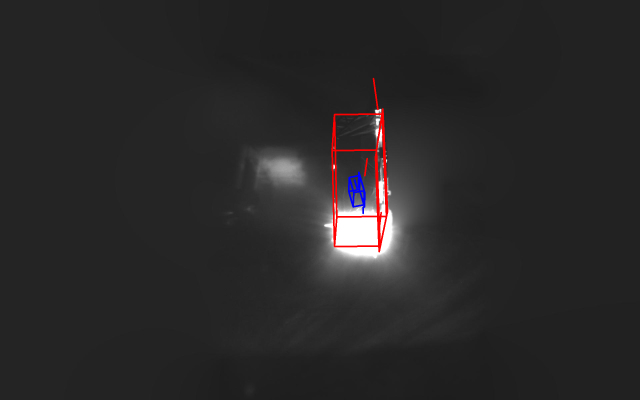}};
  \node[anchor=south west, xshift=-3pt, yshift=-3pt] at (img.south west)
  {\scalebox{0.5}{\colorbox{gray!30}{\begin{tabular}{@{}l@{}} {img002422}\end{tabular}}}};
  \node[anchor=south east, xshift=3pt, yshift=-3pt] at (img.south east)
      {\scalebox{0.5}{\colorbox{gray!30}{\begin{tabular}{@{}l@{}} {$E_r$: 147.3$^{\circ}$} \\ {$E_t$: 22.76\,$m$}\end{tabular}}}};
\end{tikzpicture} &
\begin{tikzpicture}
  \node[inner sep=0pt] (img) at (0,0) {\includegraphics[width=\linewidth]{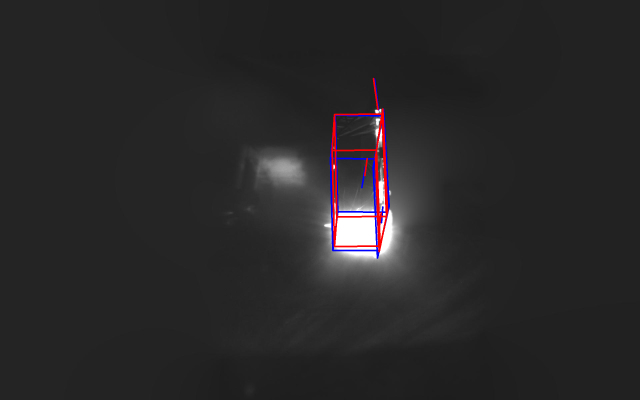}};
  \node[anchor=south west, xshift=-3pt, yshift=-3pt] at (img.south west)
  {\scalebox{0.5}{\begin{tabular}{@{}l@{}} {}\end{tabular}}};
  \node[anchor=south east, xshift=3pt, yshift=-3pt] at (img.south east)
      {\scalebox{0.5}{\colorbox{gray!30}{\begin{tabular}{@{}l@{}} {$E_r$: 6.50$^{\circ}$} \\ {$E_t$: 0.030\,$m$}\end{tabular}}}};
\end{tikzpicture}  \\

\begin{tikzpicture}
  \node[inner sep=0pt] (img) at (0,0) {\includegraphics[width=\linewidth]{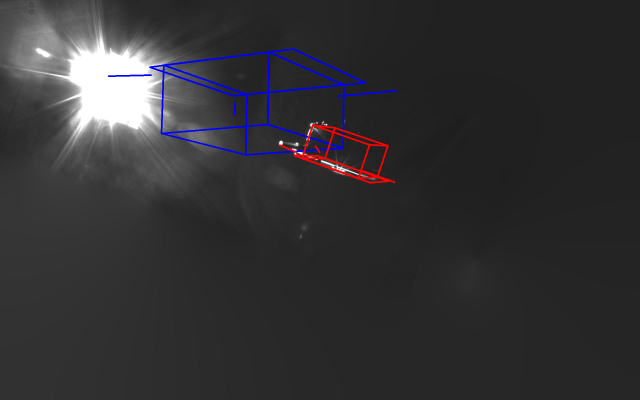}};
  \node[anchor=south west, xshift=-3pt, yshift=-3pt] at (img.south west)
  {\scalebox{0.5}{\colorbox{gray!30}{\begin{tabular}{@{}l@{}} {img002603}\end{tabular}}}};
  \node[anchor=south east, xshift=3pt, yshift=-3pt] at (img.south east)
      {\scalebox{0.5}{\colorbox{gray!30}{\begin{tabular}{@{}l@{}} {$E_r$: 174.2$^{\circ}$} \\ {$E_t$: 4.74\,$m$}\end{tabular}}}};
\end{tikzpicture} &
\begin{tikzpicture}
  \node[inner sep=0pt] (img) at (0,0) {\includegraphics[width=\linewidth]{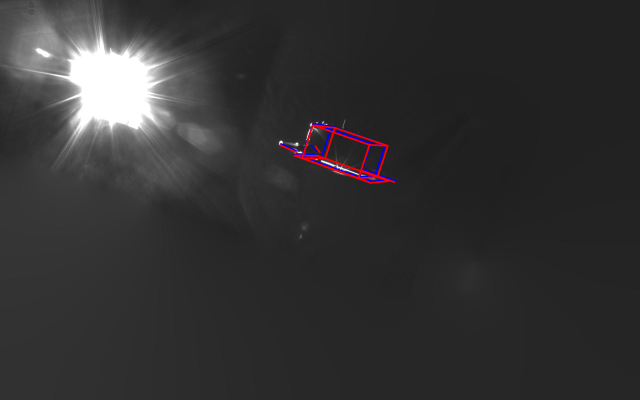}};
  \node[anchor=south west, xshift=-3pt, yshift=-3pt] at (img.south west)
  {\scalebox{0.5}{\begin{tabular}{@{}l@{}} {img000019}\end{tabular}}};
  \node[anchor=south east, xshift=3pt, yshift=-3pt] at (img.south east)
      {\scalebox{0.5}{\colorbox{gray!30}{\begin{tabular}{@{}l@{}} {$E_r$: 2.97$^{\circ}$} \\ {$E_t$: 0.049\,$m$}\end{tabular}}}};
\end{tikzpicture} &
& 
\begin{tikzpicture}
  \node[inner sep=0pt] (img) at (0,0) {\includegraphics[width=\linewidth]{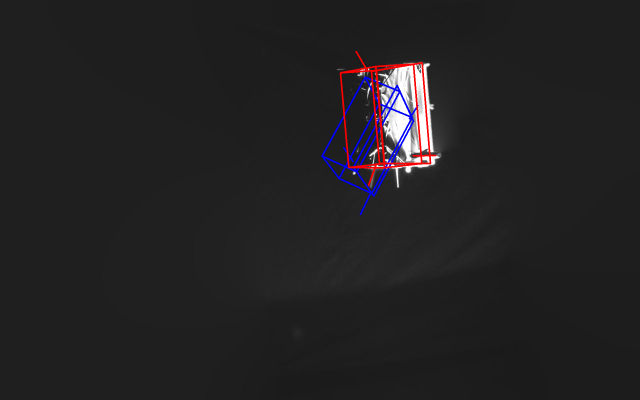}};
  \node[anchor=south west, xshift=-3pt, yshift=-3pt] at (img.south west)
  {\scalebox{0.5}{\colorbox{gray!30}{\begin{tabular}{@{}l@{}} {img002617}\end{tabular}}}};
  \node[anchor=south east, xshift=3pt, yshift=-3pt] at (img.south east)
      {\scalebox{0.5}{\colorbox{gray!30}{\begin{tabular}{@{}l@{}} {$E_r$: 57.2$^{\circ}$} \\ {$E_t$: 0.743\,$m$}\end{tabular}}}};
\end{tikzpicture} &
\begin{tikzpicture}
  \node[inner sep=0pt] (img) at (0,0) {\includegraphics[width=\linewidth]{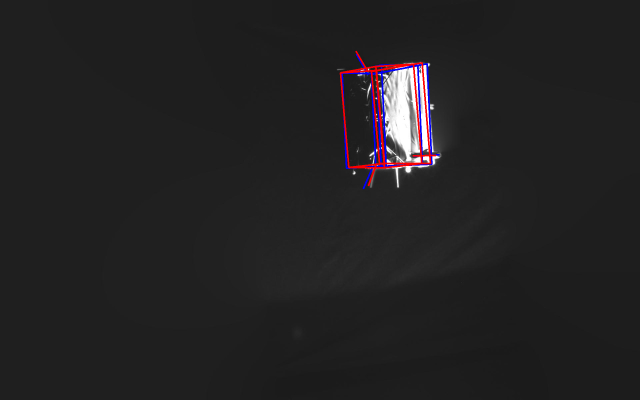}};
  \node[anchor=south west, xshift=-3pt, yshift=-3pt] at (img.south west)
  {\scalebox{0.5}{\begin{tabular}{@{}l@{}} {}\end{tabular}}};
  \node[anchor=south east, xshift=3pt, yshift=-3pt] at (img.south east)
      {\scalebox{0.5}{\colorbox{gray!30}{\begin{tabular}{@{}l@{}} {$E_r$: 5.19$^{\circ}$} \\ {$E_t$: 0.021\,$m$}\end{tabular}}}};
\end{tikzpicture}  \\

\end{tabular}
% \vspace{-1.1em}
\caption{Qualitative comparison of pose predictions on Sunlamp. Each pair shows results without UDA (left) and with \acronym (right). Rotation and translation errors ($E_r$, $E_t$) are reported. Ground-truth poses are shown in \textcolor{red}{red}, and predictions in \textcolor{blue}{blue}.}
\end{center}
\end{figure*}

%% file: references.bib
@inproceedings{denninger2020blenderproc,
  title={Blenderproc: Reducing the reality gap with photorealistic rendering},
  author={Denninger, Maximilian and Sundermeyer, Martin and Winkelbauer, Dominik and Olefir, Dmitry and Hodan, Tomas and Zidan, Youssef and Elbadrawy, Mohamad and Knauer, Markus and Katam, Harinandan and Lodhi, Ahsan},
  booktitle={16th Robotics: Science and Systems, RSS 2020, Workshops},
  year={2020}
}

@InProceedings{farahani2021,
author="Farahani, Abolfazl
and Voghoei, Sahar
and Rasheed, Khaled
and Arabnia, Hamid R.",
editor="Stahlbock, Robert
and Weiss, Gary M.
and Abou-Nasr, Mahmoud
and Yang, Cheng-Ying
and Arabnia, Hamid R.
and Deligiannidis, Leonidas",
title="A Brief Review of Domain Adaptation",
booktitle="Advances in Data Science and Information Engineering",
year="2021",
publisher="Springer International Publishing",
address="Cham",
pages="877--894",
}

@inproceedings{hinterstoisser2012model,
  title={Model based training, detection and pose estimation of texture-less 3d objects in heavily cluttered scenes},
  author={Hinterstoisser, Stefan and Lepetit, Vincent and Ilic, Slobodan and Holzer, Stefan and Bradski, Gary and Konolige, Kurt and Navab, Nassir},
  booktitle={Asian conference on computer vision},
  year={2012},
  organization={Springer}
}

@INPROCEEDINGS{speedplus2022,
  author={Park, Tae Ha and Märtens, Marcus and Lecuyer, Gurvan and Izzo, Dario and D'Amico, Simone},
  booktitle={2022 IEEE Aerospace Conference (AERO)}, 
  title={SPEED+: Next-Generation Dataset for Spacecraft Pose Estimation across Domain Gap}, 
  year={2022},
  volume={},
  number={},
  pages={1-15},
  doi={10.1109/AERO53065.2022.9843439}}

@inproceedings{zhao2019learning,
  title={On learning invariant representations for domain adaptation},
  author={Zhao, Han and Des Combes, Remi Tachet and Zhang, Kun and Gordon, Geoffrey},
  booktitle={ICML},
  year={2019},
  organization={PMLR}
}

@article{park2024robust,
  title={Robust multi-task learning and online refinement for spacecraft pose estimation across domain gap},
  author={Park, Tae Ha and D’Amico, Simone},
  journal={Advances in Space Research},
  year={2024},
  publisher={Elsevier}
}

@inproceedings{wang2021gdr,
  title={Gdr-net: Geometry-guided direct regression network for monocular 6d object pose estimation},
  author={Wang, Gu and Manhardt, Fabian and Tombari, Federico and Ji, Xiangyang},
  booktitle={Proceedings of the IEEE/CVF Conference on CVPR},
  year={2021}
}

@inproceedings{zhang2023manifold,
  title={Manifold-aware self-training for unsupervised domain adaptation on regressing 6D object pose},
  author={Zhang, Yichen and Lin, Jiehong and Chen, Ke and Xu, Zelin and Wang, Yaowei and Jia, Kui},
  booktitle={Proceedings of the International Joint Conference on Artificial Intelligence},
  year={2023}
}

@article{wang2021occlusion,
  title={Occlusion-aware self-supervised monocular 6D object pose estimation},
  author={Wang, Gu and Manhardt, Fabian and Liu, Xingyu and Ji, Xiangyang and Tombari, Federico},
  journal={IEEE Transactions on Pattern Analysis and Machine Intelligence},
  volume={46},
  pages={1788--1803},
  year={2021},
  publisher={IEEE}
}

@article{yang2024pvspe,
  title={PVSPE: A pyramid vision multitask transformer network for spacecraft pose estimation},
  author={Yang, Hong and Xiao, Xueming and Yao, Meibao and Xiong, Yonggang and Cui, Hutao and Fu, Yuegang},
  journal={Advances in Space Research},
  year={2024},
  publisher={Elsevier}
}

@inproceedings{yang2024cod,
  title={COD: Learning Conditional Invariant Representation for Domain Adaptation Regression},
  author={Yang, Hao-Ran and Ren, Chuan-Xian and Luo, You-Wei},
  booktitle={European Conference on Computer Vision},
  pages={108--125},
  year={2024},
  organization={Springer}
}

@article{oord2018representation,
  title={Representation learning with contrastive predictive coding},
  author={Oord, Aaron van den and Li, Yazhe and Vinyals, Oriol},
  journal={arXiv preprint arXiv:1807.03748},
  year={2018}
}

@INPROCEEDINGS{xiang2018posecnn, 
    AUTHOR    = {Yu Xiang AND Tanner Schmidt AND Venkatraman Narayanan AND Dieter Fox}, 
    TITLE     = {PoseCNN: A Convolutional Neural Network for 6D Object Pose Estimation in Cluttered Scenes}, 
    BOOKTITLE = {Proceedings of Robotics: Science and Systems}, 
    YEAR      = {2018}, 
    ADDRESS   = {Pittsburgh, Pennsylvania}, 
    MONTH     = {June}, 
    DOI       = {10.15607/RSS.2018.XIV.019} 
}

@article{xu2024unsupervised,
  title={Unsupervised Domain Adaptation with Contrastive Learning-Based Discriminative Feature Augmentation for RS Image Classification},
  author={Xu, Ren and Samat, Alim and Zhu, Enzhao and Li, Erzhu and Li, Wei},
  journal={Remote Sensing},
  volume={16},
  number={11},
  pages={1974},
  year={2024},
  publisher={MDPI}
}

@article{bauer2024challenges,
  title={Challenges for monocular 6d object pose estimation in robotics},
  author={Bauer, Dominik and H{\"o}nig, Peter and Weibel, Jean-Baptiste and Garc{\'\i}a-Rodr{\'\i}guez, Jos{\'e} and Vincze, Markus and others},
  journal={IEEE Transactions on Robotics},
  year={2024},
  publisher={IEEE}
}

@inproceedings{jiang2021regressive,
  title={Regressive domain adaptation for unsupervised keypoint detection},
  author={Jiang, Junguang and Ji, Yifei and Wang, Ximei and Liu, Yufeng and Wang, Jianmin and Long, Mingsheng},
  booktitle={Proceedings of the IEEE/CVF Conference on Computer Vision and Pattern Recognition},
  pages={6780--6789},
  year={2021}
}

@inproceedings{ikeda2024diffusionnocs,
  title={Diffusionnocs: Managing symmetry and uncertainty in sim2real multi-modal category-level pose estimation},
  author={Ikeda, Takuya and Zakharov, Sergey and Ko, Tianyi and Irshad, Muhammad Zubair and Lee, Robert and Liu, Katherine and Ambrus, Rares and Nishiwaki, Koichi},
  booktitle={2024 IEEE/RSJ International Conference on Intelligent Robots and Systems (IROS)},
  pages={7406--7413},
  year={2024},
  organization={IEEE}
}

@article{bechini2025robust,
  title={Robust and efficient single-CNN-based spacecraft relative pose estimation from monocular images},
  author={Bechini, Michele and Lavagna, Mich{\`e}le},
  journal={Acta Astronautica},
  year={2025},
  publisher={Elsevier}
}

@ARTICLE{9991175,

  author={Zhao, Chunhui and Qin, Boao and Feng, Shou and Zhu, Wenxiang and Zhang, Lifu and Ren, Jinchang},

  journal={IEEE Transactions on Geoscience and Remote Sensing}, 

  title={An Unsupervised Domain Adaptation Method Towards Multi-Level Features and Decision Boundaries for Cross-Scene Hyperspectral Image Classification}, 

  year={2022},

  volume={60},

  number={},

  pages={1-16},

  doi={10.1109/TGRS.2022.3230378}}

@inproceedings{
keramati2024conr,
title={ConR: Contrastive Regularizer for Deep Imbalanced Regression},
author={Mahsa Keramati and Lili Meng and R. David Evans},
booktitle={The Twelfth International Conference on Learning Representations},
year={2024},
}

@inproceedings{vasconcelos2021shape,
  title={Shape consistent 2D keypoint estimation under domain shift},
  author={Vasconcelos, Levi O and Mancini, Massimiliano and Boscaini, Davide and Bulo, Samuel Rota and Caputo, Barbara and Ricci, Elisa},
  booktitle={2020 25th International Conference on Pattern Recognition (ICPR)},
  pages={8037--8044},
  year={2021},
  organization={IEEE}
}

@inproceedings{zhang2019bridging,
  title={Bridging theory and algorithm for domain adaptation},
  author={Zhang, Yuchen and Liu, Tianle and Long, Mingsheng and Jordan, Michael},
  booktitle={ICML},
  pages={7404--7413},
  year={2019},
  organization={PMLR}
}

@inproceedings{loshchilov2017decoupled,
title={Decoupled Weight Decay Regularization},
author={Ilya Loshchilov and Frank Hutter},
booktitle={International Conference on Learning Representations},
year={2019},
}

@InProceedings{Xiao_2018_ECCV,
author = {Xiao, Bin and Wu, Haiping and Wei, Yichen},
title = {Simple Baselines for Human Pose Estimation and Tracking},
booktitle = {Proceedings of the European Conference on Computer Vision (ECCV)},
month = {September},
year = {2018}
}

@inproceedings{luo2021rethinking,
  title={Rethinking the heatmap regression for bottom-up human pose estimation},
  author={Luo, Zhengxiong and Wang, Zhicheng and Huang, Yan and Wang, Liang and Tan, Tieniu and Zhou, Erjin},
  booktitle={Proceedings of the IEEE/CVF conference on computer vision and pattern recognition},
  year={2021}
}

@inproceedings{chen2020simple,
  title={A simple framework for contrastive learning of visual representations},
  author={Chen, Ting and Kornblith, Simon and Norouzi, Mohammad and Hinton, Geoffrey},
  booktitle={International conference on machine learning},
  pages={1597--1607},
  year={2020},
  organization={PmLR}
}

@inproceedings{tan2025onda,
  title={ONDA-Pose: Occlusion-Aware Neural Domain Adaptation for Self-Supervised 6D Object Pose Estimation},
  author={T.Tan and Q.Dong},
  booktitle={Proceedings of the Computer Vision and Pattern Recognition Conference},
  year={2025}
}

@inproceedings{chen2023texpose,
  title={Texpose: Neural texture learning for self-supervised 6d object pose estimation},
  author={Chen, Hanzhi and Manhardt, Fabian and Navab, Nassir and Busam, Benjamin},
  booktitle={Proceedings of the IEEE/CVF Conference on CVPR},
  year={2023}
}

@inproceedings{tan2023smoc,
  title={SMOC-Net: leveraging camera pose for self-supervised monocular object pose estimation},
  author={Tan, Tao and Dong, Qiulei},
  booktitle={Proceedings of the IEEE/CVF Conference on CVPR},
  year={2023}
}

@inproceedings{brachmann2014learning,
  title={Learning 6d object pose estimation using 3d object coordinates},
  author={Brachmann, Eric and Krull, Alexander and Michel, Frank and Gumhold, Stefan and Shotton, Jamie and Rother, Carsten},
  booktitle={European conference on computer vision},
  pages={536--551},
  year={2014},
  organization={Springer}
}

@inproceedings{hodavn2019photorealistic,
  title={Photorealistic image synthesis for object instance detection},
  author={Hoda{\v{n}}, Tom{\'a}{\v{s}} and Vineet, Vibhav and Gal, Ran and Shalev, Emanuel and Hanzelka, Jon and Connell, Treb and Urbina, Pedro and Sinha, Sudipta N and Guenter, Brian},
  booktitle={2019 IEEE international conference on image processing (ICIP)},
  pages={66--70},
  year={2019},
  organization={IEEE}
}

@inproceedings{kaskman2019homebreweddb,
  title={Homebreweddb: RGB-D dataset for 6d pose estimation of 3d objects. In 2019 IEEE},
  author={Kaskman, Roman and Zakharov, Sergey and Shugurov, Ivan and Ilic, Slobodan},
  booktitle={CVF International Conference on Computer Vision Workshop (ICCVW)},
  year={2019}
}

@inproceedings{hodavn2020bop,
  title={BOP challenge 2020 on 6D object localization},
  author={Hoda{\v{n}}, Tom{\'a}{\v{s}} and Sundermeyer, Martin and Drost, Bertram and Labb{\'e}, Yann and Brachmann, Eric and Michel, Frank and Rother, Carsten and Matas, Ji{\v{r}}{\'\i}},
  booktitle={European Conference on Computer Vision},
  year={2020}
}

@article{hinterstoisser2011gradient,
  title={Gradient response maps for real-time detection of textureless objects},
  author={Hinterstoisser, Stefan and Cagniart, Cedric and Ilic, Slobodan and Sturm, Peter and Navab, Nassir and Fua, Pascal and Lepetit, Vincent},
  journal={IEEE transactions on pattern analysis and machine intelligence},
  volume={34},
  number={5},
  pages={876--888},
  year={2011},
  publisher={IEEE}
}

@article{shugurov2021dpodv2,
  title={Dpodv2: Dense correspondence-based 6 dof pose estimation},
  author={Shugurov, Ivan and Zakharov, Sergey and Ilic, Slobodan},
  journal={IEEE transactions on pattern analysis and machine intelligence},
  volume={44},
  number={11},
  year={2021},
  publisher={IEEE}
}

@inproceedings{labbe2020cosypose,
  title={Cosypose: Consistent multi-view multi-object 6d pose estimation},
  author={Labb{\'e}, Yann and Carpentier, Justin and Aubry, Mathieu and Sivic, Josef},
  booktitle={European Conference on Computer Vision},
  pages={574--591},
  year={2020},
  organization={Springer}
}

@article{yanfang2024feature,
  title={Feature-aided pose estimation approach based on variational auto-encoder structure for spacecrafts},
  author={Yanfang, LIU and Rui, ZHOU and Desong, DU and Shuqing, CAO and Naiming, QI},
  journal={Chinese Journal of Aeronautics},
  volume={37},
  number={8},
  pages={329--341},
  year={2024},
  publisher={Elsevier}
}

@article{ousalah2025uncertainty,
  title={Uncertainty-Aware Knowledge Distillation for Compact and Efficient 6DoF Pose Estimation},
  author={{Ali Ousalah}, Nassim and Kacem, Anis and Ghorbel, Enjie and Koumandakis, Emmanuel and Aouada, Djamila},
  journal={International Conference on Intelligent Robots and Systems (IROS)},
  year={2025}
}

@article{chen2021towards,
  title={Towards self-similarity consistency and feature discrimination for unsupervised domain adaptation},
  author={Chen, Chao and Fu, Zhihang and Chen, Zhihong and Cheng, Zhaowei and Jin, Xinyu},
  journal={Signal Processing: Image Communication},
  volume={94},
  pages={116232},
  year={2021},
  publisher={Elsevier}
}

@article{zhang2019category,
  title={Category anchor-guided unsupervised domain adaptation for semantic segmentation},
  author={Zhang, Qiming and Zhang, Jing and Liu, Wei and Tao, Dacheng},
  journal={Advances in neural information processing systems},
  volume={32},
  year={2019}
}

@article{wang2023bridging,
  title={Bridging the domain gap in satellite pose estimation: A self-training approach based on geometrical constraints},
  author={Wang, Zi and Chen, Minglin and Guo, Yulan and Li, Zhang and Yu, Qifeng},
  journal={IEEE transactions on aerospace and electronic systems},
  year={2023},
  publisher={IEEE}
}

@article{park2023satellite,
  title={Satellite pose estimation competition 2021: Results and analyses},
  author={Park, Tae Ha and M{\"a}rtens, Marcus and Jawaid, Mohsi and Wang, Zi and Chen, Bo and Chin, Tat-Jun and Izzo, Dario and D’Amico, Simone},
  journal={Acta Astronautica},
  year={2023},
  publisher={Elsevier}
}

@inproceedings{perez2022spacecraft,
  title={Spacecraft pose estimation based on unsupervised domain adaptation and on a 3d-guided loss combination},
  author={P{\'e}rez-Villar, Juan Ignacio Bravo and Garc{\'\i}a-Mart{\'\i}n, {\'A}lvaro and Besc{\'o}s, Jes{\'u}s},
  booktitle={ECCV},
  year={2022},
}

@article{schwonberg2023survey,
  title={Survey on unsupervised domain adaptation for semantic segmentation for visual perception in automated driving},
  author={Schwonberg, Manuel and Niemeijer, Joshua and Term{\"o}hlen, Jan-Aike and Schmidt, Nico M and Gottschalk, Hanno and Fingscheidt, Tim and others},
  journal={IEEE Access},
  volume={11},
  pages={54296--54336},
  year={2023},
  publisher={IEEE}
}

@inproceedings{pitkevich2024survey,
  title={A survey on Sim-to-Real transfer methods for robotic manipulation},
  author={Pitkevich, Andrei and Makarov, Ilya},
  booktitle={2024 IEEE 22nd Jubilee International Symposium on Intelligent Systems and Informatics (SISY)},
  year={2024}
}

@article{cassinis2023leveraging,
  title={Leveraging neural network uncertainty in adaptive unscented Kalman Filter for spacecraft pose estimation},
  author={Cassinis, Lorenzo Pasqualetto and Park, Tae Ha and Stacey, Nathan and D’Amico, Simone and Menicucci, Alessandra and Gill, Eberhard and Ahrns, Ingo and Sanchez-Gestido, Manuel},
  journal={Advances in Space Research},
  pages={5061--5082},
  year={2023},
  publisher={Elsevier}
}

@article{park2023adaptive,
  title={Adaptive neural-network-based unscented kalman filter for robust pose tracking of noncooperative spacecraft},
  author={Park, Tae Ha and D’Amico, Simone},
  journal={Journal of Guidance, Control, and Dynamics},
  pages={1671--1688},
  year={2023},
  publisher={American Institute of Aeronautics and Astronautics}
}
